\documentclass[twoside,11pt]{article}
\usepackage{blindtext}
\usepackage{subcaption}
\usepackage[percent]{overpic} 
\usepackage[abbrvbib, preprint]{jmlr2e}
\usepackage{amsmath, amssymb}
\usepackage{bm}
\usepackage{bbm} 
\usepackage{dsfont} 
\usepackage{mathtools} 
\usepackage{algorithm}
\usepackage{algpseudocode}
\usepackage[HTML]{xcolor}
\usepackage{pifont}
\usepackage{xurl} 
\usepackage{hyperref}
\hypersetup{breaklinks=true}
\newtheorem{example}{Example} 
\newtheorem{theorem}{Theorem}
\newtheorem{lemma}{Lemma} 
\newtheorem{proposition}{Proposition} 
\newtheorem{remark}{Remark}
\newtheorem{corollary}{Corollary}[theorem]
\newtheorem{definition}{Definition}

\newtheorem{assumptions}{Assumption}

\newcommand{\prob}[0]{\mathbf{Pr}}
\newcommand{\ntrain}[0]{n_{\mathrm{train}}}
\newcommand{\bt}{\boldsymbol{\theta}} 

\newcommand{\x}{\boldsymbol{x}}
\newcommand{\Y}{\boldsymbol{Y}}
\newcommand{\s}{\boldsymbol{s}}

\newcommand{\X}{\boldsymbol{X}}

\newcommand{\calb}[1]{#1_{\mathrm{cal}}}

\newcommand{\psibias}[0]{\psi}
\newcommand{\psibiashat}[0]{\widehat{\psi}}
\newcommand{\psibar}[0]{\overline{\psi}}
\newcommand{\psitrue}[0]{\psi_0}
\newcommand{\kernel}[2]{\mathrm{ker}_{\lambda} (#1, \, #2) }
\newcommand{\dotprod}[3]{ \langle #1, \, #2 \rangle_{#3}}
\newcommand{\bx}[0]{\boldsymbol{x}}
\newcommand{\bX}[0]{\boldsymbol{X}}
\newcommand{\yhatpsi}[0]{\hat{y}_{\psi}}
\newcommand{\ncal}[0]{n_{\mathrm{cal}}}
\newcommand{\cmark}{\ding{51}}
\newcommand{\xmark}{\ding{55}}

\newcommand{\R}{\mathbb{R}}

\newcommand{\emp}[1]{\textit{#1}}
\newcommand{\defeq}{\vcentcolon=}

\usepackage{lastpage}
\jmlrheading{23}{2025}{1-\pageref{LastPage}}{1/21; Revised 5/22}{9/22}{21-0000}{Daniel Salnikov, Dan Leonte and Kevin Michalewicz}

\ShortHeadings{MAPS: model-agnostic prediction sets for supervised learning}{salnikov, leonte and michalewicz}
\firstpageno{1}

\title{The MAPS Algorithm: Fast model-agnostic and distribution-free prediction intervals for supervised learning}

\author{\name\small Daniel Salnikov \email\small d.salnikov22@imperial.ac.uk\\
       \addr\small Department of Mathematics, Imperial College London\\
       180 Queen's Gate, London, SW7 2AZ, UK\\
       Great Ormond Street Institute of Child Health\\
       30 Guilford Street, London, WC1N 1EH, UK
       \AND
       \name\small Dan Leonte \email\small leonted@kaust.edu.sa\\
       \addr\small  Department of Statistics \\
       King Abdullah University of Science and Technology\\
       Thuwal, Saudi Arabia
       \AND
       \name\small Kevin Michalewicz \email\small k.michalewicz22@imperial.ac.uk \\
       \addr\small Department of Mathematics, Imperial College London \\
       180 Queen's Gate, London, SW7 2AZ, UK\\
        Instituto de Ingeniería Biomédica, Universidad de Buenos Aires\\
        Av. Paseo Colón 850, Buenos Aires C1063ACV, Argentina
        }
\editor{Chancellor Palpatine}

\begin{document}

\maketitle
\vspace{-15 pt}
\begin{abstract}%
\small{
A fundamental problem in modern supervised learning is computing reliable conditional prediction intervals in high-dimensional settings: existing methods often rely on restrictive modelling assumptions, do not scale as predictor dimension increases, or only guarantee marginal (population-level) rather than conditional (individual-level) coverage. We introduce the \emp{lifted predictive model} (LPM), a new conditional representation, and propose the MAPS (Model-Agnostic Prediction Sets) algorithm that produces distribution-free conditional prediction intervals and adapts to any trained predictive model. Our procedure is bootstrap-based, scales to high-dimensional inputs and accounts for heteroscedastic errors. We establish the theoretical properties of the LPM, connect prediction accuracy to interval length, and provide sufficient conditions for asymptotic conditional coverage. We evaluate the finite-sample performance of MAPS in a simulation study, and apply our method to simulation-based inference and image classification. In the former, MAPS provides the first approach for debiasing neural Bayes estimators and constructing 
valid confidence intervals for model parameters given the estimators, at any desired level. In the latter, it provides the first approach that accounts for uncertainty in model calibration and label prediction.
%
}
\vspace{-10 pt}
\end{abstract}

\addtocontents{toc}{\protect\setcounter{tocdepth}{0}{}}
\section{Introduction}\label{sec:intro}
\vspace{-10 pt}
Accurate and scalable uncertainty quantification is a core challenge in modern statistical machine learning. Practitioners increasingly rely on complex predictive models, for example, regularised linear models, gradient-boosted trees, or artificial neural networks. However, computing trustworthy prediction intervals that (i) reflect the model's actual predictive accuracy, (ii) scale to high-dimensional predictors, and (iii) do not require strong distributional assumptions, remains difficult~\citep{barber2020limitsdistributionfreeconditionalpredictive}. In this work, we introduce the lifted predictive model (LPM) framework and the computationally efficient Model-Agnostic Prediction Sets (MAPS) algorithm that produces distribution-free, optimal prediction intervals with valid asymptotic \emp{conditional} coverage for any trained pointwise predictive model.
\subsection{Notation} 
 Let $Y \in \mathcal{Y} \subseteq \mathbb{R}$ be a response (or target), 
 and $\boldsymbol{X} \in \mathcal{X} \subseteq \mathbb{R}^d$ a $d$-dimensional predictor (or covariate) 
 variable generated by an unknown joint distribution, say $P_{\X, Y}$, such that the conditional and predictor distributions $P_{Y | \bX}$ and $P_{\bX}$ 
 are absolutely continuous with respect to the Lebesgue measure. In supervised settings, we are interested in the regression function $f: \mathbb{R}^d \rightarrow \mathbb{R}$ that minimises a data-fidelity loss, e.g., the conditional mean-squared prediction error (MSPE)
\begin{equation}\label{eq: mspe definition}
    \mathbb{E} \, [(Y-f(\X))^2\mid \bX=\bx]. 
\end{equation}
\par
The idealised pointwise {\em minimiser} of~\eqref{eq: mspe definition} given an {\em out-of-sample} realisation $\bX_o = \x_o$, i.e., not ``seen" during training, is the regression function value $f(\x_o)=\mathbb{E} \, [Y\mid \X=\x_o]$, where $f$ has a finite $L_2$- or $L_1$-norm. In practice, $f$ is unknown and is replaced by a predictive model obtained from a learning algorithm applied to the training data~\citep[Chapter 2]{statLearn}. Such procedures may be parametric (e.g., linear regression), nonparametric (e.g., kernel methods), or algorithmic (e.g., neural networks). Applying our chosen procedure~\citep[see, e.g.,][]{deep_learn} to the training set yields a pointwise predictive model:
$$
\hat f \leftarrow \texttt{procedure}\{(\X_i, \, Y_i)_{\mathrm{train}}, \, i = 1, \ldots, \ntrain \}, \, \, \, \, \text{such} \, \, \text{that} \, \, \, \, \hat{f}: \mathbb{R}^d \rightarrow \mathbb{R},
$$
and the resulting pointwise predictions $\hat{f} (\bx_o) \equiv \hat{y}_o$ given $\bX_o = \bx_o$ and $\hat{f}$. 
\subsection{Motivation}
Our objective is to compute reliable model-agnostic prediction intervals for an out-of-sample response $Y_o$, which are valid for any continuous $\hat{f}$, e.g., a multilayer perceptron $\hat{f}_{\mathrm{MLP}}$, a support vector machine $\hat{f}_{\mathrm{SVM}}$, or a gradient boosted tree $\hat{f}_{\mathrm{GBT}}$. Given a confidence level $\alpha \in (0, 1)$, we seek a set $\widehat{C}(\bx_o) = [\hat{c}_1 (\bx_o), \, \hat{c}_2 (\bx_o) ] \subset \mathbb{R}$, where $\hat{c}_1, \, \hat{c}_2 : \mathcal{X} \to \mathbb{R}$ are ``learned" functions with valid distribution-free conditional coverage:
\begin{equation}\label{eq: general conditional interval}
    \mathbf{Pr} ( Y_o \in \widehat{C}(\bx_o) \, | \, \bX_o = \bx_o ) \geq 1 - \alpha, \, \forall \x_o \in \mathcal{X}.
\end{equation}
\par
In our framework, we exploit the information encoded in \(\hat f\) and an independent calibration set to construct prediction sets for \(Y_o\) with strong theoretical guarantees, namely valid asymptotic conditional coverage; see Theorem~\ref{th: aymp validity of MAPS}. The MAPS algorithm does not require $f$ to satisfy any particular (non)parametric assumptions or to impose constraints on $\hat{f}$. Instead, MAPS adapts interval width depending on how accurate \(\hat f\) is at a point \(\bx\). If \(\hat f\) is accurate, then the intervals are tighter, whereas if $\hat{f}$ is inaccurate, the intervals widen to reflect increased uncertainty. 
Rather than 
estimating the full conditional distribution of \(Y_o\) given the high-dimensional predictor $\bX_o \in \R^d$, which is difficult in large \(d\)~\citep{JMLR:v24:22-1217}, MAPS 
exploits a new lower-dimensional, conditional relation 
between \(Y_o\) and $\hat{f} (\bX_o)$, which we refer to as the lifted predictive model (LPM). By focusing on this relation
, MAPS scales 
to high-dimensional settings {\em and} adapts interval width to the model's local accuracy.
\par
We stress that the independent calibration set: $\{ (\X_{\calb{i}}, \, Y_{\calb{i}}), \, \calb{i} = 1\, \ldots, \, n_{\mathrm{cal}} \}$, composed of $\calb{n}$ ``new" response and predictor pairs 
must remain {\em unused} during model fitting. This is a common requirement for methods based on data-splitting techniques~\citep{stone_cv}. Our procedure does not require access to the training set; intuitively, it ``wraps around" any trained $\hat{f}$ and quantifies its uncertainty in a post-training step.
\begin{figure}
    \centering
\includegraphics[width=0.65\linewidth]{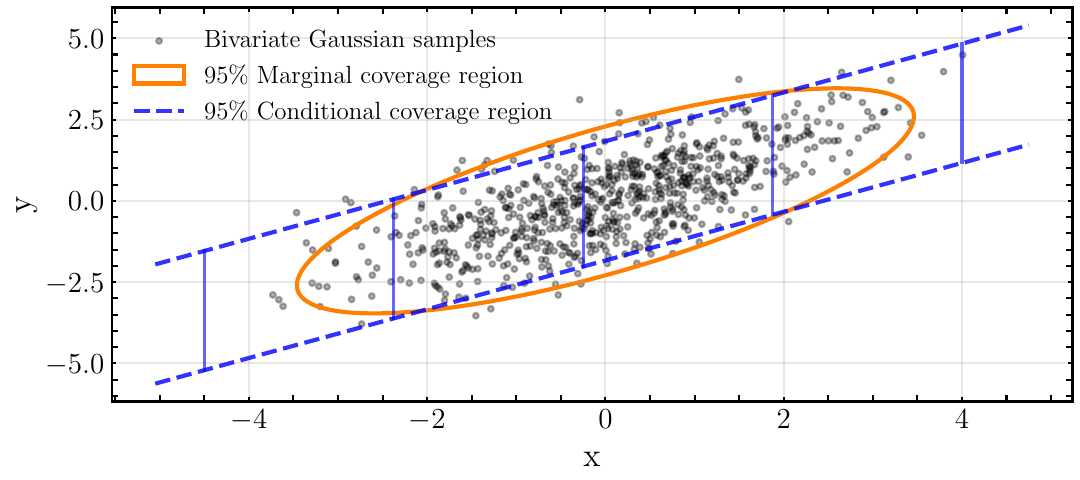}     \vspace{-12 pt}
   \caption{{\small Comparison of conditional (\textcolor{blue}{---}) and marginal (\textcolor[HTML]{FF8000}{---}) prediction intervals for a linear regression derived from $\mathbb{E} \, [Y \, | \, X ]$, where $(X, \, Y) \sim \mathcal{N}_2 ( \boldsymbol{0}, \, \mathbf{\Sigma})$. Conditional prediction intervals are defined by the theoretical $95\%$-quantiles of $Y_o \mid X_o = x_o$. The marginal region is the set of points within an ellipse that contains $95\%$ of the probability mass of the bivariate normal distribution, i.e., $\mathcal{E}_{0.95} = \left\{ (x,y) \in \mathbb{R}^2 : (x,y)^\top \mathbf{\Sigma}^{-1} (x,y) \le \chi^2_{2,0.95} \right\},$ where $\chi^2_{2,0.95}$ is the $95\%$-quantile of the chi-squared distribution with 2 degrees of freedom.} 
   }
    \label{fig:marg_cond_int_comp}
    \vspace{-20 pt}
\end{figure}
\par
The distinction between marginal and conditional prediction intervals is important. Marginal coverage is based on probability computations that depend on the {\em joint} distribution of $Y$ and $\X$. That is, for a given prediction interval $C(\x) = [c_1(\x), \, c_2(\x)]$, where $c_1, \, c_2: \mathcal{X} \to \mathbb{R}$ are functions of $\x$, marginal prediction intervals compute coverage by treating the argument of both $c_1 (\cdot)$ and $c_2 (\cdot)$ as random rather than fixed, and often 
\begin{equation}
    \underbrace{\mathbf{Pr} ( Y_o \in  C (\x_o)  \, | \, \X_o = \x_o )}_{\mathrm{conditional}} \neq \underbrace{\mathbf{Pr} ( Y_o \in  C (\X_o) )}_{\mathrm{marginal}}.
\end{equation}
\par
In other words, marginal prediction intervals do not guarantee conditional coverage and tend to concentrate in high-density regions for predictor variables~\citep{dcp}. In some cases, marginal prediction intervals for predictor points in low probability regions might be undefined. For example, Figure~\ref{fig:marg_cond_int_comp} highlights that marginal prediction intervals defined by an ellipse are {\em limited} to high-density predictor regions: $|X_o | \leq \sigma_X \, \chi_{2,0.95}$,
$$C(X_o) = \rho X_o / \sigma_X \pm \sigma_Y \, [(1 - \rho^2) (\chi^2_{2,0.95} - X_o^2 / \sigma_X^2)]^{1/2},$$
where $(X, \, Y) \sim \mathcal{N}_2 ( \boldsymbol{0}, \, \mathbf{\Sigma})$ and $\chi_{2,0.95}$ is the square root of the $95\%$-quantile of the chi-squared distribution with 2 degrees of freedom.
%
\par
In contrast, conditional coverage treats the argument of both $c_1 (\cdot)$ and $c_2 (\cdot)$ as fixed and does not depend on the likelihood of observing a given predictor value. This allows us to compute intervals for points in low-density predictor regions. In summary, marginal prediction intervals provide population-wise guarantees: on average, coverage is only valid for the entire population, whereas conditional prediction intervals provide individual-wise guarantees: coverage is valid for a targeted individual. Further, conditional coverage implies optimal marginal coverage~\citep{dist_free_conf_inference}. This distinction is paramount in practice, not just theoretically. For example, in critical care, a central objective is to predict outcomes of specific patients, which motivates the development of prognostic scoring systems based on biomarker predictors~\citep{sofa_scores}. Such systems classify patients as high-risk if the pointwise predictions are larger or smaller than fixed thresholds. This highlights the importance of conditional prediction intervals for patient-specific risk assessment. 
\subsection{Related work}
\label{subsec: model-free and related work}
\cite{conf_book_1sted} introduce conformal prediction (CP), a flexible--yet powerful--method that exploits ranks of monotone transformations to attain coverage. \cite{conformal_nonparametric_local} show that there is no algorithm that computes finite length prediction intervals with valid distribution-free conditional coverage for finite samples. \cite{dist_free_conf_inference} propose elegant split CP (SCP) methods for distribution-free prediction intervals in regression. Both CP and SCP intervals provide marginal rather than conditional coverage. \cite{barber2020limitsdistributionfreeconditionalpredictive} discuss the difficulty of attaining conditional coverage. \cite{class_images_21} propose SCP methods that provide marginal coverage for image classifiers. This is not an exhaustive list of the growing literature in conformal methods. See~\cite{conf_intro_FTML_23} for an introduction. \cite{dcp} propose distributional CP (DCP), which is a resourceful method for computing conditional prediction intervals by exploiting the probability integral transform suggested by~\cite{politis2013model}.
\par
\cite{politis2015model} propose ingenious model-free bootstrap (MFB) methods for computing prediction intervals; see, e.g.,~\cite{ts_pred_inference}. \cite{conf_mfb_boot} show that MFB confidence intervals for the mean are bootstrap consistent if $d = 1$. \cite{wang2021modelfreebootstrapconformalprediction} show that MFB  prediction intervals are asymptotically conditionally valid. However, MFB methods do not scale as $d$ increases, require estimating $\hat{f}$ at every bootstrap iteration, and are hard to implement for complex models with computationally expensive training routines. Both MFB and DCP require consistent estimators of $P_{Y | \bX}$; unfortunately, even if an estimator is consistent, the convergence rates for $d \gg 1$~\citep{conformal_nonparametric_local} highlight the need for extremely large calibration sets. These methods neither calibrate nor exploit $\hat{f}$, nor do they provide model-agnostic coverage guarantees.
\par
An alternative approach is placing a Gaussian process prior on the unknown regression function~\citep{gaussian_process_bound_two}. However, this imposes stringent smoothness and translation invariance assumptions, the prediction intervals require refitting the posterior distribution for every new test observation (i.e., computationally expensive), typically assume homoscedastic errors with known variances, and have a different notion of validity based on posterior probabilities derived from an assumed prior distribution; see~\cite{conformal_Vovk_tutorial} for an insightful discussion. Further, if the prior-likelihood is misspecified or not valid, then these tend to severely undercover in practice~\citep{dl_calb_quant}.
\begin{table}
    \centering
    \caption{
        {\small Comparison of prediction intervals. \textbf{Mod-Ag (Model-Agnostic):} calibrates a trained model to compute adaptive optimal length. \textbf{Cond:} valid asymptotic conditional coverage. \textbf{HD Pred:} scales with predictor dimension. \textbf{Shape-Free:} works for distribution-free heteroscedastic errors. \textbf{Class:} provides error bounds for class prediction probabilities.}
    }
    \begin{tabular}{lrrrrr}
         \small\bf{Method} & \small\bf{Mod-Ag} & \small\bf{Cond} & \small\bf{HD Pred} & \small\bf{Shape-Free} & \small\bf{Class}\\
         \hline
         \small MAPS~(Ours) & \color{blue}{\cmark} & \color{blue}{\cmark} & \color{blue}{\cmark} & \color{blue}{\cmark} & \color{blue}{\cmark} \\
         \small MFB~\citep{politis2015model} & \color{orange}{\xmark} & \color{blue}{\cmark} & \color{orange}{\xmark} & \color{blue}{\cmark} & \color{orange}{\xmark} \\
         \small SCP~\citep{dist_free_conf_inference} & \color{orange}{\xmark} & \color{orange}{\xmark} & \color{blue}{\cmark} & \color{orange}{\xmark} & \color{orange}{\xmark} \\
         \small DCP~\citep{dcp} & \color{orange}{\xmark} & \color{blue}{\cmark} & \color{orange}{\xmark} & \color{blue}{\cmark} & \color{orange}{\xmark} \\
         \hline
    \end{tabular}
    \vspace{-10 pt}
    \label{tab:method comparison}
\end{table}
\par
Our ideas relate to the classic work of~\cite{mz_regression} for evaluating the quality of economic forecasts, and~\cite{Platt1999} for computing class prediction probability estimates in binary classification. However, neither of them propose prediction intervals nor do they define a conditional model that links calibration responses to model predictions. A complete overview of confidence intervals is beyond the scope of this work; see, e.g.,~\cite{JMLR:v15:javanmard14a},~\cite{politis1999subsampling}, and~\cite{bootstrap}.
\vspace{-5 pt}
\subsection{Contributions}\label{sec:contributions}
The key contribution of our work is introducing a model-agnostic statistical framework that combines calibration techniques with well-understood nonparametric estimators for attaining theoretically valid model-agnostic conditional coverage. At the core of our framework lies the lifted predictive model (LPM), which leverages the conditional relationship between responses and pointwise predictions rather than responses and high-dimensional predictors. We exploit the LPM to derive the MAPS algorithm, a computationally efficient and distribution-free procedure that takes as input any previously trained pointwise predictive model and outputs optimal prediction intervals with asymptotically valid conditional coverage. That is, MAPS adapts interval width to the pointwise accuracy of any trained predictive model without assuming a specific (non)parametric form for the regression function, homoscedastic or symmetric error distributions, or all of the former. Our LPM framework enables conditional uncertainty quantification in varied settings, which facilitates generalising MAPS to binary classification. We demonstrate finite-sample performance by conducting simulation studies where the true regression function is known. We further apply MAPS in two disparate settings: neural point estimation within simulation-based inference, and image classification, as detailed below.
\vspace{5 pt}
\newline
{\bf Neural Point Estimators:}  
Neural Bayes estimators (NBE) are neural networks trained with synthetic data to infer the parameters of an intractable statistical model. This approach for inference is becoming increasingly popular when likelihood estimation is unavailable or computationally prohibitive, but simulating from the model is fast~\citep{matt_american_statistician,Sainsbury-Dale03072025, leonte2025simulation}. Although NBEs are attractive for their amortisation, inference speed, and ability to process inputs with missing or irregular observations, their statistical properties, e.g., bias and coverage guarantees, are often poor~\citep{Andre2025NeuralBE}. MAPS addresses these issues by debiasing the point estimators, and crucially, is the first method that, given the estimators, implements full uncertainty quantification across all quantiles
; see Figure~\ref{fig: sbi figure} in Section~\ref{subsec:maps_for_sbi} for a preview.
\vspace{5 pt}
\newline
\noindent
{\bf Image Classification:} Neural networks have become ubiquitous in computer vision for sensitive tasks: medical diagnostics~\citep{CHEN_CNN_review_medicine}, defect identification in manufacturing~\citep{CNN_defect_detection} and gun detection in public spaces~\citep{cnn_gun_detection}. However, accurate and scalable uncertainty quantification for predictive probabilities remains challenging~\citep{prob_calib_temp_scaling, ICLR2025_conf_impact}. MAPS for binary classification can be thought of as a generalisation of Platt scaling \citep{Platt1999}, which, to the best of our knowledge, provides the first formal model-agnostic conditional coverage guarantees for class prediction probabilities in binary classification. Our tests on \texttt{ImageNet}~\citep{image_net} with the \texttt{ConvNeXt}~\citep{convnext} architecture reveal an intriguing {\em double uncertainty} phenomenon: MAPS suggests \texttt{ConvNeXt} 
assigns near-zero confidence to ambiguous images, effectively indicating ``it does not know". See Figure~\ref{fig: dog} in Section~\ref{subsec: application to dog images} for a preview.
%
\subsection{Paper outline}
\label{subsec: outline}
Section~\ref{sec: LPM} connects the regression function to optimal prediction intervals, discusses how these are closely related to the symmetric ones from classical statistical theory, and formally introduces the LPM framework. Section~\ref{sec: maps} introduces the MAPS algorithm, discusses efficient computational implementations, and generalises MAPS to binary classification problems. Section~\ref{sec: idealised properties} presents the asymptotic conditional validity of the MAPS algorithm, introduces theoretical tools for assessing how robust LPM estimators are to different calibration sets and the notion of contour-homoscedastic distributions, and derives the optimal properties of consistent predictive models. Section~\ref{sec: numerical exp} evaluates MAPS in simulation studies and Section~\ref{sec: application} illustrates its flexibility on the two applications mentioned in Section~\ref{sec:contributions}. Author contributions are detailed at the end of the paper. 
\addtocontents{toc}{\protect\setcounter{tocdepth}{0}{}}
\section{Model-agnostic prediction intervals}
\label{sec: LPM}
We assume that the response values are generated by the data-generating process
\begin{equation} \label{eq: regression error}
     Y = f(\boldsymbol{X}) + \epsilon,
\end{equation}
where the {\em random} errors $\epsilon \, | \, \bX = \bx$ have a continuous zero-mean distribution $P_{\epsilon | \bx}$ that can depend on the predictor $\bx$, i.e., we do not assume homoscesdastic or symmetric error distributions. 
By abuse of notation, we denote both the distribution (law) and cumulative distribution function of $\epsilon_o \mid \bX_o = \bx_o$ by $P_{\epsilon_o | \bx_o}$. If $f$ and $q_{\epsilon_o | \bx_o} ( \alpha) := P_{\epsilon_o | \bx_o}^{-1} (\alpha)$ are known, the ideal symmetric prediction intervals at the $(1 - \alpha) \times 100\%$ confidence level for an observation $\x_o$ would be
\begin{equation}\label{eq: ideal symm interval}
    C_{\mathrm{ideal}\text{-}\mathrm{sym}} \left ( \boldsymbol{x}_o \right ) := \big [ f(\boldsymbol{x}_o) + q_{\epsilon_o | \bx_o} ( \alpha / 2), \, f(\boldsymbol{x}_o) + q_{\epsilon_o | \bx_o} (1 -  \alpha / 2) \big ].
\end{equation}
\par
Most statistical theory suggests computing symmetric confidence intervals given model-based asymptotic results (e.g., maximum likelihood estimators). However, in practice, estimators need not be asymptotically normal or unbiased, and estimation errors need not have a symmetric distribution. If the errors have a skewed distribution, then $C_{\mathrm{ideal}\text{-}\mathrm{sym}} (\bx_o)$ is not optimal, i.e., there exist shorter prediction intervals that guarantee the same coverage. The shortest interval achieving a given coverage level is defined by the highest density region: the set of points where the density exceeds a threshold that depends on the given coverage level. For unimodal densities, this region is a single interval whose endpoints lie at equal density. Proposition~\ref{prop: ideal optimal interval} formalises this for prediction intervals. Specifically, if $P_{\epsilon_o | \bx_o}$ has a bounded absolutely continuous unimodal density function, that is, the density function $p_{\epsilon_o | \bx_o}$ is continuous and has a unique finite maximum, then the optimal prediction region is the interval $C_{\mathrm{ideal}} (\bx_o) = [c_1(\x_o), \, c_2(\x_o)]$, where $c_1 (\x_o)$ and $c_2 (\bx_o)$ satisfy
\begin{equation}\label{eq:HDR}
    p_{Y_o | \bx_o}(c_1 (\x_o)) = p_{Y_o | \bx_o}(c_2 (\x_o)) \text{ and } \mathbf{Pr} \big ( Y_o \in  [c_1 (\x_o),\, c_2 (\x_o)]  \, | \, \boldsymbol{X}_o = \boldsymbol{x}_o  \big ) = 1 - \alpha,
\end{equation}
$p_{Y_o | \x_o}$ is the conditional density function of $ Y_o | \X_o = \x_o$, and the endpoints can be written as adjusted symmetric quantiles:
\begin{equation}\label{eq: ideal oracle confidence region}
    c_1(\x_o) = f(\x_o) + q_{\epsilon_o | \x_o} ( \alpha / 2 + h) , \ \ \  c_2(\x_o) = f(\x_o) + q_{\epsilon_o | \x_o} (1 - \alpha / 2  + h) ,
\end{equation}
where $h \in [-\alpha / 2, \, \alpha/2]$ is a shape adjustment uniquely determined by~\eqref{eq:HDR}. 
For symmetric densities, $h = 0$, i.e., we recover $C_{\mathrm{ideal}\text{-}\mathrm{sym}} (\bx_o)$, $h < 0$ for left-skewed ones, and $h > 0$ for right-skewed ones. Our results hold for general continuous distributions, and extend naturally to multimodal ones via mixture arguments (the optimal set is a union of intervals). However, we limit our discussion to unimodal densities for ease of notation; in this case, the optimal prediction interval is unique.
\begin{proposition}\label{prop: ideal optimal interval}
    The ideal prediction interval $C_\mathrm{ideal} (\bx_o)$ solves the optimisation problem
    \begin{equation}\label{eq: optmial pred interval optimisation}
         C_{\mathrm{ideal}} \left ( \boldsymbol{x}_o \right ) = \underset{C ( \bx_o) }{\operatorname{argmin}} \, \lambda_{\mathrm{Leb}} \left (   C \left ( \boldsymbol{x}_o \right ) \right ) \, \, \, s.t. \, \, \, \mathbf{Pr} \big ( Y_o \in  C (\boldsymbol{x}_o)  \, | \, \boldsymbol{X}_o = \boldsymbol{x}_o  \big ) \geq 1 - \alpha,
    \end{equation}
    for all $\alpha \in (0, 1)$, where $\bx_o \in \mathcal{X}$, $\lambda_{\mathrm{Leb}}$ is the Lebesgue measure and $C ( \bx_o)  \subset \mathbb{R}$ is a prediction interval. Further, if $p_{\epsilon_o | \bx_o}$ is unimodal, then the solution to~\eqref{eq: optmial pred interval optimisation} is unique. 
\end{proposition}
\par
See Appendix~\ref{app: proof of prop 1} for a proof of Proposition~\ref{prop: ideal optimal interval} and for a simple calculation that determines the adjustment $h$ and translated ``symmetric" quantiles. \cite{dcp} show an equivalent implicit result based on conformal scores and the joint density function $p_{Y, \X}$. Our result makes the dependence on $f (\bx_o)$ explicit. Put another way, estimation of the regression function is unavoidable for deriving consistent prediction intervals and requisite for computing optimal ones.
\par
Since neither $f$ nor the error distribution $P_{\epsilon \mid \bx_o}$ is known in practice, we extend the ideal construction to a generic predictive model $\hat{f}$ by centring the interval at $\hat{f}(\bx_o)$. The residuals associated with model $\hat{f}$ are
\begin{equation} \label{eq: prediction error}
    \hat{\epsilon}_o := Y_o - \hat{f} (\boldsymbol{x}_o ), 
\end{equation}
given $\boldsymbol{X}_o = \boldsymbol{x}_o$. If we had access to the quantiles of $\hat{\epsilon}_o$, then we could compute the oracle prediction intervals, which guarantee conditional coverage at the $(1 - \alpha) \times 100\%$ level,
\begin{align}
     \widehat{C}_{\mathrm{oracle}\text{-}\mathrm{sym}} \left ( \boldsymbol{x}_o \right ) & \defeq \big [ \hat{f} (\boldsymbol{x}_o) + q_{ \hat{\epsilon}_o | \bx_o} ( \alpha / 2), \, \hat{f} (\boldsymbol{x}_o) + q_{ \hat{\epsilon}_o | \bx_o} ( 1 - \alpha / 2) \big ]\label{eq: oracle sym pred interval},\\
      \widehat{C}_{\mathrm{oracle}} \left ( \boldsymbol{x}_o \right )  & \defeq \big [ \hat{f} (\boldsymbol{x}_o) + q_{ \hat{\epsilon}_o | \bx_o} ( \alpha / 2 + \hat{h}) , \, \hat{f} (\boldsymbol{x}_o) + q_{ \hat{\epsilon}_o | \bx_o} ( 1 - \alpha / 2 + \hat{h})  \big ]\label{eq: oracle pred interval}.
\end{align}
\par
Lemma~\ref{lem: oracle and ideal coverage} highlights that if $\hat{f}$ is accurate, then $\widehat{C}_{\mathrm{oracle}} ( \boldsymbol{x}_o )$ is asymptotically optimal, i.e., as $n_{\mathrm{train}} \to \infty$ the interval in~\eqref{eq: oracle pred interval} converges in probability to $C_\mathrm{ideal} ( \boldsymbol{x}_o )$ defined in~\eqref{eq: ideal oracle confidence region}. Recall the little-$o$ notation~\citep[Chapter 2]{asym_stats}: if $\mathbf{Pr} ( \| \hat{f} - f \|_{L_2} \leq \delta ) \to 1$ for all $\delta > 0$ as $n_{\mathrm{train}} \to \infty$, we write $ \| \hat{f} - f \|_{L_2} = o_{\mathbf{Pr}} (1)$, and say that $\hat{f}$ is an $L_2$-norm consistent estimator of $f$. See Appendix~\ref{app: proof of lemma one} for a proof of Lemma~\ref{lem: oracle and ideal coverage}. 
\begin{lemma}\label{lem: oracle and ideal coverage}
    Assume that $\| \hat{f} - f \|_{L_2} = o_{\mathbf{Pr}} (1)$ and that $P_{\epsilon_o | \bx_o}$ is continuous for all $\bx_o$. Then, for all $t \in \mathbb{R}$ we have that $P_{\hat{\epsilon}_o | \bx_o} (t) \to P_{\epsilon_o | \bx_o} (t)$, and for all $\bx_o \in \mathcal{X}$
    \begin{equation}\label{eq: oracle and ideal converge}
         \lambda_{\mathrm{Leb}} \left ( C_{\mathrm{ideal}} \left ( \boldsymbol{x}_o \right ) \Delta \widehat{C}_{\mathrm{oracle}} \left ( \boldsymbol{x}_o \right ) \right ) \overset{\mathbf{Pr}}{\longrightarrow} 0, \, \, \, \, \mathrm{where} \, \, A \Delta B := (A \cup B ) \setminus (A \cap B).
    \end{equation}
    Further, the symmetric intervals also converge in probability, that is, 
    \begin{align}\label{eq: symmetric oracle and ideal converge}
         \lambda_{\mathrm{Leb}} \left ( C_{\mathrm{ideal}\text{-}\mathrm{sym}} \left ( \boldsymbol{x}_o \right ) \Delta \widehat{C}_{\mathrm{oracle}\text{-}\mathrm{sym}} \left ( \boldsymbol{x}_o \right ) \right ) &\overset{\mathbf{Pr}}{\longrightarrow} 0.
    \end{align}
\end{lemma}
\par
Lemma~\ref{lem: oracle and ideal coverage} and~\eqref{eq: ideal oracle confidence region}--\eqref{eq: oracle pred interval} show that consistent {\em and} optimal conditional coverage requires accurate estimation of both the regression function and the distribution of $\hat{\epsilon}_o$, not of $\epsilon_o$. Two broad approaches exist for estimating the distribution of $\hat{\epsilon}_o$. At one end are methods based on parametric assumptions, e.g., pivotal quantities for linear regression with homoscedastic Gaussian errors~\citep{linear_models_book}. At the other end, there is a growing number of black-box approaches for uncertainty quantification~\citep{Brenowitz_2025}, which are difficult to analyse and validate, do not adapt to arbitrary parametric or algorithmic predictive models, and require complex training routines. Recently, adaptive ``model-free" uncertainty quantification methods have been proposed~\citep{politis2015model, conf_intro_FTML_23}. However, these do not exploit $\hat{f}$ and require estimating $P_{Y | \bX}$, which is infeasible for high-dimensional $\bx$ and constraints their use to $d \leq 3$~\citep{wang2021modelfreebootstrapconformalprediction} or to impose stringent linearity or sparsity assumptions 
 on $P_{Y | \bX}$~\citep{dcp}.
\par
To address these issues, we introduce the LPM and propose the MAPS algorithm. Our approach builds from the idea of studying the relationship between the scalar random variables $Y_o$ and $\hat{f} (\bX_o) \equiv \hat{f} (\bX_o; \, (\X, \, \Y)_{\mathrm{train}} )$ rather than the one between $Y_o$ and $\bX_o$. That is, we exploit the information encoded in $\hat{f}$ by conditioning on $\hat{f} (\bX_o)$ rather than on $\bX_o$. This allows us to derive distribution-free prediction intervals with valid asymptotic conditional coverage for $Y_o \mid \hat{f} (\bX_o)$ for any $\hat{f}$. Surprisingly, we also attain conditional coverage for $Y_o \mid \bX_o$, and under weaker conditions than homoscedasticity, both $Y_o \mid \hat{f} (\bX_o)$ and $Y_o \mid \bX_o$ have the same coverage guarantees, as detailed in Section~\ref{subsec: dist-free model assumption}.
\subsection{The lifted predictive model}\label{subsec: spline estimator LPM}
Recall that, in addition to the training data, we have access to an independent, “unseen” calibration set drawn from the same distribution as both the training data and out-of-sample observations, i.e., for $\calb{i} = 1, \dots, \calb{n}$ both $(\bX_{\calb{i}}, \, Y_{\calb{i}}) \sim P_{\bX, Y}$ and $(\bX_o, \, Y_o) \sim P_{\bX, Y}$ are {\em i.i.d.} random variables. Thus, the conditional distributions of $Y_{\calb{i}} \mid \hat{f} (\bX_{\calb{i}}) = \hat{f} (\bx_{\calb{i}})$ and $Y_o \mid \hat{f} (\bX_o) = \hat{f} (\bx_o)$ depend on $\hat{f} (\bx_{\calb{i}})$ and $\hat{f} (\bx_o)$ in identical functional forms (e.g., location, shape and scale). 
\par
Substantiated by the above, we generalise post-training calibration techniques, which adjust $\hat{f}$ to account for biases in the resulting predictions \citep{prob_calib_temp_scaling, dl_calb_quant}, into a formal statistical framework with theoretical guarantees. This lifted probability space connects $\hat{f}$ to the calibration set and out-of-sample data, and allows us to define a lifted regression function that debiases $\hat{f}$ and enables uncertainty quantification induced by $P_{\bX, Y}$. We begin by defining the lifted predictive model below.
\begin{definition}\label{def: lpm definition}
    For $(\calb{\mathbf{X}}, \, \calb{\boldsymbol{Y}})$ and $(\boldsymbol{X}_o, \, Y_o)$ the lifted predictive model {\em (LPM)} is given by
    \begin{equation} \label{eq: dist-free additive errors}
          Y_{\calb{i}} = y_{\psibias} (\bX_{\calb{i}}) + u_{\calb{i}}, \, \, \, \, and \, \, \, \, \, Y_o = y_{\psibias} (\bX_o) + u_o, 
    \end{equation}
where $y_{\psibias} ( \bx) := \psibias (\hat{f} (\bx) )$ is the conditional expectation $\mathbb{E} \, [ Y \, | \, \hat{f} (\bX) = \hat{f} (\bx) ]$, $\psi: \mathbb{R} \rightarrow \mathbb{R}$ is a continuous function with finite $L_2$-norm, and $u_{\calb{i}}$ and $u_o$ have zero-mean distributions that have the same conditional dependence on $\hat{f} (\bx_{\calb{i}})$ and $\hat{f} (\bx_o)$, that is, for $\calb{i} = 1, \dots, \calb{n}$
$$if \, \,  \hat{f} (\bx_{\calb{i}}) = \hat{f} (\bx_o), \, \, \, \text{then} \, \,\, P_{u | \hat{f} } (\cdot \mid \hat{f} (\bx_{\calb{i}}) ) \equiv P_{u | \hat{f} } (\cdot \mid \hat{f} (\bx_o) ).$$
We stress that for any fixed $\hat{f}$, the stochastic mechanisms from~\eqref{eq: regression error} and~\eqref{eq: dist-free additive errors} are equivalent, i.e., the data-generating processes of $Y_o$ given $\hat{f} (\bX_o)$ and $\calb{Y}$ given $\hat{f} (\calb{\bX})$ are the same.
\end{definition}
\par
Intuitively, if $\hat{f}$ generalises effectively, then we would expect $\hat{f}$ to produce accurate predictions and minimal adjustment by $\psi$, i.e., $y_{\psi} (\bx_o) \approx \hat{f} (\bx_o) \approx f(\bx_o)$. In contrast, if $\hat{f}$ provides little predictive information, then we would expect that $Y_o$ and $\hat{f} (\bx_o)$ are weakly correlated, and that $y_{\psi}$  is constant with respect to $\hat{f}$, i.e., $y_{\psi} (\bx_o) \approx f(\bx_o) \not\approx \hat{f} (\bx_o)$. In both cases, and for any $\hat{f}$ in between, $u_o$ accounts for the remaining random variation unexplained by $\hat{f}$, induced by $P_{\bX, \epsilon}$. The shorthand $\hat{f} (\X_o) = \hat{f} (\x_o)$ denotes the event $\X_o \in \hat{f}^{-1} (\hat{y}_o)$, where $\hat{f} (\x_o) = \hat{y}_o \in \mathbb{R}$ and $\hat{f}^{-1} (\hat{y}_o) \subset \mathcal{X}$ is the preimage of the point $\hat{f} (\x_o) = \hat{y}_o$. This also holds for data in the calibration set, so to ease notation, at times, we drop the calibration subscript. We illustrate the idea with two examples below.
%
{\small
\begin{example}[Ridge regression]
    Let $Y = \sum_{j = 1}^d \gamma_j X_j + \epsilon$, where $X_1 \equiv 1$ (intercept), $\boldsymbol{\gamma} \in \mathbb{R}^d$ are unknown coefficients, and $\epsilon \sim \mathcal{N} (0, \, \sigma_{\epsilon}^2)$ are independent of $\X$. Given a training set of size $\ntrain < d $, stack $\bx_i^T$ row-wise in the design matrix $\mathbf{X}$, where $i = 1, \dots, \ntrain$ and, for simplicity, assume that $\mathbf{X}^T \mathbf{X} = \mathbf{I_d}$. The ridge estimator is $\boldsymbol{\hat{\gamma}} = (\mathbf{X}^T \mathbf{X} + \lambda \mathbf{I_d} )^{-1} \, \mathbf{X}^T \boldsymbol{Y}_{\mathrm{train}}$ and the predictive model is $\hat{f} : \x \mapsto \boldsymbol{\hat{\gamma}}^T \x$, where $\boldsymbol{Y}_{\mathrm{train} } = (Y_1, \dots, Y_{\ntrain})^T$ and $\boldsymbol{\epsilon}_{\mathrm{train} } = (\epsilon_1, \dots, \epsilon_{\ntrain})^T$ and $\lambda > 0$ is a hyperparameter. Define $\psi : t \mapsto (1 + \lambda) t$, so for calibration and out-of-sample data,
    $$ \psi(\boldsymbol{\hat{\gamma}}^T \calb{\x}) = (1 + \lambda) \boldsymbol{\hat{\gamma}}^T \calb{\x}  \, \, \, \, \mathrm{and} \, \, \, \, \psi(\boldsymbol{\hat{\gamma}}^T \bx_o) = (1 + \lambda) \boldsymbol{\hat{\gamma}}^T \bx_o.$$
    Given $\boldsymbol{\hat{\gamma}}$, $\calb{\bX} = \calb{\x}$ and $\bX_o = \x_o$, we have that
    $$\calb{u} \mid \boldsymbol{\hat{\gamma}}^T \calb{\x} \sim \mathcal{N} (0, \, \sigma_{\epsilon}^2 (1 + (1 + \lambda)^2 \calb{\x}^T \calb{\x}) )  \, \, \, \, \mathrm{and} \, \, \, \, u_o \mid \boldsymbol{\hat{\gamma}}^T \x_o \sim \mathcal{N} (0, \, \sigma_{\epsilon}^2 (1 + (1 + \lambda)^2 \x_o^T \x_o) ),$$
    where $\calb{u} = \calb{Y} - \psi(\boldsymbol{\hat{\gamma}}^T \calb{\X})$ and $u_o = Y_o - \psi(\boldsymbol{\hat{\gamma}}^T \bX_o)$. Thus, the {\em LPM} is
    $$\calb{Y} = (1 + \lambda) \boldsymbol{\hat{\gamma}}^T \calb{\X} + \calb{u}, \, \, \, \, \mathrm{and} \, \, \, \, Y_o = (1 + \lambda) \boldsymbol{\hat{\gamma}}^T \bX_o + u_o.$$
    Denote the distribution function of $\mathcal{N} (0, \, 1)$ by $\Phi$. Then, given $\boldsymbol{\hat{\gamma}}^T \bX_o = \boldsymbol{\hat{\gamma}}^T \bx_o$, for any $t_{0} < t_1 \in \mathbb{R}$,
    $$\mathbf{Pr} ( t_0 \leq Y_o - \psi(\boldsymbol{\hat{\gamma}}^T \bx_o) \leq t_1) = \Phi \left ( \frac{t_1} {\sqrt{ \sigma_{\epsilon}^2 (1 + (1 + \lambda)^2 \x_o^T \x_o)} } \right) - \Phi \left ( \frac{t_0}{\sqrt{ \sigma_{\epsilon}^2 (1 + (1 + \lambda)^2 \x_o^T \x_o)}} \right).$$
\end{example}
\begin{example}[Smooth functions and MLPs] \label{exm: linear smoothers}
    Assume that $Y$ in~\eqref{eq: regression error} is generated by
    \begin{equation*}
        Y =  f (\bX) + \epsilon,
    \end{equation*}
    where $f: \mathbb{R}^d \rightarrow \mathbb{R}$ has a finite $L_2$-norm and we estimate $f$ with a multilayer perceptron (MLP), or any other architecture, on the training set. Let $\hat{f}_{m \ell p}$ denote the learned MLP. Then, the {\em LPM} is
    \begin{equation*}
        \calb{Y} = \psi ( \hat{f}_{m \ell p} (\calb{\bX} ) ) + \calb{u}, \, \, \, \, \mathrm{and} \, \, \, \, Y_o = \psi ( \hat{f}_{m \ell p} (\bX_o) ) + u_o,
    \end{equation*}
     where $\psi$ adjusts the, possibly biased, predictions and $u_o \sim P_{u | \hat{f}_{m \ell p}}$ accounts for unexplained variation.
\end{example}
}
\par
By Definition~\ref{def: lpm definition} the function $\psi$ minimises MSPE given $\hat{f} (\bX_o) = \hat{f} (\bx_o)$, that is,
\begin{equation}\label{eq: theoretical psi minimises}
    \psi = \underset{g \in \mathcal{H}}{\mathrm{argmin} } \{ \mathbb{E}[(Y_o - g (\hat{f} (\X_o)))^2 \mid \hat{f} (\bX_o) = \hat{f} (\bx_o) ],
\end{equation}
where $\mathcal{H}$ is a space of functions~\citep{semiparametric_reg_book} to which the identity function belongs, i.e., $\mathrm{id}(t) = t$ and $\mathrm{id} (\cdot) \in \mathcal{H}$. Thus, for any trained model $\hat{f}$ we have that
\begin{equation}\label{eq: mpse for psi and identity}
    \mathbb{E} \, [(Y_o - \psi (\hat{f} (\X_o)))^2 \mid \hat{f} (\bX_o) = \hat{f} (\bx_o) ] \leq \mathbb{E} \, [(Y_o - \hat{f} (\X_o))^2 \mid \hat{f} (\bX_o) = \hat{f} (\bx_o) ].
\end{equation}
\par
In view of~\eqref{eq: mpse for psi and identity}, we see that $\psi$ debiases $\hat{f} (\bx_o)$ to minimise~\eqref{eq: theoretical psi minimises}, and that $u_o$ in~\eqref{eq: dist-free additive errors} accounts for approximation and irreducible error. 
The LPM formally connects the lifted probability space induced by conditioning on $\hat{f}$ to the calibration set and out-of-sample data generated by $P_{\bX, Y}$. 
For practical purposes, choosing an adequate space of functions $\mathcal{H}$ in which to estimate $\psi$ is critical. 
Recent developments in supervised learning, for example, neural point estimators in statistics~\citep{matt_american_statistician, zammit_mangion_annual_review_2025}, necessitate flexible choices. 
Provided sufficient calibration data are available, flexible regression functions, such as spline estimators are well-justified. These have enough flexibility to correct systematic or local biases that cannot be reliably controlled when training neural predictors on batches via stochastic gradient descent, as we demonstrate empirically in Section~\ref{subsec:maps_for_sbi}. 
 \par
Let $\mathcal{W}^2 $ denote the space of absolutely continuous functions with square-integrable second derivatives. 
The smoothing spline estimator~\citep{wabba_splines} solves
\begin{equation}\label{eq: smooth spline criterion}
    \psibiashat = \underset{ g \in \mathcal{W}^2 }{\mathrm{argmin}} \left \{ \frac{1}{\calb{n}}\| \calb{\boldsymbol{Y}} -  \boldsymbol{\hat{y}}_{g} \|_2^2 + \lambda  \int ( g'' (\hat{f} (\bx) ) )^2 \, d P_{\bX} \right \},
\end{equation}
where $\boldsymbol{\hat{y}}_{g} = (  g(\hat{y}_{\calb{1}}), \dots, \, g (\hat{y}_{\calb{n}}) )^T$, $\hat{f} (\bx_{\calb{i}}) = \hat{y}_{\calb{i}}$, $\calb{i} = 1, \dots\, \calb{n}$ and $\lambda > 0$ is a hyperparameter that controls how strong the smoothness penalty is, which can be chosen by cross-validation, or by the generalised cross-validation (GCV) criterion. 
\par
If $\lambda = 0$, then $\psibiashat$ interpolates calibration responses, and if $\lambda = \infty$, then $\psibiashat$ recovers the least squares estimator for linear regression~\citep[Chapter 5]{statLearn}. The solution is given by generalised ridge coefficients:
    $\boldsymbol{\hat{\beta}} = ( \mathbf{S} (\hat{f})^T \mathbf{S}(\hat{f}) + \lambda \, \mathbf{\Omega} (\hat{f}) )^{-1} \mathbf{S} (\hat{f})^T \calb{\boldsymbol{Y}}$,
where $[\mathbf{S}(\hat{f})]_{i j} = s_{j} (\hat{f} (\bx_{\calb{i}}))$, $s_{j}$ is the $j$th basis function in a $\calb{n}$-dimensional basis for natural cubic splines, $[\mathbf{\Omega} (\hat{f})]_{i j} = \int  s_{i}'' (\hat{f} (\calb{x}))  s_{j}'' (\hat{f} (\bx_{\calb{}})) \, \mathrm{d} P_{\bX}$, and the lifted predictions are 
\begin{equation}\label{eq: spline basis expansion}
    \hat{y}_{\psi} (\bx_o) := \widehat{\psi} (\hat{f} (\bx_o) ) = \sum_{j = 1}^{\calb{n}} \hat{\beta}_{j} s_{j} ( \hat{f} (\bx_o)).
\end{equation}
\par
For computational efficiency, we can minimise~\eqref{eq: smooth spline criterion} using an $\calb{N}$-dimensional basis, $\calb{N} = o(\calb{n})$. These spline predictors have attractive properties; see Section~\ref{subsec: LPM theory}. However, $\psi$ can be estimated by alternative nonparametric methods, e.g., wavelets~\citep{guys_book}.
\addtocontents{toc}{\protect\setcounter{tocdepth}{0}{}}
\section{The model-agnostic prediction sets (MAPS) algorithm}
\label{sec: maps}
\label{subsec: real_valued_maps}
Given a lifted predictor $\psibiashat$, we can mimic the construction in Section~\ref{sec: LPM} and construct a lifted prediction interval for $Y_o$. If we could generate {\em i.i.d.} lifted residuals, i.e., sample $M$ points from the distribution of
\begin{equation} \label{eq: lifted pred error}
    \hat{u}_o := Y_o - \hat{y}_{\psibias} (\bx_o),
\end{equation}
then we could use the empirical quantiles of the sample $\{ \hat{u}_o^{(m)}  \}_{m=1}^M$ to estimate the desired ones. However, this requires access to the distribution of $\hat{u}_o$ given $ \hat{f} (\bX_o) = \hat{f} (\bx_o)$. One can circumvent this by estimating the conditional distribution using the calibration set. We propose estimating the conditional distribution function, say $P_{\hat{u} | \hat{f}}$, of $\hat{u}_o$ using kernel density estimation (KDE) techniques~\citep{dens_est_silverman}. This induces further estimation errors, which MAPS accounts for via bootstrapping~\citep{bootstrap}.
\begin{algorithm}[t]
    \caption{ {\bf MAPS} } \label{alg: MAPS add errors}
    \setlength{\abovedisplayskip}{6pt}
\setlength{\belowdisplayskip}{6pt}
\setlength{\abovedisplayshortskip}{4pt}
\setlength{\belowdisplayshortskip}{4pt}
    \begin{algorithmic}[1]
        \Require  $(\calb{\mathbf{X}}, \, \calb{\boldsymbol{Y}})$, $\boldsymbol{x}_o$, $\hat{f}$, miscoverage level $\alpha \in (0, \, 1)$
        \State {\small Apply a nonparametric procedure to estimate $\psi$ in~\eqref{eq: dist-free additive errors}, e.g., minimise~\eqref{eq: smooth spline criterion} to get $\psibiashat$}
        \State Compute $\hat{u}_{\calb{i}} = Y_{\calb{i}} - \hat{y}_{\psibias} (\bx_{\calb{i}})$ and the lifted prediction $\hat{y}_{\psibias} (\boldsymbol{x}_{o} )$
        \State Using nonparametric methods, estimate $P_{\hat{u} | \hat{f}}$ to get $\widehat{P}_{\hat{u} | \hat{f}}$, e.g., $\widehat{P}_{\hat{u} | \hat{f}}$ in~\eqref{eq: error CDF estimation}
        \For{$b = 1$ to $b = B$}
            \State Sample $V_o, \, V_i \sim \operatorname{Unif} (0, \, 1)$ for $i = 1, \dots, \calb{n}$, and generate pivotal residuals:
            $$\hat{u}_i^{*b} = \widehat{P}_{\calb{\hat{u}} | \hat{f}}^{-1} ( V_i \,| \, \hat{f} (\bx_{\calb{i}} ) ) \,\, \, \mathrm{and} \, \, \, \widehat{U}_o^{*b} = \widehat{P}_{\hat{u}_o | \hat{f}}^{-1} ( V_o \,| \, \hat{f} (\bx_o ) )$$
            \State Bootstrap calibration responses for $i = 1, \dots, \calb{n}$ and an out-of-sample one:
            $$Y_{i}^{*b} = \hat{y}_{\psibias} (\bx_{\calb{i}}) + \hat{u}_i^{*b} \,\, \, \mathrm{and} \, \, \,Y_o^{*b} = \hat{y}_{\psibias} (\boldsymbol{x}_{o} ) + \widehat{U}_o^{*b}$$
            \State Use $\boldsymbol{Y}^{*b} = (Y_{1}^{*b}, \dots, Y_{\calb{n}}^{*b})$ to bootstrap $\psibiashat$ and the out-of-sample prediction:
            $$\psibiashat^{*b} = \mathrm{argmin}_{\psi} \{ \mathcal{L}_{\ell p m} ( \boldsymbol{Y}^{*b}, \, \hat{f} ) \} \,\, \, \mathrm{and} \, \, \, \hat{y}_{\psibias}^{*b} (\boldsymbol{x}_{o} ) = \psibiashat^{*b} (\hat{f} (\bx_o)).$$
            \State Bootstrap the out-of-sample prediction error: 
            $$\hat{u}^{*b}_o = Y_o^{*b} - \hat{y}_{\psibias}^{*b} (\boldsymbol{x}_{o} ).$$
        \EndFor
    \end{algorithmic}
     $\mathbf{Return:} \, \hat{y}_{\psibias} (\boldsymbol{x}_{o} ), \, \{ \hat{u}^{*b}_o \}_{b=1}^B$ \Comment{Lifted prediction and residual bootstrap sample}
\end{algorithm}
\par
To this end, we compute the calibration set residuals and estimate the conditional distribution function of $\hat{u}_o$ given $\hat{f}(\bx_o)$ by
\begin{equation} \label{eq: error CDF estimation}
     \widehat{P}_{\hat{u} | \hat{f}} ( \hat{u}_o \, | \,  \hat{f} (\bx_o) ) = \frac{ \sum_{\calb{i} = 1}^{\calb{n}} k_{\hat{f}} \left ( [ \hat{f} (\bx_o) - \hat{f} (\bx_{\calb{i}}) ] / h_{\hat{f}} \right ) K_{u} \left ( [ \hat{u}_o - \hat{u}_{\calb{i}} ] / h_u \right )}{ h_{\hat{f}} \, 
        \sum_{\calb{i} = 1}^{\calb{n}} k_{\hat{f}} \left ( [ \hat{f} (\bx_o) - \hat{f} (\bx_{\calb{i}}) ] / h_{\hat{f}} \right )
    },
\end{equation}
where $k_{\hat{f}}$ is a symmetric univariate density function, $K_u$ is a continuous cumulative distribution function, and $h_{\hat{f}}, \, h_u > 0$ are bandwidths. The KDE estimator in~\eqref{eq: error CDF estimation}, or alternative choices, might need adjusting if $\hat{f} (\bx_o)$ is a {\em boundary} point; see~\cite{DasCDF_estimator}, i.e., $\hat{f} (\bx_o)$ is close to either $\min \{ \, \hat{f} (\bx_{\calb{i}}) \}$ or $\max \{ \,\hat{f} (\bx_{\calb{i}}) \}$.
\par
If~\eqref{eq: error CDF estimation} is a nonparametric mixture of Gaussian kernels, then the estimator provides a built-in mechanism for generating conditional samples. However, other estimators may not admit immediate sampling, e.g., quantile regression. In these cases, we could use the probability integral transform 
to generate response variables, as noted by~\cite{politis2015model}. We generalise this by introducing pivotal residuals, i.e., residuals with a known standard distribution, even if the errors are heteroscedastic. As $\calb{n}$ increases this could become computationally prohibitive. See Appendix~\ref{app: comp details} for general computational details and an implementation that allows for efficient sampling regardless of $\calb{n}$.
\begin{definition}\label{def: pivotal residual}
    The pivotal residual $V_o$ for an out-of-sample predictor $\bx_o$ is given by the {\em probability integral transform} 
    of the out-of-sample lifted residual $\hat{u}_o$ given $\hat{f} (\bX_o) = \hat{f}(\bx_o)$
    $$V_o := P_{\hat{u}_o | \hat{f}} (\hat{u}_o \, | \, \hat{f} (\bx_o) ).$$
\end{definition}
\par
For any $\hat{f}$, $V_o$ has a standard uniform distribution and is independent of $\hat{f} (\bX_o)$. Thus, it can be thought of as a {\em pivotal quantity}, and using a sample of $V_o \sim \mathrm{Unif} (0, \, 1)$, we can generate lifted residuals by inversion. Further, we can generate response values in~\eqref{eq: lifted pred error} by using the estimator $\widehat{P}_{\hat{u} | \hat{f}}$ rather than the unknown $P_{\hat{u}_o | \hat{f}}$, and then bootstrap these steps to account for estimation error. Algorithm~\ref{alg: MAPS add errors} presents the MAPS procedure in detail.
%
\par
The ``symmetric" MAPS prediction intervals are given by
\begin{equation} \label{eq: maps interval dist-free}
    \widehat{C}_{\mathrm{maps}\text{-}\mathrm{sym}} (\bx_o) := \left [ \hat{y}_{\psibias} (\boldsymbol{x}_{o} ) + \widehat{q}_{\hat{u}^{*b}_o | \hat{f}} ( \alpha / 2), \, \hat{y}_{\psibias} (\boldsymbol{x}_{o} ) + \widehat{q}_{\hat{u}^{*b}_o | \hat{f}} (1 - \alpha / 2) \right ], 
\end{equation}
where the estimated quantiles $\widehat{q}_{\hat{u}^{*b}_o | \hat{f}}(\cdot)$ are the empirical quantiles of the bootstrap sample $\{ \hat{u}^{*b}_o \}$ generated by Algorithm~\ref{alg: MAPS add errors}. 
\par
If desired, these can be adjusted to optimality by using the quick procedure in Appendix~\ref{app: proof of prop 1} to find the shape adjustment $\hat{h}^* \in [-\alpha/2, \, \alpha/2]$, translated ``symmetric" quantiles and the resulting optimal prediction interval
\vspace{-10 pt}
\begin{equation} \label{eq: maps optimal interval dist-free}
    \widehat{C}_{\mathrm{maps}} (\bx_o) := \left [ \hat{y}_{\psibias} (\boldsymbol{x}_{o} ) + \widehat{q}_{\hat{u}^{*b}_o | \hat{f}} ( \alpha / 2 + \hat{h}^*), \, \hat{y}_{\psibias} (\boldsymbol{x}_{o} ) + \widehat{q}_{\hat{u}^{*b}_o | \hat{f}} (1 - \alpha / 2 + \hat{h}^*) \right ],
\end{equation}
\vspace{-10 pt}
where $\widehat{C}_{\mathrm{maps}} (\bx_o) \equiv \widehat{C}_{\mathrm{maps}} (\hat{f} (\bx_o))$ depends on $\bx_o$ only through $\hat{f} (\bx_o)$ and $\hat{u}_o^*$.
\subsection{Generalised MAPS prediction intervals for binary classification}\label{subsec:MAPS_binary}
Let $Y \in \{ 0, \, 1 \}$ be a binary response and $\X \in \mathcal{X}$ a $d$-dimensional predictor, such that there exist two functions, 
$f: \mathcal{X} \to \mathbb{R}$ and a link function $\varphi: \mathbb{R} \rightarrow (0, \, 1)$, that map a point $\x$ to a conditional probability for $Y$, i.e., $\mathbf{Pr} (Y = 1 \, | \, \bX = \bx) = \varphi ( f (\bx) ) $. 
\par
In this setting, a predictive model (probabilistic classifier) $\hat{f} : \mathcal{X} \to \mathbb{R}$ estimates the 
discriminant function $f$. Since $Y_o \mid \X_o = \x_o \sim \operatorname{Bernoulli}[ \varphi( f(\bx_o)) ]$, a natural extension to~\eqref{eq: dist-free additive errors} would be to define $Y_o \mid \hat{f} (\X_o) = \hat{f} (\x_o) \sim \operatorname{Bernoulli}[ \varphi( y_{\psi} (\bx_o)) ]$, where the lifted regression function $y_{\psi}$ adjusts $\hat{f}$. However, the predicted ``logits" $\hat{f} (\x_o), \, \hat{f} (\calb{\x}) \in \mathbb{R}$ but the responses $\calb{Y}, \, Y_o \in \{0, \, 1\}$, hence, predictions could be vacuous, i.e., $\hat{y}_{\psi} (\x_o) \notin \{0, \, 1\}$.
\par
Motivated by optimisation algorithms that exploit a ``working" response~\citep[Chapter 5]{statLearn}, we propose a hierarchical model that generalises the working response idea, and enables us to perform prediction and uncertainty quantification. To this end, consider the generalised logistic LPM
\begin{align}
    Y_o \, &| \, \hat{f} (\bX_o) = \hat{f} (\x_o) \sim \operatorname{Bernoulli}[\varphi_{\psi} (\bx_o) ], \label{eq: bernoulli LPM} \\
    Z_o &= y_{\psi} (\bx_o) + \frac{Y_o - \varphi_{\psi} (\bx_o) }{\varphi_{\psi} (\bx_o)  \, \{ 1 - \varphi_{\psi} (\bx_o)\} }, \label{eq: logistic LPM}
\end{align}
where $Z_o \in \mathbb{R}$ are latent responses, $\varphi_{\psi} (\bx) := \varphi( y_{\psi} (\bx) )$, and we assume that~\eqref{eq: bernoulli LPM}--\eqref{eq: logistic LPM} also holds for calibration set pairs $(\hat{f} (\calb{\X}), \calb{Y})$ and latent responses $\calb{Z}$.
\par
By construction, $ \mathbb{E} \, [ Z_o \, | \,\hat{f} (\bX_o) = \hat{f} ( \bx_o) ] = y_{\psi} (\bx_o)$  and $\mathrm{var} (Z_o \, | \, \hat{f} (\bX_o) = \hat{f} ( \bx_o) ) = 1$, hence, $Z_o$ can be interpreted as a latent response, such that its conditional mean $y_{\psi} (\x_o)$ is the value of the lifted discriminant function evaluated at $\hat{f} (\x_o)$, and the errors in~\eqref{eq: dist-free additive errors} are standardised Bernoulli ``errors". The lifted logit $y_{\psi} (\x_o)$ minimises MSPE for predicting $Z_o$, which connects~\eqref{eq: logistic LPM} to~\eqref{eq: mpse for psi and identity}. So, to predict $Z_o$ in~\eqref{eq: logistic LPM}, we only need to estimate $y_{\psi} (\x_o)$. The hierarchical model in~\eqref{eq: bernoulli LPM}--\eqref{eq: logistic LPM} allows us to estimate $y_{\psi}$ in alternating steps, which recovers the IRLS Algorithm~\citep[Chapter 4][]{statLearn}. See Algorithm~\ref{alg: IRLS for lifted logistic LPM} in Appendix~\ref{app: comp details}. 
\newline
Under~\eqref{eq: bernoulli LPM}--\eqref{eq: logistic LPM}, the penalised negative log-likelihood (deviance loss), say $\mathcal{L}_{\ell p m} (\calb{\boldsymbol{y}} ; \, \varphi, \psi )$, for our spline estimator is
\begin{equation}\label{eq: lifted deviance}
     - \sum_{\calb{i} = 1}^{\calb{n}} Y_{\calb{i}} \operatorname{logit} \{ \varphi_{\psi} (\bx_{\calb{i}}) \} +
     \log \{ 1 - \varphi_{\psi} (\bx_{\calb{i}}) \} + \frac{\lambda}{2}  \int ( \psibias'' (\hat{f} (\bx) ) )^2 \, d P_{\bX},
\end{equation}
where $\lambda > 0$ can be chosen by cross-validation or GCV, as in~\eqref{eq: smooth spline criterion}. In view of~\eqref{eq: lifted deviance}, we choose the \texttt{sigmoid} function, $\varphi(t) = \exp(t) \, / (1 + \exp(t) )$, as the link function because its inverse is the \texttt{logit}. However, our procedure can be extended to any choice of link function. Given our choice of link function (\texttt{sigmoid}), the loss function in~\eqref{eq: lifted deviance} reduces to a penalised deviance loss~\citep[Chapter 5]{statLearn} for spline coefficients $\boldsymbol{\beta}$,
\begin{equation}\label{eq: logistic loss for beta}
    \mathcal{L}_{\ell p m} (\calb{\boldsymbol{Y}} ; \, \varphi, \, \boldsymbol{\beta}) = - \sum_{\calb{i} = 1}^{\calb{n}} Y_{\calb{i}} y_{\psi} (\bx_{\calb{i}}) +
     \log \{ 1 - \varphi_{\psi} (\bx_{\calb{i}}) \} + \frac{\lambda}{2} \, \boldsymbol{\beta}^T \mathbf{\Omega} (\hat{f}) \, \boldsymbol{\beta},
\end{equation}
where $\mathbf{S}(\hat{f})$ is the natural cubic spline basis matrix evaluated at the logits $\hat{f} (\bx_{\calb{i}})$, and the entries in $\mathbf{\Omega} (\hat{f})$ are the penalty terms in~\eqref{eq: lifted deviance}. Minimising~\eqref{eq: logistic loss for beta} can be done efficiently by Newton's method
; see Algorithm~\ref{alg: IRLS for lifted logistic LPM} in Appendix~\ref{app: comp details}. The second-order approximation induced by the Hessian of \eqref{eq: logistic loss for beta} requires solving the generalised weighted least squares (WLS) problem
\begin{align}\label{eq: logistic LPM ridge regression}
    \boldsymbol{\hat{\beta}} &= \underset{\boldsymbol{\beta}}{\operatorname{argmin}} \left \{ \big ( \calb{\boldsymbol{Z}} - \mathbf{S} (\hat{f}) \, \boldsymbol{\beta} )^T \, \mathbf{W} (\varphi_{\psi}) \, (\calb{\boldsymbol{Z}} - \mathbf{S} (\hat{f}) \, \boldsymbol{\beta} \big ) + \lambda \, \boldsymbol{\beta}^T \mathbf{\Omega} (\hat{f}) \, \boldsymbol{\beta} \right \} \notag \\
     &= ( \mathbf{S} (\hat{f})^T \mathbf{W} (\varphi_{\psi} ) \, \mathbf{S} (\hat{f}) + \lambda \, \mathbf{\Omega} (\hat{f}) )^{-1} \, \mathbf{S} (\hat{f})^T \mathbf{W} (\varphi_{\psi} ) \calb{\boldsymbol{Z}},
\end{align}
where $\calb{\boldsymbol{Z}} = (Z_{\calb{1}}, \dots, Z_{\calb{n}})^T$ are the latent responses in~\eqref{eq: logistic LPM}, and the weight matrix $[\mathbf{W} (\varphi_{\psi})]_{i i} = \varphi_{\psi} (\bx_{\calb{i}})  \, ( 1 - \varphi_{\psi} (\bx_{\calb{i}}) )$ standardises the Bernoulli ``errors" $Y_{\calb{i}} -  \varphi_{\psi} (\bx_{\calb{i}})$.
\par
The fitted predictive logits are $\hat{y}_{\psi} (\bx_o) = \sum_{j = 1}^{\calb{n}} \hat{\beta}_{j} s_{j} ( \hat{f} (\bx_o)),$ where $s_j$ is the $j$th spline in the basis expansion, c.f., with $\hat{y}_{\psi} (\x_o)$ in~\eqref{eq: spline basis expansion}. Given $\varphi_{\psi} (\bx_{\calb{i}})$ and $\calb{\boldsymbol{Z}}$, the closed-form solution in~\eqref{eq: logistic LPM ridge regression} allows us to quickly compute $\hat{y}_{\psi} (\x_o)$ and ``predict" $y_{\psi} (\x_o)$. Under the LPM in~\eqref{eq: bernoulli LPM}--\eqref{eq: logistic LPM}, the lifted logit $y_{\psi} (\x_o)$ is the optimal discriminant value between labels given the logit $\hat{f} (\bX_o) = \hat{f} ( \bx_o)$.
\par
To this end, let $\overline{y}_{\psi} (\bx_o) := \mathbb{E} \, [ \hat{y}_{\psi} (\bx_o) \mid \hat{f} (\bX_o) = \hat{f} ( \bx_o) ]$ and define the predictive root $R_o := \hat{y}_{\psi} (\x_o) - \overline{y}_{\psi} (\bx_o)$. We propose constructing a prediction interval for $\varphi (\hat{y}_{\psi} (\bx_o) + R_o) $, which accounts for lifted logit estimation error and irreducible uncertainty, i.e., the idealised value $\varphi_{\psi} (\bx_o)$. For example, even if $\varphi(\hat{y}_{\psi} (\bx_o)) = 0.85$, the prediction interval could contain $1/2$ if estimation errors are non-negligible. Put another way, at a given confidence level $\alpha \in (0, \, 1)$, if $1/2$ {\em does not} belong to the prediction interval for $\varphi (\hat{y}_{\psi} (\bx_o))$, then the classifier is ``confident" in its label predictions; on the other hand, if $1/2$ belongs to the prediction interval, then the classifier is not ``confident", even if $\varphi(\hat{y}_{\psi} (\bx_o)) \not\approx 1/2$; see Figure~\ref{fig: dog}.
\begin{algorithm}[t]
\caption{ {\bf MAPS} for binary classification}\label{alg: maps for binary class}
\setlength{\abovedisplayskip}{6pt}
\setlength{\belowdisplayskip}{6pt}
\setlength{\abovedisplayshortskip}{4pt}
\setlength{\belowdisplayshortskip}{4pt}
\begin{algorithmic}[1]
        \Require  $(\calb{\X}, \, \calb{\boldsymbol{Y}})$, $\boldsymbol{x}_o$, $\hat{f}$, link function $\varphi$, miscoverage level $\alpha \in (0, \, 1)$ 
        \State Estimate the logistic LPM in~\eqref{eq: logistic LPM}, e.g., minimise~\eqref{eq: lifted deviance} to get $\psibiashat$ 
        \State Compute the lifted predictive logit $\hat{y}_{\psi} (\bx_o) = \widehat{\psi} ( \hat{f} (\bx_o))$
        
        \For{$b = 1$ to $b = B$}
            \State For $i = 1, \dots, \calb{n}$, generate a bootstrap binary response:
                $$Y_{i}^{*b} \sim \operatorname{Bernoulli} \left [ \varphi \left (\hat{y}_{\psi} (\bx_{\calb{i}}) \right ) \right ]$$ 
            \State Use $\boldsymbol{Y}^{*b} = (Y_{1}^{*b}, \dots, Y_{\calb{n}}^{*b})$ to bootstrap $\psibiashat$, e.g., by minimising~\eqref{eq: lifted deviance}: 
                $$\psibiashat^{*b} = \mathrm{argmin}_{\psi} \{ \mathcal{L}_{\ell p m} ( \boldsymbol{Y}^{*b}; \, \varphi, \psi ) \}$$
            \State For $i = 1, \dots, \calb{n}$, generate $Z_{i}^{*b}$ by substituting $Y_{i}^{*b}$ and $\varphi (\hat{y}_{\psi}^{*b} (\bx_{\calb{i}}))$ into~\eqref{eq: logistic LPM}:
            \begin{equation*}
                Z_i^{*b} = \hat{y}_{\psi}^{*b} (\bx_{\calb{i}}) + \{ Y_i^{*b} - \varphi (\hat{y}_{\psi}^{*b} (\bx_{\calb{i}})) \} \, [\varphi (\hat{y}_{\psi}^{*b} (\bx_{\calb{i}}))  \, \{1 - \varphi (\hat{y}_{\psi}^{*b} (\bx_{\calb{i}})) \} ]^{-1}.
            \end{equation*} 
            \State Use $\boldsymbol{Z}^{*b} = (Z_1^{*b}, \dots, Z_{\calb{n}}^{*b})$ to bootstrap $\boldsymbol{\hat{\beta}}^{*b}$ and $\hat{y}_{\psi}^{*b} (\bx_o)$: \Comment{WLS problem in~\eqref{eq: logistic LPM ridge regression}}
             $$\boldsymbol{\hat{\beta}}^{*b} = \mathrm{argmin}_{\boldsymbol{\beta}} \{ \mathcal{L}_{\ell p m} ( \boldsymbol{Z}^{*b}; \, \hat{f}, \, \boldsymbol{\beta} ) \} \,\, \, \mathrm{and} \, \, \, \hat{y}_{\psibias}^{*b} (\boldsymbol{x}_{o} ) = \textstyle\sum_{j = 1}^{\calb{N}} \hat{\beta}_{j}^* s_{j} ( \hat{f} (\bx_o)). $$
            \State Bootstrap the out-of-sample predictive logit root: $$R_o^{*b} = \hat{y}_{\psi}^{*b} (\bx_o) - \hat{y}_{\psi} (\bx_o).$$
        \EndFor
    \end{algorithmic}
     $\mathbf{Return:} \, \hat{y}_{\psi} (\bx_o), \, \{R^{*b}_o\}_{b=1}^B$ \Comment{Lifted predictive logit and bootstrap root sample}
\end{algorithm}

\par
Given $\widehat{C} (\bx_o)$, if $\mathbf{Pr} ( \hat{y}_{\psi} (\bx_o) + R_o \in \widehat{C} (\bx_o) \, | \, \hat{f} (\bX_o) = \hat{f} ( \bx_o) ) = 1 - \alpha$, then, since $\varphi$ is {\em monotonic}, we can $\varphi$-transform the interval's endpoints, denoted by $\varphi_{\psi} ( \widehat{C} (\bx_o) ) \subseteq [0, \, 1]$, to obtain the prediction interval
\begin{equation}\label{eq: maps intervals for binary classification}
    \mathbf{Pr} \left (\varphi(\hat{y}_{\psi} (\bx_o) + R_o)  \in \varphi_{\psi} \left ( \widehat{C} (\bx_o) \right ) \, | \, \hat{f} (\bX_o) = \hat{f} ( \bx_o) \right ) = 1 - \alpha,
\end{equation} 
which quantifies how ``confident" $\hat{f}$ is in its label predictions.
\par
If $\hat{f}$ discriminates consistently between classes, then we would expect $R_o$ to collapse at zero. Put another way, the lifted logits $y_{\psi} (\x_o)$ are consistent estimators of $\overline{y}_{\psi} (\x_o)$. Unfortunately, we cannot ignore the lifted bias, $\overline{y}_{\psi} (\bx_o) - y_{\psi} (\bx_o)$, without imposing further assumptions; see Section~\ref{sec: idealised properties}. On the other hand, if $\hat{y}_{\psi} (\x_o)$ does not have a concentration point, then the classifier $\hat{f}$ cannot discriminate consistently (``confidently") between classes given $\x_o$, the logit $\hat{f} (\x_o)$ is uninformative.
\par
In practice, neither $\overline{y}_{\psi} (\x_o)$ nor the distribution of $R_o$ are known. To amend this, we propose a bootstrap-based algorithm that extends Algorithm~\ref{alg: MAPS add errors} to binary classification. We can bootstrap predictive roots as follows. Given a calibration set and classifier $\hat{f}$, we minimise~\eqref{eq: logistic loss for beta} to obtain the fitted $\hat{y}_{\psi}$, then we generate a ``new" bootstrap sample of calibration responses $Y_i^* \sim \operatorname{Bernoulli}[\varphi \big ( \widehat{\psi} ( \hat{f} (\bx_{\calb{i}})) \big ) ]$, for $i = 1, \ldots, \calb{n}$. Given $Y_i^*$, we bootstrap the LPM by minimising~\eqref{eq: lifted deviance} to get $\widehat{\psi}^*$, and generate $Z_i^*$ by substituting $Y_i^*$ and $\varphi \big ( \widehat{\psi}^* (\hat{f} (\x_{\calb{i}}) ) \big )$ into~\eqref{eq: logistic LPM} for $i = 1, \ldots, \calb{n}$. Given $Z_i^*$, we bootstrap a ``prediction" of $\overline{y}_{\psi} (\x_o)$ by substituting $\boldsymbol{Z}^* = (Z_1^*, \ldots, Z_{\calb{n}}^*)^T$ into~\eqref{eq: logistic LPM ridge regression} and solving for $\boldsymbol{\hat{\beta}}^*$, which gives the bootstrap lifted predictive logit $\hat{y}^*_{\psi} (\x_o) = \sum_{j = 1}^{\calb{N}} \hat{\beta}_{j}^* s_{j} ( \hat{f} (\bx_o))$, where $\calb{N}$ is the number of spline basis functions. Finally, the bootstrap root is $R_o^* := \hat{y}_{\psi}^* (\bx_o) - \hat{y}_{\psi} (\bx_o)$; see Algorithm~\ref{alg: maps for binary class}.
\begin{remark}\label{rem: hierarchichal bootstrap}
We can think of Algorithm~\ref{alg: maps for binary class} as a ``hierarchical" bootstrap, where $Y^*_i \mid  \hat{y}_{\psi} (\bx_{\calb{i}})$ is a ``conditional posterior" for sampling latent responses $\boldsymbol{Z}^* \, | \, \boldsymbol{Y}^* $, and subsequently generating ``predictions" $\hat{y}_{\psi}^* (\bx_o) \, | \, \boldsymbol{Z}^*$.
\end{remark}
\par
Combining~\eqref{eq: logistic LPM} and~\eqref{eq: maps intervals for binary classification} with Algorithm~\ref{alg: maps for binary class}, we estimate the ``symmetric" quantiles $\widehat{q}_{R_o^{*b}} (\alpha/2)$ and $\widehat{q}_{R_o^{*b}} (1 - \alpha/2)$ to get the resulting confidence intervals for prediction probabilities, and can compute the adjustment $\hat{h}^* \in [-\alpha/2, \, \alpha/2]$ to obtain the optimal ones, 
\begin{align}
    \widehat{C}_{\operatorname{maps}\text{-}\mathrm{sym} } (\bx_o; \, \varphi ) &= \left [ \varphi (\hat{y}_{\psi} (\boldsymbol{x}_{o} ) +\widehat{q}_{R_o^{*b}} (\alpha/2) ), \, \varphi (\hat{y}_{\psi} (\boldsymbol{x}_{o} ) +\widehat{q}_{R_o^{*b}} (1 - \alpha/2) ) \right ], \label{eq: MAPS binary class} \\
     \widehat{C}_{\operatorname{maps} } (\bx_o; \, \varphi ) &= \left [ \varphi (\hat{y}_{\psi} (\boldsymbol{x}_{o} ) +\widehat{q}_{R_o^{*b}} (\alpha/2 + \hat{h}^*) ), \, \varphi (\hat{y}_{\psi} (\boldsymbol{x}_{o} ) +\widehat{q}_{R_o^{*b}} (1 - \alpha/2 + \hat{h}^*) ) \right ]. \label{eq: MAPS binary class opt}
\end{align}
%
\par
The lifted prediction probabilities $\varphi( \hat{y}_{\psi} (\bx_o) )$ from Algorithm~\ref{alg: maps for binary class} reduce to well-known post-training calibration procedures when $\psi$ is restricted to simple parametric classes.  If \(\psi(t) \equiv t/T\) for \(T>0\), then the lifted probabilities recover temperature scaling~\citep{prob_calib_temp_scaling}, and if \(\psi(t) \equiv \beta_0 + \beta_1 t\), then these recover Platt scaling \citep{Platt1999}.
\par
More generally, choosing $\psi$ from a flexible family of functions (e.g., splines) extends standard binary-classification calibration techniques, while providing more expressive local corrections if needed, the same is true for regression tasks. Algorithms~\ref{alg: MAPS add errors} and~\ref{alg: maps for binary class} can also estimate one-sided intervals, which is hard to do with conformal scores, due to their dependence on functions of absolute values~\citep{wang2021modelfreebootstrapconformalprediction}, e.g., $|\calb{Y} - \hat{f}(\calb{\X})|$. See Appendix~\ref{app: cat and multiple splits} for MAPS extensions to ``studentised" lifted residuals, multiple splits and categorical predictors.

\addtocontents{toc}{\protect\setcounter{tocdepth}{0}{}}
\vspace{-10 pt}
\section{Asymptotic properties of the MAPS algorithm}
\label{sec: idealised properties}
Recall that in the distribution-free setting it is not possible to find algorithms that produce valid non-trivial finite-sample conditional prediction intervals~\citep{conformal_nonparametric_local}. 
So, we focus on valid asymptotic conditional coverage.
\begin{definition}[\citealp{conformal_nonparametric_local}]
    We say that $\widehat{C} (\bx_o)$ is asymptotically conditionally valid if for all possible $\bx_o \in \mathcal{X}$ and $\alpha \in (0, 1)$, we have that
    \begin{equation}\label{eq: asymp con valid def}
        \mathbf{Pr} \big ( Y_o \notin \widehat{C} (\bx_o ) \, | \, \bX_o = \bx_o \big ) = \alpha + o_{\mathbf{Pr}} (1).
    \end{equation}    
\end{definition}
\par
Before presenting our main model-agnostic results, we introduce the notion of an asymptotically linear LPM. This allows us to analyse lifted pointwise predictions as ``estimators" for different choices of loss functions. That is, given a lifted loss function, say $\mathcal{L}_{\ell pm}$, does $\hat{y}_{\psi}(\bx_o)$ have statistical guarantees as $\calb{n} \to \infty$, and are these pointwise predictions robust.
\begin{definition}\label{def: asymp linear lpm}
    We say that the {\em LPM} linked to $\mathcal{L}_{\ell pm}$ is asymptotically linear if there exists a function $\ell_{\psi_{\hat{f}}}$ and a nonvanishing sequence $a (\calb{n}) > 0$ such that as $\calb{n} \to \infty$, we have that
    \begin{equation}
         \{ \yhatpsi (\bx_o) - y_{\psibias} (\bx_o) \} = \frac{ 1 }{ \calb{n} } \, \sum_{\calb{i} = 1}^{\calb{n}} \ell_{\psi_{\hat{f}}} ( Y_{\calb{i}}, \, \bx_o, \, \bx_{\calb{i}}) + o_{\prob} (1 / \sqrt{a (\calb{n})} ),  \tag{C1}\label{eq: asymp: linear lpm}
    \end{equation}
     $\mathbb{E} \, [ \ell_{\psi_{\hat{f}}} (  Y_{\calb{i}}, \, \bx_o, \, \bx_{\calb{i}}) ] = 0$, $\sum_{\calb{i} = 1}^{\calb{n}} \mathbb{E} \, [\ell_{\psi_{\hat{f}}}^2 ( Y_{\calb{i}}, \, \bx_o, \, \bx_{\calb{i}}) ] / \calb{n}^2 = O (1/ a (\calb{n})) \to 0$ for all $\bx_o, \, \bx_{\calb{i}} \in \mathcal{X}$. We then call $\ell_{\psi_{\hat{f}}}$ a lifted prediction influence function.
\end{definition}
\par
Definition~\ref{def: asymp linear lpm} generalises the idea of an influence function~\citep{geer2000empirical} from parameter estimation to pointwise prediction. Intuitively, if our chosen $\psibiashat$ is pointwise consistent, then $\yhatpsi (\bx_o)$ should concentrate around $y_{\psibias} (\bx_o)$, and we would compute approximately equal pointwise predictions given different calibration sets.
\begin{lemma}\label{lem: stability of the lpm}
     Assume that the {\em LPM} in~\eqref{eq: dist-free additive errors} is valid, and that the lifted predictions $\yhatpsi (\bx_o)$ satisfy~\eqref{eq: asymp: linear lpm}. Then, for all $\bx_o, \,  \bx_{\calb{i}}$ and $\hat{f}$, as $\calb{n} \to \infty$ we have that
    \begin{align*}
        &\sqrt{a (\calb{n}) } \, \{ \yhatpsi (\bx_o) - y_{\psibias} (\bx_o) \} \overset{\boldsymbol{\mathrm{d}}}{\longrightarrow} \mathcal{N} (0, \, \sigma_{\hat{y}}^2 (\bx_o) ),
        &\sqrt{a (\calb{n})} \, | \mathbb{E}[ \yhatpsi (\bx_o) ] - y_{\psibias} (\bx_o) | = o_{\mathbf{Pr}}(1).
    \end{align*}
\end{lemma}
\par
We attain model-agnostic asymptotic conditional validity and pertinence by exploiting Lemma~\ref{lem: stability of the lpm}, generalising model-free bootstrap results of~\cite{conf_mfb_boot}, and thinking of $\hat{y}_{\psibias} (\bx_o)$ as an M-estimator~\citep{geer2000empirical}. Before stating Theorem~\ref{th: aymp validity of MAPS}, we mention our technical assumptions.
\begin{assumptions}\label{assum: A1} For all $\bx_o \in \mathcal{X}$ and given $\hat{f} (\bX_o) = \hat{f} (\bx_o)$ as $\calb{n} \to \infty$ we have that 
        \begin{align}
            \| \widehat{P}_{\hat{u} | \hat{f} } - P_{\hat{u} | \hat{f} } \|_{L_2}  \overset{\prob}{\longrightarrow} 0, \tag{A1}\label{eq: consistent dist estimator} \\
             \int \left (  \hat{y}_{\psi} (\bx) - \hat{y}_{\psi}^* (\bx) \right)^2 d P_{\bX} \overset{\prob}{\longrightarrow} 0, \tag{A2}\label{eq: consistent psi etimator}
        \end{align} 
        where $\hat{y}_{\psi}^* (\bx)$ are the bootstrapped lifted predictions.
\end{assumptions}
\par
\eqref{eq: consistent dist estimator} states that $\widehat{P}_{\hat{u} | \hat{f} }$ converges in $L_2$-norm to $P_{\hat{u} | \hat{f} }$ for every fixed $\hat{f} (\bx_o)$. This assumption can be changed to the stronger supremum-norm consistency; however, one can show this holds, except for sets of zero-probability, by using Markov's inequality and Egorov's theorem~\citep{bartle1982elements}, as noted by~\cite{dist_free_conf_inference}. The kernel smoother in~\eqref{eq: error CDF estimation} is $L_2$-norm consistent, but the convergence rates depend on $d$, e.g., the rate for simpler marginal estimators~\citep{conformal_nonparametric_local} is 
\begin{equation}\label{eq: kernel smoother rates}
    \| \widehat{P}_{\hat{u}} - P_{\hat{u} } \|_{L_2} = O_{\prob} \left ( \calb{n}^{ - 4 / (d + 4)} \right ).
\end{equation}
\par
An attractive property of the LPM is that we have to estimate a conditional distribution function when conditioning on a {\em one-dimensional} variable, $\hat{f} (\bx_o)$, rather than on the original $d$-dimensional predictor. This enables us to efficiently estimate conditional prediction intervals for high-dimensional predictors, as evidenced by avoiding $d \gg 1$ in~\eqref{eq: kernel smoother rates}, e.g., conditional prediction intervals for multilayer perceptrons.
\par
\eqref{eq: consistent psi etimator} says that given the calibration set the bootstrap estimates converge in $L_2$-norm to the lifted predictions. Neither~\eqref{eq: consistent dist estimator} nor~\eqref{eq: consistent psi etimator} require accurate predictive models, i.e., if $\hat{f}$ is inaccurate, then $ \hat{y}_{\psi} (\x) \approx \calb{\overline{Y}}$. Further, both~\eqref{eq: consistent dist estimator} and~\eqref{eq: consistent psi etimator} are satisfied by a large number of nonparametric estimators, which usually perform better in one-dimensional problems~\citep{wasser_nonparam}. Both conditions are equivalent to the ones of~\cite{dcp, dist_free_conf_inference, wang2021modelfreebootstrapconformalprediction}. In Appendix~\ref{app: proof lemma two}, we show that condition~\eqref{eq: asymp: linear lpm} is sufficient for~\eqref{eq: consistent psi etimator}.
\begin{theorem}
\label{th: aymp validity of MAPS}
     Assume that~\eqref{eq: dist-free additive errors} is valid, and that~\eqref{eq: consistent dist estimator} and~\eqref{eq: consistent psi etimator} hold. Then, $\widehat{C}_{\mathrm{maps}} (\bx_o)$ given by~\eqref{eq: maps interval dist-free} is model-agnostic asymptotically conditionally valid, that is,
    \begin{equation} \label{eq: maps asymp validity}
        \underset{\bx_o}{\sup} \left \{ \mathbf{Pr} \left ( Y_o \notin \widehat{C}_{\mathrm{maps}} (\bx_o ) \, | \, \hat{f} (\bX_o) = \hat{f} (\bx_o) \right ) - \alpha \right \} = o_{\mathbf{Pr}}(1).
    \end{equation}
    Further, if $\hat{y}_{\psi} (\bx_o)$ is asymptotically linear, as in~\eqref{eq: asymp: linear lpm}, then for all predictions $\hat{f} (\bx_o)$
    \begin{equation} \label{eq: pertinent model-agnostic}
          \| P_{\hat{u} | \hat{f} }^* -  P_{u | \hat{f} } \|_{\infty} = o_{\mathbf{Pr}} \left ( \max \{1 / \sqrt{a_{\calb{n}}}, \, b_{\calb{n}} \} \right),
    \end{equation} 
    where $ P_{\hat{u} | \hat{f} }^*$ is the distribution of $\hat{u}^*_o$, and $b_{\calb{n}} \to 0$ is the asymptotic rate in~\eqref{eq: consistent dist estimator}.
\end{theorem}
\par
Theorem~\ref{th: aymp validity of MAPS} shows that under commonly accepted regularity conditions MAPS (Algorithm~\ref{alg: MAPS add errors}) produces asymptotically valid model-agnostic conditional prediction intervals, and if the estimated LPM is asymptotically linear, then we also have that the prediction intervals are bootstrap consistent. In other words, these results tell us that the {\em conditional} coverage of both~\eqref{eq: maps interval dist-free} and~\eqref{eq: maps optimal interval dist-free} converges to $(1 - \alpha)$. If~\eqref{eq: asymp: linear lpm} holds, then these prediction intervals are model-agnostic pertinent, and hence, bootstrap consistent; see Appendix~\ref{supp: proof theorem two}. Pertinent model-agnostic prediction intervals generalise pertinent prediction intervals, which have been shown to have better small sample performance than conformal ones~\citep{wang2021modelfreebootstrapconformalprediction}. This suggests that pertinent properties could aid to control MSPE. In contrast, conformal prediction intervals are not pertinent, since these are not concerned with MSPE, and tend to underperform for finite samples~\citep{ICLR2025_conf_impact}.
\begin{corollary}\label{cor: binary class interval}
    Assume the logistic {\em LPM} in~\eqref{eq: logistic LPM} and~\eqref{eq: consistent psi etimator} are valid. Then, $ \widehat{C}_{\operatorname{maps} } (\bx_o; \, \varphi )$ is asymptotically conditionally valid for $\varphi (\hat{y}_{\psi} (\bx_o) )$. Further, if $\hat{y}_{\psi} (\x_o) = \overline{y}_{\psi} (\x_o) + o_{\mathbf{Pr}}(1)$, then~\eqref{eq: MAPS binary class opt} is also an asymptotic confidence interval for $\varphi (\overline{y}_{\psi} (\x_o) )$, and if~\eqref{eq: asymp: linear lpm} holds, then 
    \begin{equation} \label{eq: binary class maps asymp validity}
        \underset{\bx_o}{\sup} \left \{ \mathbf{Pr} \left ( \varphi (y_{\psi} (\bx_o) ) \notin \widehat{C}_{\operatorname{maps} } (\bx_o; \, \varphi ) \, | \, \hat{f} (\bX_o) = \hat{f} (\bx_o) \right ) - \alpha \right \} = o_{\mathbf{Pr}}(1).
    \end{equation}
\end{corollary}
\par
Corollary~\ref{cor: binary class interval} suggests assessing how reliable label predictions are by computing ``prediction" intervals for $\varphi (\hat{y}_{\psi} (\bx_o) )$, which concentrate at the value $\varphi (\overline{y}_{\psi} (\bx_o) )$ if $\hat{y}_{\psi} (\x_o)$ converges in probability to $\overline{y}_{\psi} (\bx_o)$. That is, if the model produces reliable predictions for $\x_o$, otherwise the lifted predictive probability does not discriminate between a coin toss and surer cases. Further, if~\eqref{eq: asymp: linear lpm} holds, then~\eqref{eq: MAPS binary class} contains the ``true" prediction probability at the $(1 - \alpha) \times 100\%$ confidence level {\em and} collapses to the singleton $\{\varphi_{\psi} (\bx_o) \}$ in the limit, as $\calb{n} \to \infty$. See Appendix~\ref{supp: proof theorem two} for proofs of Corollary~\ref{cor: binary class interval} and Theorem~\ref{th: aymp validity of MAPS}.
%
\subsection{Distribution-free model assessment}\label{subsec: dist-free model assumption}
Consistent predictive models have optimal properties that suggest model-agnostic comparison statistics. Before stating them, we introduce $f$-contour homoscedastic distributions and study conditional coverage with respect to the regression function.
\begin{definition}\label{def: f-homoscedastic errors}
    We say that the errors in~\eqref{eq: regression error} are $f$-contour homoscedastic if for all $t \in \mathbb{R}$ and for each fixed $y \in \mathbb{R} \, \, s.t. \, \, f^{-1} (y) \neq \varnothing$ we have that 
    \begin{equation}
          \forall \bx \in f^{-1} (y), \, \, \, \, \,  \mathbf{Pr} ( \epsilon \leq t \mid f (\bX) = y) =  \mathbf{Pr} ( \epsilon \leq t \mid \bX = \bx) \tag{A3}\label{eq: f-homos assump},
    \end{equation}
     where $f^{-1} (y) := \{ \bx \in \mathcal{X} : \, f (\bx) = y, \, \, y \in \mathbb{R} \}$ denotes the preimage of a fixed point.
\end{definition}
\par
Definition~\ref{def: f-homoscedastic errors} says that the error distribution $P_{\epsilon | \bx} $ depends on the pointwise predictions $f(\bx)$ rather than on the predictor points $\bx$, i.e., error variation is identical for points that produce equal conditional means. This can be thought of as being in-between homoscedastic and heteroscedastic errors, say a ``sparse" heteroscedastic distribution.
\begin{lemma}\label{lem: coverage comparison and optimal f-homos}
    Assume that~\eqref{eq: regression error} is valid, denote the distribution of $\epsilon \mid f(\bX) = y$ by $P_{\epsilon | f}$, and define $C_{f\text{-}\mathrm{ideal}}  (\bx_o)$ analogously to $C_{\mathrm{ideal}} (\bx_o)$ in~\eqref{eq: ideal oracle confidence region} using the quantiles of $P_{\epsilon | f}$. Then, there exist $\alpha^{-}, \, \alpha^{+} \in (0, \, 1)$ such that
    \begin{equation}\label{eq: coverage comparison global}
        \alpha^{-} \leq \mathbf{Pr} ( Y_o \notin C_{f\text{-}\mathrm{ideal}}  (\bx_o) \mid \bX_o = \bx_o) \leq \alpha^{+},  \, \, \, \mathrm{and} \, \, \, \, \alpha^{-} \leq \alpha \leq \alpha^{+}.
    \end{equation}
    Further, if~\eqref{eq: f-homos assump} holds, then $\alpha^{-} = \alpha = \alpha^{+}$, and $C_{f\text{-}\mathrm{ideal}}  (\bx_o)$ is optimal, that is,
    \begin{equation}\label{eq: optimal interval for f-homos}
        \lambda_{\mathrm{Leb}} \left ( C_{f\text{-}\mathrm{ideal}}  (\bx_o) \Delta C_{\mathrm{ideal}} (\bx_o) \right ) = 0.
    \end{equation}
\end{lemma}
\par
We use the results in Lemma~\ref{lem: coverage comparison and optimal f-homos} to attain optimal prediction intervals for the LPM if the error distribution is $f$-contour homoscedastic and $\hat{f}$ is consistent.
\begin{theorem}\label{th: dist-free consistent model properties}
    Assume that~\eqref{eq: dist-free additive errors} is valid, $\| \hat{f} - f \|_{L_2} = o_{\mathbf{Pr}} (1)$, and that~\eqref{eq: consistent dist estimator} and~\eqref{eq: asymp: linear lpm} hold. Then, as $\calb{n} \to \infty$, for all $\bx_o \in \mathcal{X}$ we have that the {\em LPM} is also consistent, that is,
     \begin{equation}\label{eq: maps consistency}
         \| \yhatpsi - f \|_{\infty} = o_{\mathbf{Pr}} (1), \, \, \,  \psi(\hat{f} (\bx) ) = f (\bx) + o_{\mathbf{Pr}} (1),
     \end{equation}
     and for all $\x_o \in f^{-1} (y)$, the residuals $\hat{u}_o^*$ converge in distribution to $\epsilon_o \mid f(\X_o) = y$,
     \begin{equation}\label{eq: maps f-homos optimal}
        \| P_{\hat{u}_o | \hat{f} }^* -  P_{\epsilon_o | f } \|_{\infty} = o_{\mathbf{Pr}} (1). 
     \end{equation}
     \par
     Hence, if~\eqref{eq: f-homos assump} holds, then the intervals $\widehat{C}_{\mathrm{maps}} (\bx_o)$ are asymptotically optimal,
     \begin{equation}\label{eq: maps optimality}
         \lambda_{\mathrm{Leb}} \left ( \widehat{C}_{\mathrm{maps}} (\bx_o) \Delta  C_{\mathrm{ideal}} (\bx_o) \right ) \overset{\prob}{\longrightarrow} 0.
     \end{equation}
\end{theorem}
\par
Theorem~\ref{th: dist-free consistent model properties} shows that the LPM either ``inherits" accurate predictions or ``kills" them. MAPS uses the lifted predictions to adjust interval length and reflect larger uncertainty for more inaccurate models. If $Y$ is independent of $\bX$, then $f$ is a constant function. In this case, it is sensible to predict response values using the calibration set mean, i.e., $\hat{y}_{\mathrm{null}} (\bx_o)  := \calb{\overline{Y}}$. By Theorem~\ref{th: dist-free consistent model properties}, we see that as $\calb{n}$ tends to infinity the inequality
\begin{equation}\label{eq: mspe and interval length}
     \lambda_{\mathrm{Leb}} \Big ( C_{\mathrm{ideal}} (\bx_o) \Big ) \leq \lambda_{\mathrm{Leb}} \left ( \widehat{C}_{\mathrm{maps}} (\bx_o) \right ) \leq \lambda_{\mathrm{Leb}} \left ( \widehat{C}_{\mathrm{null}} (\bx_o) \right ),
\end{equation}
holds with high probability. This suggests comparing predictive models by looking at the length of $\widehat{C}_{\mathrm{maps}} (\bx_o)$ and at the difference $\yhatpsi (\bx_o) - \hat{f}(\bx_o)$. These comparison statistics extend to local or global ones, or to both of the former, by averaging them over $\bx \in \mathcal{X}$. The result in~\eqref{eq: mspe and interval length} coincides with experiments of~\cite{dist_free_conf_inference} for homoscedastic errors.
\subsection{Analysis of the spline estimator for the LPM}
\label{subsec: LPM theory}
The spline estimators in Section~\ref{subsec: spline estimator LPM} can be thought of as smoothing or penalised splines, which are computationally efficient~\citep{penalised_splines_consistency}. Before presenting their attractive theoretical properties, we mention our technical assumptions.
\begin{assumptions}\label{assum: spline assumptions}
    Let $\mathcal{W}^2 = \{ \, g:  [ y_{\min}, \,  y_{\max}] \to \mathbb{R}, \, \, \mathrm{s.t.} \, \, \int_{y_{\min}}^{y_{\max}} g''(y)^2 \, dy < \infty \}$ be the Sobolev space of order two, where $y_{\min}, \, y_{\max} \in \mathbb{R}$. The function $\psi$ belongs to the set $\mathcal{G} \subsetneq \mathcal{W}^2 $,
    \begin{equation}
        \mathcal{G} := \left \{ \, g \in \mathcal{W}^2 : \, \forall y \in [ y_{\min}, \, y_{\max}] \, \, \, \exists \beta_0, \, \beta_1 \in \mathbb{R} \, \, \mathrm{s.t.} \,  \, | g(y) - (\beta_0 + \beta_1 \, y) |  \leq  K_{\mathcal{G}} \right \}, \tag{S1}\label{eq: spline family assumption}
    \end{equation}
    where the constant $K_{\mathcal{G}} > 0$ \underline{does not} depend on $g \in \mathcal{G}$ or $y \in  [ y_{\min}, \,  y_{\max}]$. Further, for all $\bx \in \hat{f}^{-1} ([ y_{\min}, \, y_{\max} ])$ 
    we have that
    \begin{equation}
        0 < \textstyle\sum_{j = 1}^{d} (-1)^{j + 1} \left | \partial \hat{f} / \partial x_j \, (\bx) \right | < \infty, \, \, \, \, \mathrm{and} \, \, \, \, 0 < K_1 \leq p_{\bX} (\bx) \leq K_2 < \infty,\tag{S2}\label{eq: spline cond proper lifted support}
    \end{equation}
   where $p_{\X}$ is the continuous density function of $\X$ and $K_1, \, K_2$ are constants.
\end{assumptions}
\par
The function $\psi$ belongs to the family in~\eqref{eq: spline family assumption} if it concentrates around a linear function. That is, the family $\mathcal{G}$ can be thought of as the set of functions in $\mathcal{W}^2$ that do not oscillate too roughly around a linear function. Intuitively, either $\hat{f}$ is useful and aligns with the line of best fit or $\psi$ ``kills" its predictions, such that local bias adjustments do not become ``dominant". Recall that by Theorem~\ref{th: dist-free consistent model properties}, if $\hat{f}$ is consistent, then $\psi$ is the identity function. However, if $\hat{f}$ is highly inaccurate, then $\psi$ resembles a horizontal line, that is, $\psi$ is constant with respect to $\hat{f} (\bx_o)$. Our experiments in Section~\ref{sec: application} further suggest this is sensible.
\par
Assumption~\eqref{eq: spline cond proper lifted support} says that $\hat{f}$ has a countable number of critical points in the range defined by $y_{\min}$ and $y_{\max}$. Intuitively, both calibration and out-of-sample predictions fall within this range with non-zero probability. As a function $\hat{f}$ neither explodes nor vanishes, is bounded between a countable number of local minima or maxima, and the continuous density function of $\bX$ is bounded away from zero for all $\bx \in \hat{f}^{-1} ([ y_{\min}, \,  y_{\max}] )$. 
\begin{lemma}\label{lem: L2 spline lpm influence}
    Assume that~\eqref{eq: spline family assumption},~\eqref{eq: spline cond proper lifted support} and~\eqref{eq: dist-free additive errors} hold, and $0 < \sigma_{u}^2(\bX_o) := \mathbb{E} \, [u^2_o \, | \, \hat{f} (\bX_o) ] \leq M_u$ for all $\bX_o \in \hat{f}^{-1} ([ y_{\min}, \,  y_{\max}] )$, where $M_u < \infty$ is a constant. Then,~\eqref{eq: spline basis expansion} satisfies~\eqref{eq: asymp: linear lpm} with 
    \begin{equation}
        \ell_{\psi_{\hat{f}}} (Y_{\calb{i}}, \, \bX_o, \, \bX_{\calb{i}} ) = \frac{ \big (Y_{\calb{i}} - y_{\psi} (\bX_{\calb{i}}) \big )}{\sigma_{u} (\X_i) } \, \kernel{t}{\hat{f} (\X_{\calb{i}})},
    \end{equation}
    where $\kernel{\cdot}{\cdot}: [ y_{\min}, \,  y_{\max}] \times [ y_{\min}, \,  y_{\max}] \to \mathbb{R}$ is positive-definite reproducing kernel.
\end{lemma}
\par
\cite{silverman_kernel} obtained an approximation for $\kernel{\cdot}{\cdot}$, under a different setting and paved the way for future local asymptotic analyses~\citep{spline_asymp_subExp}. Lemma~\ref{lem: L2 spline lpm influence} is valid for any error distribution with a finite variance, not just sub-Exponential ones. Lemmas~\ref{lem: stability of the lpm}--\ref{lem: L2 spline lpm influence} combined with Theorems~\ref{th: aymp validity of MAPS}--\ref{th: dist-free consistent model properties} convey that the spline estimators in~\eqref{eq: spline basis expansion} can achieve conditional coverage given $\X_o$ for $f$-contour homoscedastic distributions. These results can be used to obtain bounds that become tighter the closer $P_{\epsilon_o | \bX_o}$ is to $P_{\epsilon_o | f}$.
\begin{theorem}\label{th: cond coverage spline}
    Assume~\eqref{eq: dist-free additive errors},~\eqref{eq: consistent dist estimator},~\eqref{eq: f-homos assump},~\eqref{eq: spline family assumption} and~\eqref{eq: spline cond proper lifted support} are valid. Then, we have that
    \begin{equation} \label{eq: maps asymp validity at xo}
        \textstyle\sup_{\x_o \in \hat{f}^{-1} ([ y_{\min}, \,  y_{\max}] ) } \left \{ \mathbf{Pr} \left ( Y_o \notin \widehat{C}_{\mathrm{maps}} (\bx_o ) \, | \, \bX_o = \bx_o \right ) - \alpha \right \} = o_{\mathbf{Pr}}(1), 
    \end{equation}
    and if $\| \hat{f} - f\|_{\infty} = o_{\prob} (1)$ in $\hat{f}^{-1} ([ y_{\min}, \,  y_{\max}] )$, then these are asymptotically optimal, 
    \begin{equation}\label{eq: maps spline optimal consistent}
        \textstyle\sup_{\x_o \in \hat{f}^{-1} ([ y_{\min}, \,  y_{\max}] ) } \left \{ \lambda_{\mathrm{Leb}} \left ( \widehat{C}_{\mathrm{maps}} (\bx_o) \Delta  C_{\mathrm{ideal}} (\bx_o) \right ) \right \} = o_{\mathbf{Pr}}(1).
    \end{equation}
\end{theorem}
\vspace{-2 pt}
\par
In the interest of robustness, we extend the LPM framework by considering the $L_1$-norm rather than the $L_2$-norm, via $\mathbb{E} \, [\mathbb{I} ( Y_o \leq y_{\psi} (\bX_o)  ) \, | \, \hat{f} (\bX_o) ] = 1 / 2$. 

\addtocontents{toc}{\protect\setcounter{tocdepth}{0}{}}
\section{Simulation study}
\label{sec: numerical exp}
We generate independent training and calibration sets of the same size from a known data-generating process and fit three different predictive models: a support vector machine~\citep[Chapter 12]{statLearn}, a generalised additive model~\citep{gam_book}, and a random forest~\citep{rf_breiman}. Using KDEs with Gaussian kernels in~\eqref{eq: error CDF estimation} and the spline estimators in~\eqref{eq: spline basis expansion}, we apply MAPS to derive conditional prediction intervals for each of them. That is, we apply Algorithm~\ref{alg: MAPS add errors} to the three fitted models: $\hat{f}_\textrm{SVM}, \, \hat{f}_\textrm{RF}$ and $ \hat{f}_{\textrm{GAM}}$, and compute the optimal prediction intervals in~\eqref{eq: maps optimal interval dist-free} that estimate the optimal ones in~\eqref{eq: optmial pred interval optimisation}. 
\par
The synthetic data are generated by
\begin{equation} \label{eq: add sim}
    Y = f(\bX) + \sigma_{\epsilon} (\bX) \, \epsilon,
\end{equation}
where $\epsilon \sim t_3$, $X_{j} \sim \mathrm{Unif} (0, \, 1)$ is independent of $\epsilon$ for $j = 1, \, 2, \, 3$, $\bX = (X_1, \, X_2, \, X_3)^T$, and $t_3$ denotes Student's $t$-distribution with three degrees of freedom. The regression function, $f (\bx) = f_1 (x_1) + f_2 (x_2) + f_3 (x_3)$, is the sum of \texttt{doppler}, \texttt{sinc} and \texttt{sin} terms:
\begin{align*}
    f_1 (x) &= 5 \left \{ x (1 - x) \right \}^{1/2}  \sin \left \{ (10 \pi) / (x + 0.05) \right \}, \\
    f_2 (x) &= \sin \left \{ 10 (x + 0.05 ) \right \} / (x + 0.05) , \\
    f_3 (x) &= \sin \left \{ 10 (x + 0.05 ) \right \}  (10 x + 0.05),
\end{align*}
which are spatially inhomogeneous and difficult to estimate~\citep{guys_book}, and the variance function is $\sigma_{\epsilon} (\bx) = 2 / 3 \, \{ (1 + 2 x_1) \, \cdot x_2 \, \cdot x_3 (1 - x_3) \}^{1/2}$. The errors in~\eqref{eq: add sim} are {\em heteroscedastic} and have the ``fattest" possible tails with a finite variance.
\begin{figure}
    \centering
\includegraphics[width=\linewidth]{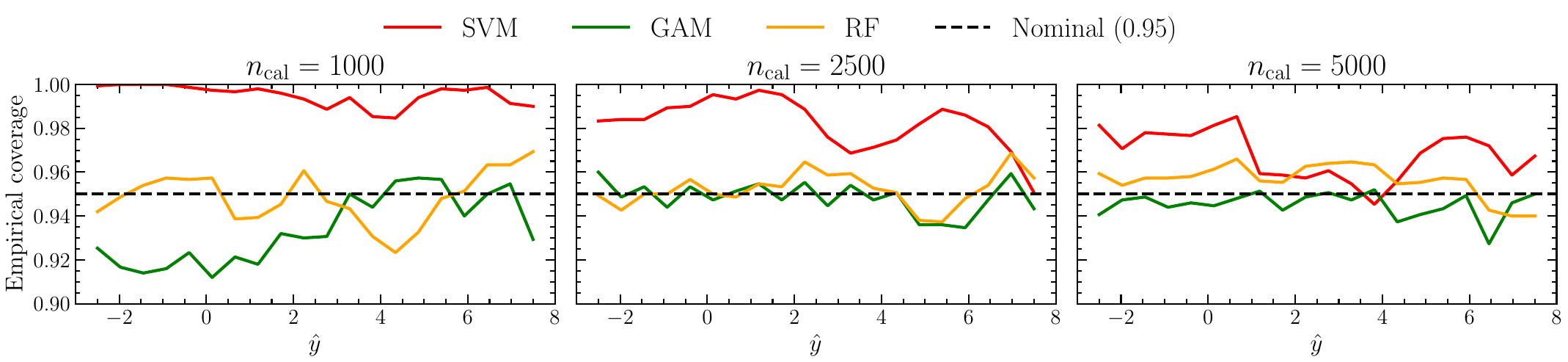}
\caption{
{\small Estimated conditional coverage of $\widehat{C}_{\mathrm{maps}} (\hat{y}_k)$ given by~\eqref{eq: cond coverage estimator}, where $\hat{f} (\bx_o) = \hat{y}_k $ and $ \hat{y}_k \in \texttt{linspace}(-2.5, \, 7.5, \, 0.5)$, $k = 1, \dots, 21$, and $\hat{f} \in \{ \hat{f}_\textrm{SVM}, \, \hat{f}_\textrm{GAM}, \, \hat{f}_\textrm{RF} \}$, the data are generated by~\eqref{eq: add sim} with $10,000$ out-of-sample test observations, and $1 - \alpha = 95\%.$ }
}
\label{fig: cond coverage estimation}
\vspace{-25 pt}
\end{figure}
\par
To validate our methodology, we evaluate finite-sample conditional coverage, given $\hat{f} (\bX_o) = \hat{f} (\bx_o)$, using a binning approach. Let $\{y_k\}_{k=1}^{21} \defeq \texttt{linspace}(-2.5,7.5,0.5)$ be the grid of evenly spaced points between $-2.5$ and $7.5$ with increment $0.5$. We choose this grid to fall in the interior of the overlap of the three models' prediction ranges, so that each conditioning value $\hat{y}_k$ is meaningful for all fitted models $\hat{f} \in \{\hat{f}_\textrm{SVM}, \hat{f}_\textrm{GAM}, \hat{f}_\textrm{RF}\}$, and so that estimation of $\psi$ and the conditional error distribution is done in regions with adequate empirical support. Exact conditioning on the 
events $\hat{f}(\X_o) = \hat{y}_k$ is infeasible in practice. We therefore generate new independent pairs $(X_{i_o}, \, Y_{i_o})$ from~\eqref{eq: add sim}, $i_o = 1, \dots, n_o$, and approximate these events by retaining only those realisations $\hat{f} (\bx_{i_o})$ for which $| \hat{y}_k - \hat{f} (\bx_{i_o})| \leq \texttt{tol}$, where we set $\texttt{tol} = 0.01$. 
We then fit the spline $\widehat{\psi}$ from~\eqref{eq: spline basis expansion} and estimate the conditional coverage of 
$\widehat{C}_{\mathrm{maps}} (\hat{y}_{k}) \equiv \widehat{C}_{\mathrm{maps}} (\x_o)$ in~\eqref{eq: maps optimal interval dist-free} with its empirical analogue:
\begin{equation}\label{eq: cond coverage estimator}
    \widehat{p} (\widehat{C}_{\mathrm{maps}} (\hat{y}_{k}) ) := \frac{\sum_{i_o =1}^{n_o} \mathbb{I} (Y_{i_o} \in \widehat{C}_{\mathrm{maps}} (\hat{y_k}) \, \, \,  \mathrm{and} \, \, \, \, | \hat{y}_k - \hat{f} (\bx_{i_o})| \leq 0.01) }{ \sum_{i_o =1}^{n_o} \mathbb{I} ( | \hat{y}_k - \hat{f} (\bx_{i_o})| \leq 0.01 )},
\end{equation}
where $\hat{f} (\bx_o) = \hat{y}_k$, $k=1,\ldots,21$. Figure 2 shows that MAPS achieves model-agnostic conditional coverage, which converges to the desired $95\%$ level as $\calb{n}$ increases and is robust for $\ntrain, \, \calb{n} \in \{ 1000, \, 2500, \, 5000\}$.
\par
\begin{figure}
    \centering
\includegraphics[width=\linewidth]{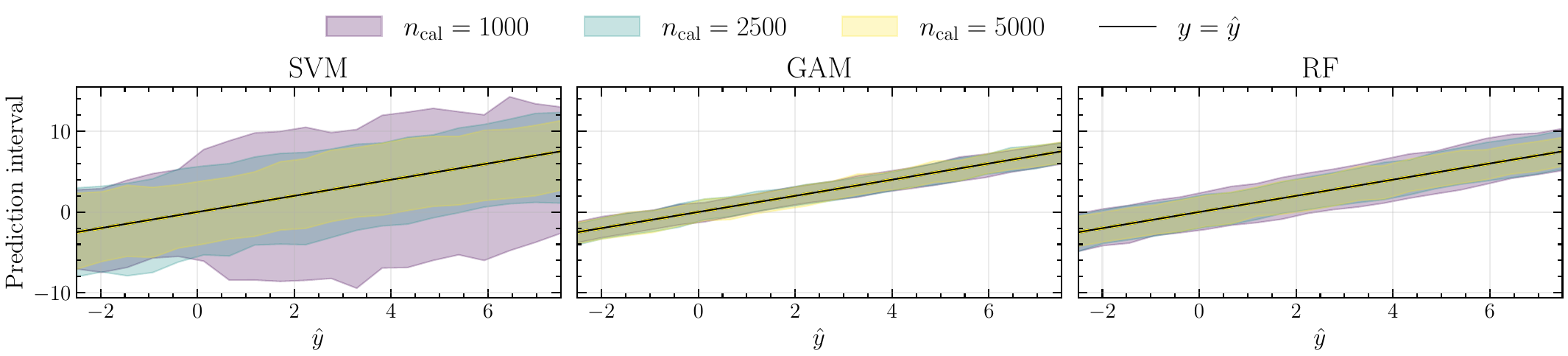}
\caption{{\small Estimated prediction intervals $\widehat{C}_{\mathrm{maps}} (\hat{y}_k)$ for $\hat{f} (\bx_o) = \hat{y}_k \in \texttt{linspace}(-2.5, \, 7.5, \, 0.5)$, $k = 1, \dots, 21$, and $\hat{f} \in \{ \hat{f}_\textrm{SVM}, \, \hat{f}_\textrm{GAM}, \, \hat{f}_\textrm{RF} \}$, the data are generated by~\eqref{eq: add sim} with $10,000$ out-of-sample test observations, $1 - \alpha = 95\%$, and $\ntrain, \, \calb{n} \in \{1000, \, 2500, \, 5000\}$. }
}
\label{fig: pred interval visualisation}
\vspace{-20 pt}
\end{figure}

Table~\ref{tab: experiment results optimal intervals} shows that MAPS provides {\em valid} conditional coverage for $Y_o \mid \hat{f} (\bX_o) = \hat{f} (\bx_o)$, even if the errors are heteroscedastic. For calibration set sizes as small as $n_\text{cal}= 1000$, the empirical conditional coverage $\widehat{p} (\widehat{C}_{\mathrm{maps}} (\hat{y}_k))$ lies between $91\%$ and $100\%$ (see Figure \ref{fig: cond coverage estimation}) 
across all grid-points 
$\hat{y}_k$, $k = 1, \dots, 21$, and for all fitted models $\hat{f} \in \{ \hat{f}_\textrm{SVM}, \, \hat{f}_\textrm{GAM}, \, \hat{f}_\textrm{RF} \}$. As the calibration set size increases, the estimated conditional coverage concentrates increasingly tightly around the nominal $95\%$ level, in line with the asymptotic guarantees of MAPS. 
Further, the calibrated predictor $\hat{y}_{\psi}$ is at least as accurate as the base $\hat{f}$ for all three models; see Table \ref{tab: experiment results optimal intervals}. 
Table~\ref{tab: experiment results optimal intervals} also shows that the largest improvements in MSPE are for $\hat{f}_{\mathrm{SVM}}$, which also produces the longest prediction intervals. In contrast, $\hat{f}_{\mathrm{GAM}}$ produces the shortest prediction intervals. As expected from Theorem~\ref{th: dist-free consistent model properties}, the former is the least accurate model, and the latter one is the most accurate one.
\begin{table}[t]
    \centering
    \caption{ {\small Simulation study results for MAPS applied to base estimators $\hat{f} \in \{ \hat{f}_\textrm{SVM}, \, \hat{f}_\textrm{GAM}, \, \hat{f}_\textrm{RF} \}$ on data generated by~\eqref{eq: add sim}.  {\bf MSPE}$(\mathbf{\hat{f}})$: 
    sample MSPE of the base predictor $\hat{f}$ computed on an independent test set drawn from~\eqref{eq: add sim}. 
    {\bf MSPE}$(\mathbf{\hat{y}_{\psi}})$: sample MSPE of $\hat{y}_{\psi}$ computed on the same test set as for $\hat{f}$. {\bf MSPE Diff}: percentage change in MSPE due to calibration by $\hat{\psi}$.
    {\bf MAE}$( 0.95, \, \widehat{p} (\widehat{C}_{\mathrm{maps}} (\hat{y}_k)) )$: mean-absolute error between the nominal $95\%$ level and the empirical conditional coverage estimate in~\eqref{eq: cond coverage estimator}. Conditional coverage $\hat{p}(\hat{C}_\textrm{maps}(\cdot))$ is evaluated at conditioning values $\hat{y}_k     \in \texttt{linspace}(-2.5, \, 7.5, \, 0.5)$. {\bf Length}: average length of the prediction intervals given by~\eqref{eq: maps optimal interval dist-free}. All values are reported to {\em one} decimal place.
    }
    }
\renewcommand{\arraystretch}{1.03}
\setlength{\tabcolsep}{5pt}

\begin{tabular}{lrrrrr}
$\mathbf{\ntrain, \, \calb{n}}$ 
& \small\bf{MSPE}$(\mathbf{\hat{f}})$ 
& \small\bf{MSPE}$(\mathbf{\hat{y}_{\psi}})$ 
& \small\bf{MSPE Diff}$ $ 
& \small\bf{MAE}$( 0.95, \widehat{p}(\hat{y}_k) )$ 
& \small\bf{Length}\\
\hline

\multicolumn{6}{l}{\small\bf Results for SVM: $\hat{f}_\textrm{SVM}$}\\[-2pt]
\cline{2-6}
\small 1,000 & 15.1 & 14.4 & $4.3\%$  & $4.5\%$ & 16.0 \\
\small 2,500 & 10.2 & 7.8  & $26.3\%$ & $3.2\%$ & 11.0 \\
\small 5,000 & 6.1  & 4.9  & $19.7\%$ & $1.7\%$ & 8.5 \\
\hline

\multicolumn{6}{l}{\small\bf Results for GAM: $\hat{f}_\textrm{GAM}$}\\[-2pt]
\cline{2-6}
\small 1,000 & 0.4 & 0.4 & $0.0\%$ & $1.8\%$ & 2.5 \\
\small 2,500 & 0.4 & 0.4 & $0.0\%$ & $0.6\%$ & 2.6 \\
\small 5,000 & 0.4 & 0.4 & $0.0\%$ & $0.4\%$ & 2.6 \\
\hline

\multicolumn{6}{l}{\small\bf Results for RF: $\hat{f}_\textrm{RF}$}\\[-2pt]
\cline{2-6}
\small 1,000 & 1.5 & 1.5 & $0.6\%$ & $1.0\%$ & 5.0 \\
\small 2,500 & 1.2 & 1.2 & $1.1\%$ & $0.6\%$ & 4.3 \\
\small 5,000 & 0.9 & 0.8 & $0.5\%$ & $0.8\%$ & 3.9 \\
\hline
\end{tabular}

    \label{tab: experiment results optimal intervals}
    \vspace{-2.5 pt}
\end{table}
Figure~\ref{fig: pred interval visualisation} shows the optimal prediction intervals in~\eqref{eq: maps optimal interval dist-free} account for heteroscedasticity and heavy-tailed skewed residuals, 
and that as $\calb{n}$ increases MAPS shortens interval length. Even if $\epsilon$ are homoscedastic Gaussian, neither $\hat{\epsilon}$ nor $u_o$ need be Gaussian too. Appendix~\ref{app: sim study details} provides an example where $f$ is a sparse ten-dimensional spatially inhomogeneous function, $\hat{f}$ is an MLP and $\epsilon \sim \mathcal{N}(0, \, \sigma_{\epsilon}^2)$.
%
\par
Our framework assumes access to sufficient calibration data. This is trivially satisfied in simulation-based inference, which motivates the application in Section~\ref{subsec:maps_for_sbi}, and is also increasingly met in practice, in modern large-scale datasets. However, an important consideration is how to choose $n_\text{cal}$ when data are scarce. This is a general open problem for data-splitting techniques~\citep{dist_free_conf_inference}. The LPM provides, to the best of our knowledge, the first framework to quantify the trade-off between training and calibration accuracy by studying how $\hat{f}$ and $\hat{y}_{\psi}$ relate under varying modelling assumptions on $f$ and $\epsilon$. This approach might enable us to attain non-asymptotic conditional coverage bounds, and as suggested by Lemma~\ref{lem: coverage comparison and optimal f-homos}, bound coverage for $Y_o \mid \X_o = \x_o$ in terms of the one for $Y_o \mid \hat{f} (\bX_o) = \hat{f} (\bx_o)$. We leave these investigations for future work.
\vspace{-10 pt}
\addtocontents{toc}{\protect\setcounter{tocdepth}{0}{}}
\section{Applications in regression and classification}
\label{sec: application}
We illustrate the versatility of MAPS through two evocative applications, one for each of the two core paradigms in supervised learning: regression and classification. The predictive models are neural networks trained on high-dimensional data in both cases. However, alternative predictive models and fitting procedures could equally be used.
\vspace{5 pt}
\newline
{\bf Regression:} In Section~\ref{subsec:maps_for_sbi}, we apply MAPS to neural point estimators within simulation-based inference (SBI), where it serves as an automated post-training calibration step that corrects residual biases and delivers uncertainty quantification without retraining the underlying model. This is particularly attractive for complex statistical models, where reliable uncertainty quantification is difficult to achieve with neural estimators \citep{Andre2025NeuralBE}.
\vspace{3 pt}
\newline
{\bf Classification:} In Section~\ref{subsec: application to dog images}, we turn to image classification and show that MAPS enables uncertainty quantification for predictive probabilities produced by modern deep classifiers. In a first, MAPS identifies ambiguous inputs and assigns low-confidence when the model is uncertain about the label, even if predicted probabilities are misleadingly high or low. This is a new interpretable notion of label uncertainty not captured by standard methods.
\subsection{MAPS for neural point estimators} 
\label{subsec:maps_for_sbi}
SBI is a class of methods in which estimators, often given by neural networks, are trained on large, synthetic datasets generated from complex statistical models in order to perform parameter inference. In this setting, one simulates parameter-observation pairs $(\bt,\x)$ from the statistical model of interest and trains a network to predict $\bt$ as a function of $\x$; see \cite{zammit_mangion_annual_review_2025} for a review. SBI is an ideal application setup for MAPS, as arbitrarily large
calibration datasets can be produced.
\par
{\bf Notation equivalence.} In this section, the response $Y$ is given by the scalar components of $\bt$, i.e., $Y \equiv \theta_j$, the estimator $\hat{f}(\x)$ by the corresponding components of $\hat{\bt}(\x)$, i.e., $\hat{f} (\x) \equiv \hat{\theta}_j (\x)$, and the lifted debiased means $\hat{y}_{\psi}$ by the components of the spline-calibrated $\hat{\bt}_\psi (\x)$, i.e., $\hat{y}_{\psi} \equiv \hat{\bt}_{\psi,j} (\x)$. 
If clear from context, we omit dependence of $\hat{\bt}$ and $\hat{\bt}_\psi$ on $\x$. 
\vspace{2.5 pt}
\par
Consider a random spatial field $\X = \left(X(\s_1),\ldots,X(\s_d)\right)$ observed at discrete locations $\s_1,\ldots,\s_d$ in the domain $\mathcal{D} = [0,1] \times [0,1] \subset \R^2$. Following \cite{Zhang2021ModelingSE}, we assume that $\X$ has a normal mean–variance mixture (NMVM) distribution 
 and focus on the generalised hyperbolic (GH) distribution subclass, which we describe next. The random variable $\X \in \R^d$ follows a NMVM distribution if we can write it as the sum $\X = \boldsymbol{\mu} + W \boldsymbol{\xi} + \sqrt{W} \boldsymbol{V}$, where $\boldsymbol{\mu}, \, \boldsymbol{\xi} \in \R^d$, $\boldsymbol{V} \sim \mathcal{N}(\boldsymbol{0}, \,\Sigma)$ is a multivariate Gaussian with covariance matrix $\Sigma$, and $W$ is a scalar random variable independent of $\boldsymbol{V}$. In particular, let $W$ follow the generalised inverse Gaussian (GIG) distribution, with density given by
\begin{equation*}
    p_{\mathrm{GIG}}(w; \, \zeta,\phi,\lambda) = \frac{(w/\phi)^{\lambda-1}}{2 \phi K_\lambda(\zeta)} \exp{\left(-\frac{\zeta}{2}\left(\frac{\phi}{w} + \frac{w}{\phi}\right)\right)}, \, w > 0,
\end{equation*}
where $K_\lambda$ is the modified Bessel function of the second kind of order $\lambda \in \R$, and $ \zeta, \, \phi > 0$. 
\noindent
To fully specify the model, set $\boldsymbol{\xi} = \xi \boldsymbol{1} = [\xi,\ldots,\xi] \in \R^d$ 
and the Mat\'ern covariance function
\begin{equation*}
    \mathrm{Cov}(\X(\s_i), \, \X(\s_j)) =   \frac{2^{1-\nu}}{\Gamma(\nu)} \left(\frac{|| s_i - s_j||}{\rho}\right)^\nu K_\nu\left(\frac{|| s_i - s_j||}{\rho}\right), \ \,  i,j = 1 ,\ldots, d, 
\end{equation*}
where $\Gamma$ is the Gamma function and $ \nu, \, \rho > 0$.
Following \cite{sainsbury2025neural}, assume $X$ is sampled at $d = 16 ^2$ locations $s_1,\ldots,s_d$ on an equidistant grid in $\mathcal{D}$. To ensure parameter identifiability, set $\phi =1$ and let each data-point $\x = (X^i(\s_1),\ldots,X^i(\s_d))_{i=1}^{150})$ be $150$ iid stacked realisations (replicates) of the spatial field $\X$, i.e., $\x$ is a $(150 \times 16^2)$-dimensional ``predictor" of $\bt$.
\par
The vector of parameters characterizing $\X$ is $\bt := (\xi,\zeta,\lambda,\rho,\nu)$. Even though this model admits a closed-form likelihood, evaluating it requires special functions and is difficult to scale. Even for univariate GH distributions, maximum likelihood estimation is challenging: the likelihood surface has flat regions, and optimisation algorithms are still an active area of research~\citep{10.5555/1195100, GH_tricky2}. This further motivates the use of neural networks for parameter inference. To this end, consider a dataset of simulated pairs $(\bt,\, \x)$, where the vector of parameters $\bt = (\theta_1, \ldots, \theta_5)$ is drawn from the priors: 
\begin{equation}
    \theta_1 \sim \mathcal{U}(-0.3,0.3), \, \theta_2 \sim \mathcal{U}(0.1,1), \, \theta_3 \sim \mathcal{U}(-1,1), \, \theta_4 \sim \mathcal{U}(0.05,0.35), \, \theta_5 \sim \mathcal{U}(0.5,2),
    \label{eq:supports}
\end{equation}
and $\x$ is then generated independently from the model specified by the value of $\bt$.
\begin{table}[t]
    \centering
    \caption{\small{Results for the SBI application described in Section~\ref{subsec:maps_for_sbi}. Summary statistics are defined equivalently to the ones reported in Table~\ref{tab: experiment results optimal intervals} for the simulation study in Section~\ref{sec: numerical exp}.  Due to the support of $\bt$ in~\eqref{eq:supports}, we use {\em two} decimal places to accurately report {\bf Length}.}
    }
    \label{tab: sbi results}
    \vspace{2pt}
    \resizebox{\linewidth}{!}{
    \begin{tabular}{lrrrrrrr}
    & \small\bf{MSPE}$(\hat{\theta}_j)$ & \small\bf{MSPE}$(\hat{\theta}_{\psi,j})$ & \small\bf{MSPE Diff} & \small\bf{MAE}$(95\%)$ & \small\bf{MAE}$(99\%)$ & \small\bf{Length}$(95\%)$ & \small\bf{Length}$(99\%)$ \\
    \hline
    $\theta_1$ & $1.3\mathrm{e}{-3}$ & $1.2\mathrm{e}{-3}$ & 4.3\% & 0.5\% & 0.1\% & 0.14 & 0.20 \\
    $\theta_2$ & $6.9\mathrm{e}{-3}$ & $6.7\mathrm{e}{-3}$ & 2.9\% & 0.5\% & 0.2\% & 0.30 & 0.38 \\
    $\theta_3$ & $1.0\mathrm{e}{-2}$ & $1.0\mathrm{e}{-2}$ & 2.9\% & 1.4\% & 0.4\% & 0.41 & 0.54 \\
    $\theta_4$ & $2.0\mathrm{e}{-4}$ & $2.0\mathrm{e}{-4}$ & 12.9\% & 1.8\% & 0.4\% & 0.05 & 0.07 \\
    $\theta_5$ & $1.4\mathrm{e}{-3}$ & $1.2\mathrm{e}{-3}$ & 16.2\% & 1.7\% & 0.4\% & 0.15 & 0.21 \\
    \hline
    \end{tabular}
    }
    \vspace{-2.5pt}
\end{table}
\par
The choice of priors is common with SBI methods, and ensures the model can ``learn" to predict parameter values in the entirety of parameter space. 
In this setup, a neural network is trained to ``predict" $\bt$ as a function of $\x$ by minimising squared-error-loss. This procedure yields a point estimator $\hat{\bt}$ 
 of the posterior mean $\mathbb{E} \, [\bt \, | \, \x]$. 
  However, one can use alternative loss functions; see Section~\ref{sec:sbi_discussion}. We do not train these neural point estimators ourselves; rather, we apply MAPS to the ones in~\cite{sainsbury2025neural}. 
Details about the network architecture and training procedure can be found in~\cite{sainsbury2025neural}. We thank the authors for making their estimators available.
\par
By applying Algorithm~\ref{alg: MAPS add errors}, we calibrate (debias) the provided neural point estimators and perform component-wise, not joint, uncertainty quantification for
$\bt$ given $\hat{\bt}(\x) =  
\big ( \hat{\theta}_j (\x) \big )$, where $ 
j = 1, \dots, 5$; see Section \ref{sec:sbi_discussion} for extensions. 
Specifically, for each component $j=1,\ldots,5$, MAPS estimates the {\em univariate} calibration function $\psi_j$  in~\eqref{eq: theoretical psi minimises} and fits the calibrated estimator $\hat{\theta}_{\psi,j}(\x) = \hat{\psi}_j(\hat{\theta}_j(\x))$. The vector-valued debiased estimator is then
\begin{equation*}
\hat{\bt}_{\psi}(\x) = \big ( \hat{\theta}_{\psi,1}(\x), \, \hat{\theta}_{\psi,2}(\x), \, \hat{\theta}_{\psi,3}(\x), \, \hat{\theta}_{\psi,4}(\x), \, \hat{\theta}_{\psi,5}(\x) \big ). 
\end{equation*}
\par
\begin{figure}[t]
\centering
    \begin{subfigure}{\textwidth}
        \makebox[1.5em][l]{\small{(a)}}%
        \raisebox{-0.5\height}{\includegraphics[width=0.95\textwidth]{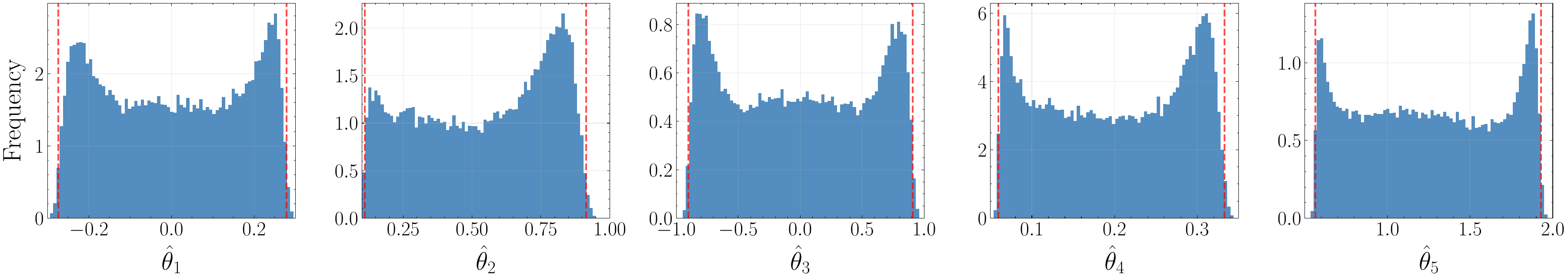}}
        \phantomcaption\label{figure:inf}
    \end{subfigure}
    \begin{subfigure}{\textwidth}
        \makebox[1.5em][l]{{\small (b)}}%
        \raisebox{-0.5\height}{\includegraphics[width=0.95\textwidth]{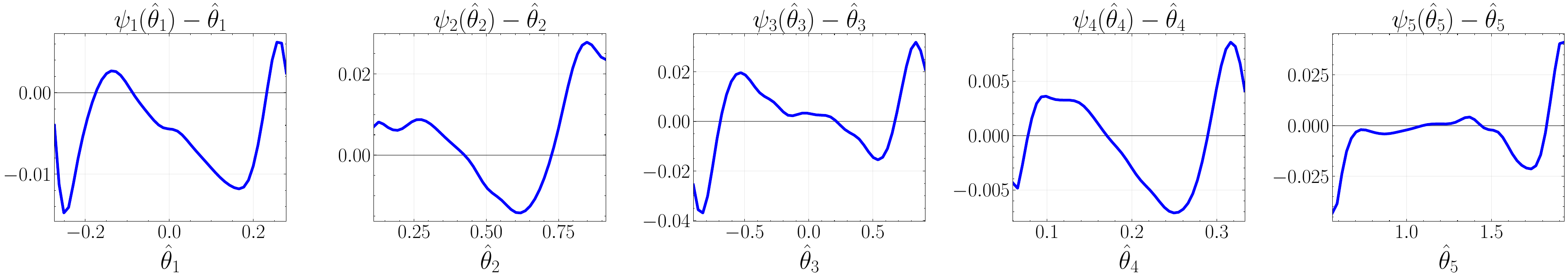}}
        \phantomcaption\label{fig:spline_minus_theta_hat}
    \end{subfigure}
    \begin{subfigure}{\textwidth}
        \makebox[1.5em][l]{{\small (c)}}%
        \raisebox{-0.5\height}{\includegraphics[width=0.95\textwidth]{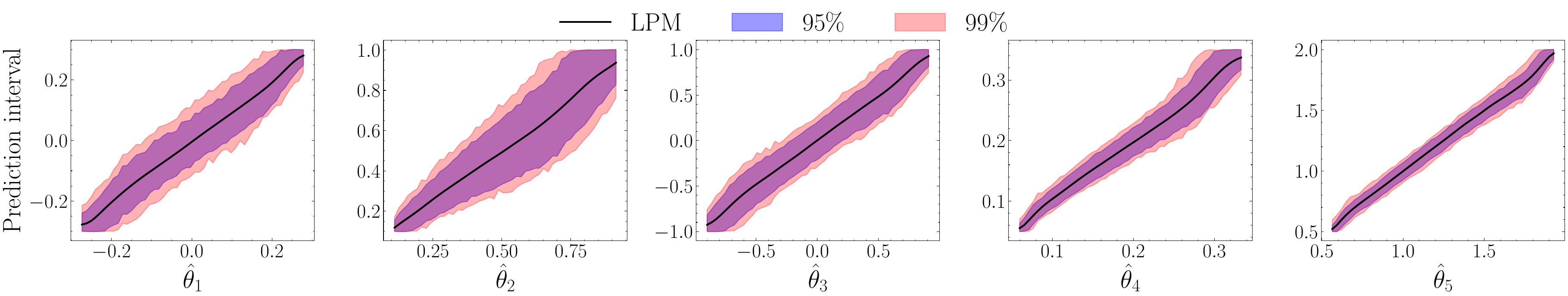}}
        \phantomcaption\label{fig:pred_intervals_sbi}
    \end{subfigure}
    \begin{subfigure}{\textwidth}
        \makebox[1.5em][l]{{\small (d)}}%
        \raisebox{-0.5\height}{\includegraphics[width=0.95\textwidth]{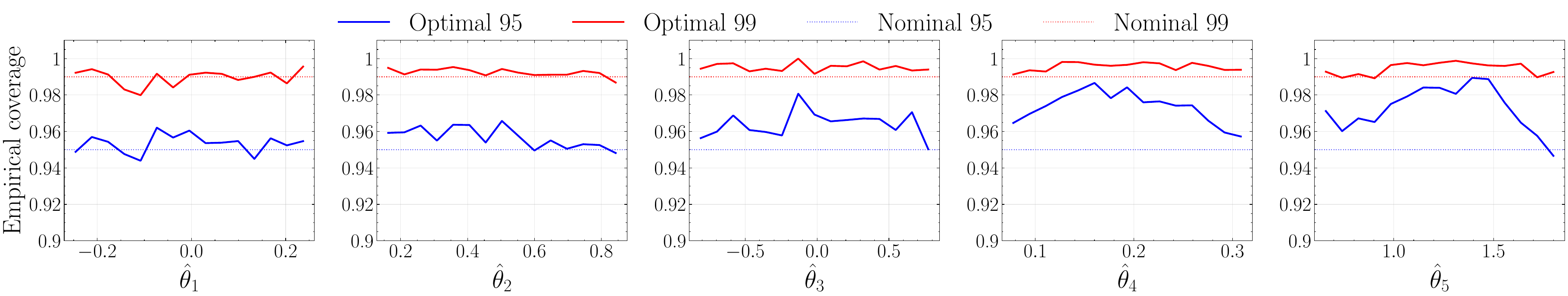}}
        \phantomcaption\label{fig:cond_coverage-sbi}
    \end{subfigure}
\caption{\small{Results for the application in SBI for parameter inference of spatial data under a NMVMN model~\citep{sainsbury2025neural} using a neural point estimator $\hat{\bt}_j$, where $j = 1, \dots, 5$. Fig.~\ref{figure:inf}: Histograms of $\hat{\bt}_j$, where the red lines highlight the $0.5\%$, $99.5\%$ empirical quantiles. Fig.~\ref{fig:spline_minus_theta_hat}: LPM bias adjustments $\hat{\psi}(\hat{\theta}_j) - \hat{\theta}_j$. Fig.~\ref{fig:pred_intervals_sbi}: $95\%$ and $99\%$ prediction intervals for $\theta_j \mid \hat{\theta}_j$. Fig.~\ref{fig:cond_coverage-sbi}: Empirical coverage of the intervals displayed in Fig.~\ref{fig:pred_intervals_sbi}. Fig.~\ref{fig:spline_minus_theta_hat}--\ref{fig:cond_coverage-sbi} are for $\hat{\bt}_j$ in the ranges highlighted by Fig.~\ref{figure:inf}. See Section~\ref{subsec:maps_for_sbi} for details.}}\label{fig: sbi figure}
\vspace{-20 pt}
\end{figure}

 
We start from a calibration set $(\bt_{i_\textrm{cal}},\x_{i_\textrm{cal}})_{i=1}^{n_\textrm{cal}}$ of size $n_\textrm{cal} = 5\cdot 10^4$, and for $i = 1, \ldots, n_\textrm{cal}$, evaluate the trained neural network at the values $\x_{i_\textrm{cal}}$ to compute the point estimators $\hat{\bt}_{i_\textrm{cal}}$. Figure \ref{figure:inf} displays histograms of the scalar components of $\hat{\bt}_{i_\textrm{cal}}$, where the $0.5\%$ and $99.5\%$ quantiles of each component are highlighted in red. To exclude possible boundary effects in calibration, we restrict our analysis to $\hat{\theta}_j$ in the highlighted ranges: $\hat{\theta}_1 \in (-0.27, \, 0.28), \, \hat{\theta}_2 \in (0.11, \, 0.91 ), \, \hat{\theta}_3 \in (-0.90, \, 0.91), \, \hat{\theta}_4 \in (0.06, \,0.33), \, \hat{\theta}_5 \in (0.56, \,1.92)$.
\par
If the neural network produced the optimal estimator: $\hat{\theta}_j(\x)  = \mathbb{E} \, [\theta_j \, | \, \x], \, j=1,\ldots,5$, then the fitted splines $\hat{\psi_j}$, estimated from $(\hat{\bt}_{i_\textrm{cal}}, \bt_{i_\textrm{cal}})_{i=1}^{n_\textrm{cal}}$, would coincide with the identity function $\textrm{id}(\cdot)$. Departures from optimality can thus be assessed by how much $\hat{\psi}_j(\cdot) -\mathrm{id}(\cdot)$ deviates from $0$, $j=1,\ldots,5$. Figure \ref{fig:spline_minus_theta_hat} corroborates significant deviations from zero. For example, at $\hat{\theta}_2 \approx 0.12$, the local adjustment $\hat{\theta}_{\psi,2} - \hat{\theta}_2 \approx 0.1$ corresponds to an almost $10\%$ correction. More generally, the corrections are largest in regions of high posterior mass for the scalar components of $\hat{\bt}$, indicating that MAPS improves accuracy where it matters most for inference. Formally, these gains are reflected in the \textbf{MSPE Diff} column in Table~\ref{tab: sbi results}. This demonstrates that the spline component of the lifted predictive model enables a post-training calibration step, correcting residual bias that is difficult to eliminate during neural network training. Importantly, this improvement is achieved without retraining the underlying neural estimator, which is computationally intensive.
\par
Figure~\ref{fig:pred_intervals_sbi} shows the optimal prediction intervals at the $95\%$ and $99\%$ levels for $\theta_j \mid \hat{\theta}_j$, where $j=1,\ldots,5$, Figure~\ref{fig:cond_coverage-sbi} displays their empirical conditional coverage, which closely match the nominal levels. Table \ref{tab: sbi results} also reports the average interval length and absolute error between the nominal and empirical levels, {\bf MAE}$(1 - \alpha) := \mathrm{mean} (| (1 - \alpha) - \hat{p} (\hat{\theta}_{j}(\x))|)$, where $\hat{p} (\hat{\theta}_{j})$ is the empirical coverage of the MAPS prediction intervals given $\hat{\theta}_{j}$ and  centred at $\hat{\theta}_{\psi,j}$, both at the $95\%$ and $99\%$ levels for each component in $\bt$. We use $15$ equidistant values for $\hat{\theta}_j$ in the ranges from Figure~\ref{figure:inf}. These results agree with the ones in Section~\ref{sec: idealised properties}, and illustrate that MAPS produces reliable prediction intervals for neural estimators.
\subsubsection{Discussion for MAPS in regression} \label{sec:sbi_discussion}
Fitting the deterministic part of the LPM has similarities with post-training calibration, which has been shown to improve estimator performance across numerous studies \citep{caruana2004predicting, Walchessen_2023_neural_likelihood_surfaces}. 
Further, the LPM can be interpreted as a debiasing, model-agnostic method applied to an available estimator---although in our case, this is only an intermediary step for instantiating the MAPS algorithm, rather than the sole objective, as it is in post-training calibration. While post-training calibration is most commonly studied in binary classification, to the best of our knowledge, our analysis represents the first application of such techniques to neural point estimators beyond classification, and in particular, within simulation-based inference.
\par
Based on the above results, this step should be applied automatically once a (neural) point estimator has been trained for quick uncertainty quantification and diagnostics, which can be used to define hypotheses tests. If the fitted spline deviates substantially from the identity, these could suggest that local biases persist after training. Importantly, such biases are made worse, but not solely caused by overfitting. Indeed, \cite{leonte2025simulation} showed that calibration can improve performance even when neural networks are trained with data simulated on the fly. Further, \cite{bai2021don} proved that even theoretically principled estimators that do not involve neural networks, e.g., the maximum likelihood estimator for logistic regression, exhibit the undesirable property of overconfidence. It is thus clear that these biases cannot, in general, be efficiently eliminated in the context of neural estimators through additional training or larger batch sizes, as they reflect structural properties of the estimator. Since fitting the LPM is computationally inexpensive, this adjustment can be incorporated as a routine post-training step in the SBI pipeline, and more generally, in neural network estimation. 
\par
Finally, in our experiments, we condition on scalar point estimates, although the method readily extends to vector-valued point estimates. The LPM framework also accommodates other loss functions, e.g., MAE instead of MSE, in which case the LPM targets the conditional median rather than the mean. While closed-form solutions are not available for \eqref{eq: smooth spline criterion} with MAE replacing MSE, the optimisation task is convex and can be efficiently solved with iterative methods. Crucially, estimating the lifted regression function $\psi$ on the calibration set does not require repeated evaluations of the trained neural network or fine-tuning it. 
\vspace{-5 pt}
\subsection{MAPS for binary image classification}
\label{subsec: application to dog images}
We test MAPS for binary classification (see Algorithm~\ref{alg: maps for binary class}) on ImageNet~\citep{image_net}. The goal is to quantify uncertainty for the \textit{dog} versus \textit{not dog} predictive probabilities produced by a modern deep classifier. We used a pretrained \texttt{ConvNeXt}~\citep{convnext} and converted the original 1000-way output into a binary score by aggregating the softmax probability mass over the ImageNet dog breed synsets, yielding $\mathbf{Pr}(\text{dog})$, which is then mapped to log-odds (logits). The calibration set is a subset of $n_{\mathrm{cal}}=2500$ images from the validation set, which is used to fit the logistic LPM in~\eqref{eq: logistic LPM} and compute the prediction intervals in~\eqref{eq: MAPS binary class} for a test set of images.
\par
\begin{figure}
    \centering
\includegraphics[width=\linewidth]{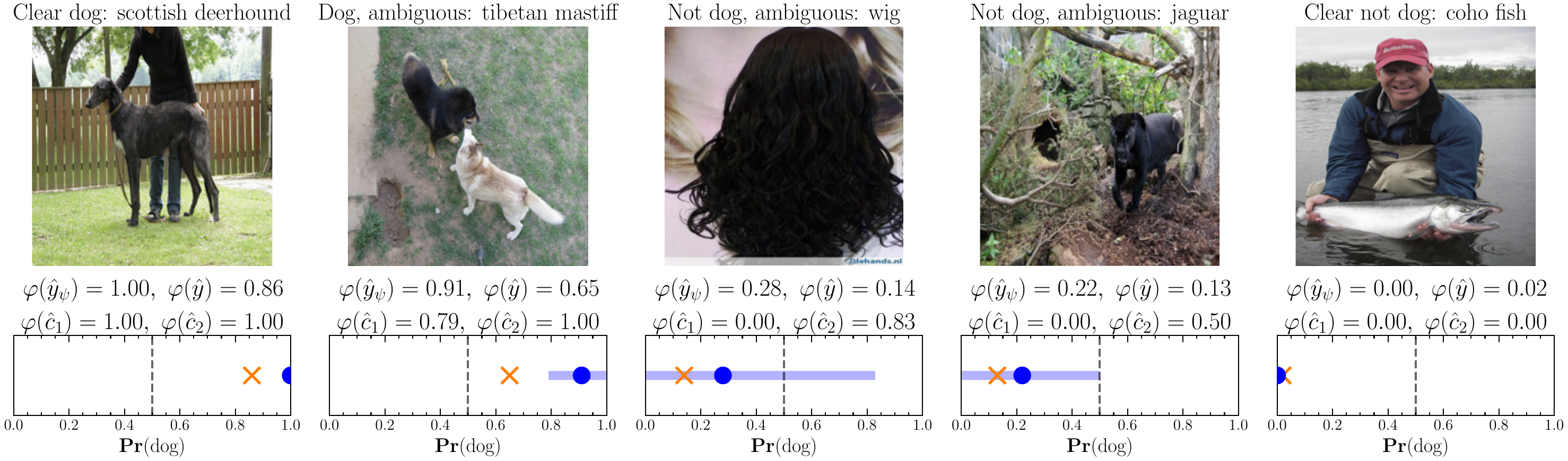}
\vspace{-20 pt}
\caption{\small{
From left to right, prediction intervals for $\mathbf{Pr}(\text{dog})$ computed with a \texttt{ConvNeXt} classifier~\citep{convnext}. Far left, the interval collapses at $\mathbf{Pr}(\text{dog}) = 1$ for the clear {\em dog} image. At the centre, the most ambiguous case {\em and} the longest interval. Far right, the interval collapses at $\mathbf{Pr}(\text{dog}) = 0$ for the clear {\em not dog} image. For each example, we report the lifted prediction probability---$\varphi( \hat{y}_{\psi})$ {\color{blue} blue dot}---\texttt{ConvNeXt} prediction probability---$\varphi( \hat{y})$ {\color{orange} orange cross}---and the lower $\varphi(\hat{c}_1)$ and upper $\varphi(\hat{c}_2)$ interval endpoints.
}
}
\vspace{-15 pt}
\label{fig: dog}
\end{figure}

Figure~\ref{fig: dog} reports five out-of-sample examples selected from a candidate pool that is disjoint from the calibration set. The first (far left) and last (far right) panels correspond to a \textit{clear dog} and a \textit{clear not dog} image, respectively, and provide reference points where MAPS returns tight intervals around the \texttt{ConvNeXt} point estimate. The three middle panels illustrate intermediate regimes: an image that corresponds to a dog according to its ground truth label but appears ambiguous, for which the predicted label is positive and the MAPS interval lies largely above $0.5$ while remaining non-degenerate. The image at the centre is an uncertain case with $\varphi (\hat{c}_1 (\x_o)) < 0.5< \varphi (\hat{c}_2 (\x_o))$, where $\varphi$ is the \texttt{sigmoid}, and $\hat{c}_1 (\x_o) = \hat{y}_{\psi} (\boldsymbol{x}_{o} ) + \widehat{q}_{R_o^{*b}} (\alpha/2) $ and $\hat{c}_2 (\x_o) = \hat{y}_{\psi} (\boldsymbol{x}_{o} ) +\widehat{q}_{R_o^{*b}} (1 - \alpha/2) $ are the endpoints in~\eqref{eq: MAPS binary class}. 
\par
MAPS identifies ambiguity at the level of the dog probability itself; for instance, in the jaguar image, where the predicted label is $\textit{not dog}$, yet the MAPS interval is not concentrated near $\mathbf{Pr}(\text{dog})=0$. When comparing the prediction probabilities $\varphi( \hat{y})$ from the original pretrained model with the lifted probabilities $\varphi( \hat{y}_{\psi})$, we observe that the logistic LPM shifts predictions towards greater certainty for non-ambiguous images, while shifting them towards $\mathbf{Pr}(\text{dog})=0.5$ for ambiguous inputs. Across these five new observations, MAPS produces an interpretable decomposition into confident and ambiguous inputs in terms of the intervals, thereby refining the raw predictive probabilities produced by the classifier. This is a result of non-negligible estimation errors, which could arise if, given the test image, the model $\hat{f}$ cannot discriminate consistently between different labels. For example, $y_{\psi} (\x_o)$ can itself be modelled as a random variable that quantifies the uncertainty linked to the ambiguous logit $\hat{f} (\x_o)$ produced by a given image. We refer to this as {\em double uncertainty} (e.g., the image at the centre in Fig.~\ref{fig: dog}). These five images show how the confidence intervals increase in length as the certainty of {\em dog} decreases, are longest for the most ambiguous cases, and start to shorten once the certainty of {\em not dog} begins to increase. Figure~\ref{fig: dog} can be thought of as a confidence gradient for predictive probabilities.
\vspace{-5 pt}
\subsection{Discussion for supervised learning}\label{sec:final_discussion}
From a high-angle perspective, statistical machine learning consists of finding the adequate objective function for a given task. In supervised learning, this is usually achieved by selecting a loss function that balances a data-fidelity term with a regularisation one. This is done to avoid overfitting, and produce accurate predictions for out-of-sample data. The LPM exploits the information encoded in a given predictive model by ``wrapping" itself around it and quickly assess its reliability in a calibration set. Intuitively, we can think of a predictive model as a ``machine" that {\em compresses} the information in a high-dimensional covariate useful for predicting a response variable of interest, and of the LPM as a  quick off-the-shelf ``scanner" that quickly assesses its effectiveness, which can perform model assessment and uncertainty quantification.
\par
Recently, \cite{wilson2025deeplearningmysteriousdifferent} noted that soft-inductive biases (flexible regularisation schemes) can lead to good generalisation for a variety of model classes, including deep neural networks. These models ``learn" functions of interactions between predictors, so called features, that can produce accurate predictions in high-dimensional problems. For example, the neural estimators in Section~\ref{subsec:maps_for_sbi} effectively use a $38,400$-dimensional predictor to estimate the target response, and the classifier in Section~\ref{subsec:MAPS_binary} uses resolutions of at least $224 \times 224$, i.e., the predictors are at least $50,176$-dimensional, to classify images. These models ``learn" to exploit complex interactions between predictor variables, which can be thought of as useful contours in high-dimensional space. We believe the success of these methods could be related to contour-homoscedasticity, which would enable learning algorithms to quickly identify features along contours, rather than having to explore vast high-dimensional spaces. \cite{whiteley2025statisticalexplorationmanifoldhypothesis} discuss a similar notion to explore the manifold hypothesis.
\par
In other words, an accurate predictive model is composed of optimal compressions of interactions between covariates for predicting out-of-sample responses. These compressions can be used to unwrap the black-box from above. The richer the lifted conditional structure, the more we can unravel a given black-box predictive model.
\addtocontents{toc}{\protect\setcounter{tocdepth}{0}{}}
\vspace{-5 pt}
\section*{Conclusion}
\label{sec: conclusion}
We have introduced a model-agnostic framework for supervised learning that provides valid conditional coverage guarantees and scales with the number of predictors or features, or both of the former. MAPS is a flexible, easily deployable method that wraps around complex predictive models. Our coverage guarantees are asymptotic, given the impossibility of finite-sample guarantees. However, these enable uncertainty quantification for a targeted individual rather than an abstract population and, crucially, depend on scalar estimators that often converge quickly to optimal prediction intervals, that is, approximation errors are negligible for sensible calibration sets. If, however, spline or KDE-based estimators perform poorly for a particular model or calibration set, MAPS readily accommodates alternative choices, such as local linear regression or quantile regression, without altering the overall framework. The LPM is still at a preliminary stage of development; some exciting generalisations we intend to pursue in future work are: multivariate responses, richer lifted conditional structures, and prediction intervals for time series data. Code for reproducing our results is available at~\url{github.com/danleonte/MAPS}. 
\vspace{-5 pt}
{\small
\subsection*{Author contributions} Conceptualisation: D.S., D.L.; Methodology: D.S., D.L.; Software: D.L., K.M.;  Validation: D.S., D.L., K.M.; Formal analysis: D.S., D.L., K.M.; Investigation: D.S., D.L., K.M.; Data curation: D.L., K.M.; Writing--original draft: D.S., D.L., K.M.; Writing--review and editing: D.S, D.L.; Visualisation: D.L., K.M.; Project admin: D.S., D.L.; Supervision: D.S.  
\vspace{-5 pt}
\subsection*{Acknowledgments}
We thank Mario Cortina-Borja, Daniyar Ghani, Rapha\"el Huser, Guy Nason, Tobias Schr\"oder, Jakub Rybak, Jianwei Shi, Yan Song and Yanbo Tang for insightful discussions and suggestions. We thank Matthew Sainsbury-Dale for making available the neural point estimators used in Section \ref{subsec:maps_for_sbi}. D.S. acknowledges support from the Institute of Child Health Great Ormond Street, Imperial College London, the Bank of Mexico and EPSRC NeST Programme grant EP/X002195/1. K.M. acknowledges support from the President's PhD Scholarship at Imperial College London. Part of this research was conducted during a visit to AIMS, South Africa.
}

\addtocontents{toc}{\protect\setcounter{tocdepth}{1}}
\vskip 0.2in
\small{
\newpage
\begin{center}
    {\large\bf Supplementary Material for \vspace{0.05cm}\\``The MAPS algorithm: Fast model-agnostic and \vspace{0.075cm}\\ distribution-free prediction intervals for supervised learning"} 
    \vspace{0.25cm}\\
    { Daniel Salnikov \quad Dan Leonte \quad Kevin Michalewicz}
\end{center}
\renewcommand*\contentsname{Appendices}
\appendix
\setcounter{equation}{0}
\counterwithin{equation}{section}
\tableofcontents
\label{supp_sec: notation summary}
\section{Computational details}
\label{app: comp details}
%
We provide a \texttt{Python} implementation and use the \texttt{JAX} library to parallelise spline and kernel density fitting across bootstrap samples, and conditional sampling for different $\bx_o$. Our implementation computes bivariate Gaussian KDEs to estimate the joint density of $(\hat{u},\hat{f})$ using the pairs $(\hat{u}_{\calb{i}}, \, \hat{f} (\bx_{\calb{i}}))$ and the resulting conditional distribution of $\hat{u} \mid \hat{f} (\bx)$, which is a mixture of Gaussians with $\calb{n}$ terms, that is, the nonparametric estimator in~\eqref{eq: error CDF estimation}. It also performs automatic data-driven selection of both kernel bandwidths, and of the spline penalty hyperparameter $\lambda$ via GCV for each bootstrap draw. MAPS can be tailored to homoscedastic errors to further increase computational efficiency. Increasing $n_\text{cal}$ improves the performance of MAPS but can also make conditional sampling computationally expensive. In such cases, we can decouple the cost of evaluating and sampling from the KDE from $n_\text{cal}$ by building a low-rank approximation using products of Chebyshev polynomials, which results in efficient sampling by the inverse transform \citep{olver2013fast, leonte2025simulation}, we leave this for future work. The bootstrap step may be omitted for large $n_\text{cal}$, as it mainly improves performance when data are scarce, and asymptotic guarantees remain valid without it.
\newline
\\
On a different note, Algorithm~\ref{alg: IRLS for lifted logistic LPM} details the IRLS Algorithm~\citep[Chapter 4][]{statLearn} used to fit the generalised LPM in~\eqref{eq: bernoulli LPM}--\eqref{eq: logistic LPM} in Section~\ref{subsec:MAPS_binary}.
\begin{algorithm}[H]
\caption{ {\bf IRLS} for lifted binary classification} \label{alg: IRLS for lifted logistic LPM}
    \begin{algorithmic}[1]
        \Require  {\small $(\calb{\mathbf{X}}, \calb{\boldsymbol{y}})$, $\varphi$, $\hat{f}$, $\boldsymbol{\beta}^{(0)}$}
        \newline
        {\small {\bf Repeat until convergence} $(t \gets t + 1)$:}
        \State {\small $\boldsymbol{z}^{(t)} \gets \mathbf{S} (\hat{f}) \, \boldsymbol{\beta}^{(t - 1)} + \mathbf{W} (\varphi_{\psi})^{-1} \, \left ( \calb{\boldsymbol{y}} - \boldsymbol{\varphi}_{\psi} \right )$}
        \State {\small $\boldsymbol{\beta}^{(t)} \gets \underset{\beta}{\operatorname{argmin}} \left \{ \big ( \boldsymbol{z}^{(t)} - \mathbf{S} (\hat{f}) \, \boldsymbol{\beta} )^T \, \mathbf{W} (\varphi_{\psi}) \, (\boldsymbol{z}^{(t)} - \mathbf{S} (\hat{f}) \, \boldsymbol{\beta} \big ) + \lambda \, \boldsymbol{\beta}^T \mathbf{\Omega} (\hat{f}) \, \boldsymbol{\beta} \right \}$} \Comment{{\small WLS problem}}
        \State {\small $\boldsymbol{\varphi}_{\psi} \gets \left \{ \varphi \left ( \sum_{j = 1}^{\calb{N}} \hat{\beta}_{j} s_{j} ( \hat{f} (\bx_{\calb{i}})) \right ) \right \}_{\calb{i} = 1}^{\calb{i} = \calb{n}}$} \Comment{\small $\calb{N}$ is the number of B-splines}
        \State {\small $\mathbf{W} (\varphi_{\psi}) \gets \operatorname{diag} \left ( \boldsymbol{\varphi}_{\psi} \right )$}
    \end{algorithmic}
     {\small $\mathbf{Return:} \, \boldsymbol{\hat{\beta}}$}
\end{algorithm}
For smaller calibration sets, it might be useful to sample estimated pivots rather than directly by inversion or from $\widehat{P}_{\hat{u} | \hat{f}}$. Remark~\ref{rem: quant transform sampling} details the sampling procedure. 
\begin{remark}\label{rem: quant transform sampling}
    We could estimate the pivotal residuals using the conditional distribution function estimator, i.e., $\widehat{V}_{\calb{i}} = \widehat{P}_{\hat{u} | \hat{f}} ( \hat{u}_{\calb{i}} \, | \,  \hat{f} (\bx_{\calb{i}}) )$, and sample with replacement before applying the quantile transform. This could increase accuracy for small samples, however, these procedures are asymptotically equivalent, i.e., as $\calb{n} \to \infty$ both guarantee asymptotic conditional coverage. 
\end{remark}

\vspace{-10 pt}
\section{{\small Multimodal distributions, categorical predictors and multiple splits}}
\label{app: cat and multiple splits}
Let $M \in \{ 1, \dots, N_m \}$ be a discrete random variable with $N_m \in \mathbb{N}$ categories, such that each number is linked to a label, e.g., $1$ denotes controls and $2$ treatments. Then, the multimodal regression function is the weighted sum of $m$-label regression functions,
\begin{equation}\label{eq: m-modal reg function}
    f (\bx_o) = \sum_{m_o = 1}^{N_m} f_{m_o} (\bx_o) \, p_{M} (m_o),
\end{equation}
where $f_{m_o} (\bx_o) = \mathbb{E} \left ( Y_o \, | \, \bX_o = \bx_o, \, M_o = m_o \right )$ and $p_{M} (m_o) = \mathbf{Pr} ( M_o = m_o )$ is label probability. Similarly, the $m$-label lifted residual is
$$
    \hat{u} ( \bx_o, \, m_o ) := Y_o - y_{\psi} (\bx_o, \, m_o),
$$
where $y_{\psi} (\bx_o, \, m_o) = \psi_{m_o} ( \hat{f}_{m_o} (\bx_o) )$ is the $m$-label LPM, $\hat{u} ( \bx_o, \, m_o ) \sim P_{\hat{u} | m_o }$, and $ P_{\hat{u} | m_o }$ is the residual distribution function given $\hat{f}_{m_o} (\bX_o)$ and $M_o = m_o$. Let $\widehat{C}_{\operatorname{maps}} (\bx_o, \, m_o)$ denote~\eqref{eq: maps interval dist-free} at the $(1 - \alpha \cdot  p_{M} (m_o) ) \times 100 \%$ confidence level for a fixed $m_o$, i.e., its centre is $\hat{y}_{\psi} (\bx_o, \, m_o) = \psibiashat_{m_o} ( \hat{f}_{m_o} (\bx_o) )$, where we estimate $ \psi_{m_o}$ in~\eqref{eq: dist-free additive errors} only using calibration data linked to label $m_o$. Then, by Bonferroni arguments, we have that
\begin{equation}\label{eq: z-label maps sets}
    \widehat{C}_{\operatorname{maps}} (\bx_o ) = \bigcup_{m_o = 1}^{N_m} \widehat{C}_{\operatorname{maps}} (\bx_o, \, m_o),
\end{equation}
is a prediction interval, without knowledge of $m_o$, at the $(1 - \alpha ) \times 100 \%$ confidence level. If there is only one label, then~\eqref{eq: z-label maps sets} is equal to~\eqref{eq: maps interval dist-free}.
\par
Further, if the labels are equiprobable (i.e., $ p_{M} (m_o) = 1 / N_m$), then the adjusted confidence levels are $(1 - \alpha / N_m ) \times 100 \%$, which is the standard Bonferroni adjustment. If the $p_{M} (m_o)$ are unknown, then we can use an estimator, $\widehat{p}_{M} (m_o)$, for adjusting confidence levels. This estimation procedure can be incorporated into Algorithm~\ref{alg: MAPS add errors}, as well as for multiple splits. If we compute $K$ intervals, $\widehat{C}_{\operatorname{maps},  k} \,  (\bx_o ) $, each at the $(1 - \alpha / K ) \times 100 \%$ confidence level, where $k \in \{ 1, \dots, K \}$ denotes the $k$th-split, then $\widehat{C}_{\operatorname{maps}} (\bx_o; \, K ) := \cap_{k = 1}^K \widehat{C}_{\operatorname{maps},  k} \,  (\bx_o ) $ is a prediction interval at the  $(1 - \alpha ) \times 100 \%$ confidence level. These extensions are valid~\citep{dist_free_conf_inference} for prediction intervals. 
\begin{remark}
\label{rem: studentised lifted residuals}
MAPS can also be adapted to sample ``studentised" residuals, as mentioned in Section~\ref{sec: maps}. 
Indeed, {\em MAPS} can be tailored to incorporate additional model-agnostic scale structure. For example, if good estimators $\hat{\sigma}^2_{\hat{f}} (\bx_o) $ of $\, \mathbb{E} \, [ u_o^2 \, | \, \hat{f} (\bx_o) ]$ and $\hat{\tau}_{\hat{f}} (\bx_o) $ of $\, \mathbb{E} \, [ |u_o| \, | \, \hat{f} (\bx_o) ]$ are available, then {\em MAPS} can directly estimate the ``studentised" lifted residuals given by
   \begin{equation}\label{eq: studentised lifted residuals}
       \frac{Y_o - \hat{y}_{\psi}(\bx_o)}{\hat{\sigma}_{\hat{f}} (\bx_o) } \, \, \, \, and \, \, \, \, \frac{Y_o - \hat{y}_{\psi}(\bx_o)}{\hat{\tau}_{\hat{f}} (\bx_o) }. 
   \end{equation}
\end{remark}
\section{MLP experiments with Gaussian errors}\label{app: sim study details}
We generate data using a regression function given by the sum of two test functions that do not have principal directions of variation, and iid predictior with standard normal distributions. This regression function is a challenging estimation problem for MLPs because the two base functions do not vary in one particular direction, i.e., these are radial functions. The test functions are a \texttt{doppler} and damped \texttt{cosine} wave~\citep[Chapter 11]{statLearn}. 
\begin{equation}
    f_{\mathrm{doppler}} (z) = a \sqrt{|z| ( 1 - |z|)} \sin \left \{ b \pi (z + c)^{-1} \right \}, 
\end{equation}
\begin{equation}
     f_{\mathrm{damcos}} (z_1, z_2) = a \exp \{ - b r(z_1, z_2) \} \cos \{ a \pi r(z_1, z_2) \},
\end{equation}
$$r(z_1, z_2) = \{ (z_1 - c_1)^2 + (z_2 - c_2)^2 \}^{1 / 2}.$$
Let $z_i  ( \boldsymbol{x} ) := \sum_{j = 1}^{10} w_{ij} x_j$, where $w_{ij} = \mathbb{I} (j \in I_{z_i}) / 10$ is the weight for the $i$th interaction path for predictor $X_j \sim \mathcal{N} (0, 1)$, $i = 1, 2, 3$, $j = 1, \dots, 10$, and $I_{z_i}$ is the set of active indices for $z_i$. The data-generating process is
\begin{equation}
    f_{\mathrm{sim}} ( \boldsymbol{X} ) = f_{\mathrm{doppler}} (z_1  ( \boldsymbol{X} ) ) +  f_{\mathrm{damcos}} (z_2  ( \boldsymbol{X} ), z_3  ( \boldsymbol{X} ) ) + \epsilon,
\end{equation}
where $\epsilon \sim \mathcal{N}(0, \sigma_{\epsilon}^2)$, and $\sigma_{\epsilon}=0.1$. We use the same training settings for the MLP as in~\cite{salnikov2024liftedcoefficientdeterminationfast}.

\begin{figure}
    \centering
\includegraphics[width=\linewidth]{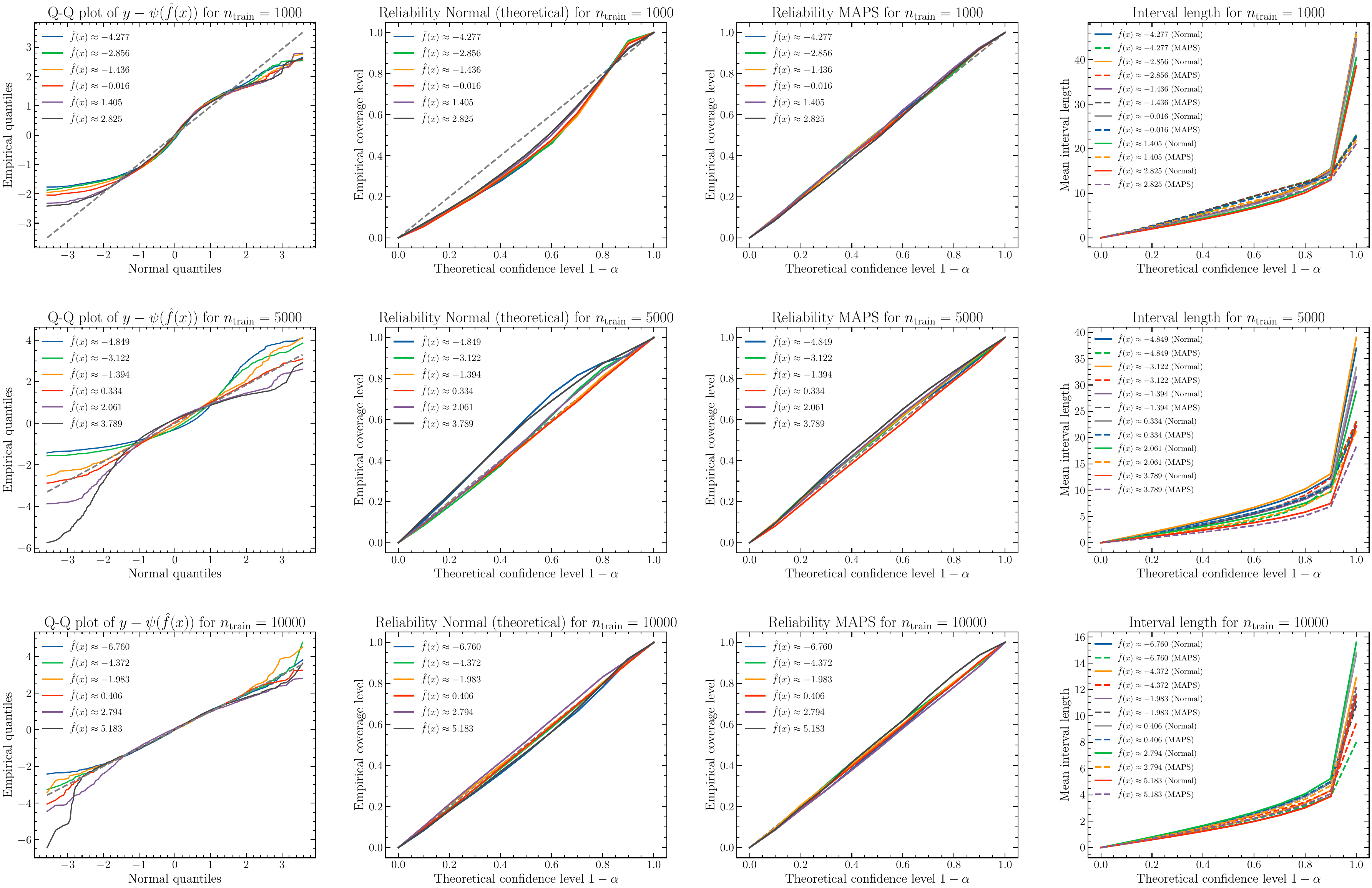}
\caption{{\small Results for the experiment in Appendix~\ref{app: sim study details} for $n_{\mathrm{train}}\in\lbrace 1000, 5000, 10000\rbrace$.}}
\label{fig: sim study fig}
\end{figure}
From left to right, the first column in Figure~\ref{fig: sim study fig} shows the QQ-plot of lifted testing set residuals, which evidence the non-Gaussian errors. The second one shows reliability plots for prediction intervals computed using the theoretical normal quantiles, and the third one shows reliability plots for prediction intervals computed with MAPS. The fourth column shows mean interval length and highlights that MAPS attains coverage with shorter intervals for smaller samples. These comparisons highlight that MAPS achieves finite-sample coverage and converges to optimality, as the theoretical ones do, as sample size increases.
\subsection{Reliability diagram details}
The reliability diagram from Figure \ref{fig: sim study fig} is obtained by considering $\alpha_i = i/10$ for $0 \le i \le 10$ and plotting the empirical coverage at significance levels $\alpha_i$, on the $y$ axis, against the theoretical confidence levels $1-\alpha_i$, on the $x$ axis. Whence, if these two lines agree, then the prediction intervals properly describe out-of-sample variation and are useful for computing generalisation errors using model $\hat{f}$. Put another way, they provide statistical evidence that model $\hat{f}$ generalises and a quick tool for computing out-of-sample prediction intervals.
\begin{definition}Let $\left\{\left(\hat{f}(x_i), y_i \right)\right\}_{i=1}^{n_{\mathrm{test}}}$ be a test set and $\alpha \in (0,1)$. Further let $\boldsymbol{x} \mapsto I_\alpha(\boldsymbol{x})$ be a mapping that returns prediction intervals at significance level $\alpha$ for $y | \hat{f}\left(\boldsymbol{x}\right)$. The empirical coverage at level $\alpha$ is given by
\begin{equation*}
    \frac{1}{n}\sum_{i=1}^{n_\mathrm{test}} \boldsymbol{1}\left(y_i \in I_\alpha(\boldsymbol{x}_i)\right).
\end{equation*}
\end{definition}
%

\section{Proofs for the MAPS Algorithm}
\subsection{Proof of Proposition~\ref{prop: ideal optimal interval}}
\label{app: proof of prop 1}
{\bf Step I:} Show that $C_{\mathrm{ideal}} (\bx_o)$ has the desired coverage.
By assumption $P_{\epsilon_o | \bx_o} $ is continuous, so there is a one-to-one map between the quantiles of $Y_o$ and $\epsilon_o$, that is, given $\bX_o = \bx_o$ we have that
\begin{equation}\label{eq: qunatile mapping}
    q_{y_o | \bx_o} ( \alpha) = f(\bx_o) + q_{\epsilon_o | \bx_o} (\alpha),
\end{equation}
for all $\alpha \in (0, \, 1)$ and $\bx_o \in \mathcal{X}$, and we see that
\begin{equation}\label{eq: ideal coverage guarantee}
     \mathbf{Pr} \big ( Y_o \in  C_{\mathrm{ideal}} (\boldsymbol{x}_o)  \, | \, \boldsymbol{X}_o = \boldsymbol{x}_o  \big ) = 1 - \alpha,
\end{equation}
for all $\alpha \in (0, \, 1)$ and $\bx_o \in \mathcal{X}$, where $C_{\mathrm{ideal}} (\bx)$ is given by~\eqref{eq: ideal oracle confidence region}.
\ \\
\newline
{\bf Step II:} Show that any solution $ [f (\bx_o) + q_0, \, f(\bx_o) + q_1]$, satisfies $$| q_1 - q_0 | \leq | q_{\epsilon_o | \bx_o} ( \alpha / 2) - q_{\epsilon_o | \bx_o} (1 - \alpha / 2)|,$$
\begin{equation}\label{eq: solution conditions}
     P_{\epsilon_o | \bx_o } (q_1) = P_{\epsilon_o | \bx_o } (q_0)  + 1 - \alpha \leq 1.
\end{equation}
\begin{proof}
Suppose that there exist $q_0, \, q_1 \in \mathbb{R}$ such that the solution to~\eqref{eq: optmial pred interval optimisation} is
\begin{equation}\label{eq: contradiction ineq equation}
    [f (\bx_o) + q_0, \, f(\bx_o) + q_1], \, \, \, \, and  \, \, \, \, P_{\epsilon_o | \bx_o } (q_1) - P_{\epsilon_o | \bx_o } (q_0) > 1 - \alpha. 
\end{equation}
Since $P_{\epsilon_o | \bx_o }$ is monotone increasing, for all $q \in (q_0, \, q_1)$ we have that
\begin{equation*}
    P_{\epsilon_o | \bx_o } (q_0) < P_{\epsilon_o | \bx_o } (q) < P_{\epsilon_o | \bx_o } (q_1). 
\end{equation*}
Hence, after rearranging, we have that
\begin{equation*}
    0 \leq P_{\epsilon_o | \bx_o } (q_1) - P_{\epsilon_o | \bx_o } (q) \leq 1 - \alpha. 
\end{equation*}
Since $P_{\epsilon_o | \bx_o }$ is continuous, by the intermediate value theorem there exists a $q_* \in (q_0, \, q_1)$ such that
\begin{equation}\label{eq: contradiction prop one proof}
     P_{\epsilon_o | \bx_o } (q_1) - P_{\epsilon_o | \bx_o } (q_*) = 1 - \alpha.
\end{equation}
Since $| q_1 - q_* | < | q_1 - q_0|$, by~\eqref{eq: qunatile mapping}, we see that $[f (\bx_o) + q_*, \, f(\bx_o) + q_1]$ has coverage $1 - \alpha$ and a {\em shorter} length. Hence,~\eqref{eq: contradiction ineq equation} being the solution to~\eqref{eq: optmial pred interval optimisation} is a {\em contradiction}. Thus, any solution satisfies $P_{\epsilon_o | \bx_o } (q_1) - P_{\epsilon_o | \bx_o } (q_0)  = 1 - \alpha $, and we arrive at~\eqref{eq: solution conditions}.
\end{proof}
{\bf Step III:} Show that the unique solution for unimodal densities is given by~\eqref{eq: ideal oracle confidence region},
\begin{equation*}
    [ f(\boldsymbol{x}_o) + q_{\epsilon_o | \bx_o} ( \alpha / 2) + h_0, \, f(\boldsymbol{x}_o) + q_{\epsilon_o | \bx_o} (1 -  \alpha / 2) + h_1 \big ],
\end{equation*}
and the sign of the shape adjustments $h_0, \, h_1$.
\begin{proof}
By~\eqref{eq: solution conditions}, we can find the solution to~\eqref{eq: optmial pred interval optimisation} by minimising
\begin{equation}\label{eq: optimisation problem}
    P_{\epsilon_o | \bx_o }^{-1} (1 - \alpha / 2 + t ) -  P_{\epsilon_o | \bx_o }^{-1} (\alpha/ 2 + t) 
\end{equation}
for $t \in [-\alpha / 2, \, \alpha / 2]$. This shows that the optimal quantiles can be recovered directly from the ``symmetric" ones as an adjustment that depends on the shape of $P_{\epsilon_o | \bx_o }$. These solutions need not be unique, e.g., a standard uniform distribution. 
\par
However, if we further assume that the density function exists and has a unique maxima, i.e., $ P_{\epsilon_o | \bx_o }$ is unimodal with a continuos density, then by differentiating~\eqref{eq: optimisation problem} and solving for the critical point, we see that the minimiser $t_*$ satisfies
\begin{equation}\label{eq: critical point cond prop 1}
    p_{\epsilon | \bx_o} \left (  P_{\epsilon_o | \bx_o }^{-1} (1 - \alpha / 2 + t_* ) \right ) =  p_{\epsilon | \bx_o} \left (  P_{\epsilon_o | \bx_o }^{-1} ( \alpha / 2 + t* ) \right ),
\end{equation}
hence, if $t_* < 0$, then the quantiles are adjusted to the left, if $t_* > 0$, to the right, and if $t_* = 0$ these are not adjusted. Let $h_0, \, h_1 \in \mathbb{R}$ be the unique solutions to $ q_{\epsilon_o | \bx_o} (1 - \alpha / 2) + h_1 = P_{\epsilon_o | \bx_o }^{-1} (1 - \alpha / 2 + t_* ) $ and $ q_{\epsilon_o | \bx_o} (\alpha / 2) + h_0 = P_{\epsilon_o | \bx_o }^{-1} (\alpha / 2 + t_* )$. Then, we verify three cases. If the density is symmetric about zero, we have that $q_{\epsilon_o | \bx_o} (1 - \alpha / 2) = - q_{\epsilon_o | \bx_o} (\alpha / 2)$, so $t_* = 0$ solves~\eqref{eq: critical point cond prop 1} and we have that $h_0 = h_1 = 0$. Thus, the solution for symmetric densities is
\begin{equation*}
    [ f(\boldsymbol{x}_o) + q_{\epsilon_o | \bx_o} ( \alpha / 2), \, f(\boldsymbol{x}_o) + q_{\epsilon_o | \bx_o} (1 -  \alpha / 2) \big ].
\end{equation*}
{\bf Case (ii):} right-skewed density. That is, $$p_{\epsilon | \bx_o} (q_{\epsilon_o | \bx_o} (1 - \alpha / 2)) > p_{\epsilon | \bx_o} (q_{\epsilon_o | \bx_o} (\alpha / 2)).$$ 
\par
If $q_{\epsilon_o | \bx_o} (1 - \alpha / 2) > \mathrm{mode} (p_{\epsilon | \bx_o} ) > q_{\epsilon_o | \bx_o} (\alpha / 2)$, then we can find $q_* < \mathrm{mode} (p_{\epsilon | \bx_o} )$ such that $p_{\epsilon | \bx_o} (q_{\epsilon_o | \bx_o} (1 - \alpha / 2)) = p_{\epsilon | \bx} (q_*)$. Intuitively, this is the point that is ``symmetric" and closest to the mode of $p_{\epsilon |\bx_o}$ (i.e., a horizontal line that intersects $p_{\epsilon | \bx_o} (\cdot)$ at these two points). Since $p_{\epsilon | \bx_o}$ is monotone increasing for $q < \mathrm{mode} (p_{\epsilon | \bx_o} )$, we have that $q_{\epsilon_o | \bx_o} (\alpha / 2) < q_*$, so we have that 
\begin{equation*}
   1 - \alpha =  \int_{q_{(\alpha / 2)}}^{q_*}  p_{\epsilon | \bx_o} (t) \, dt +  \int_{q_*}^{q (1 - \alpha / 2)}  p_{\epsilon | \bx_o} (t) \, dt,
\end{equation*}
hence, the ``symmetric" interval does not attain the required coverage,
\begin{equation*}
     \int_{q_*}^{q_{\epsilon_o | \bx_o} (1 - \alpha / 2)}  p_{\epsilon | \bx_o} (t) \, dt < 1 - \alpha.
\end{equation*}
We proceed to ``push" the horizontal line down by decreasing $q_*$ and increasing $q_{\epsilon_o | \bx_o} (1 - \alpha / 2)$. The integral above is continuous, hence, we can find $b_0 < 0$ and $h_1 > 0$ such that 
\begin{equation*}
     \int_{q_* + b_0}^{q_{\epsilon_o | \bx_o} (1 - \alpha / 2) + h_1}  p_{\epsilon | \bx_o} (t) \, dt = 1 - \alpha  \, \, \, \, and  \, \, \, \, p_{\epsilon | \bx_o} (q_* + b_0) = p_{\epsilon | \bx_o} (q_{\epsilon_o | \bx_o} (1 - \alpha / 2) + h_1).
\end{equation*}
By~\eqref{eq: critical point cond prop 1} this is the unique solution to~\eqref{eq: optmial pred interval optimisation}, thus, we see that
\begin{equation*}
    q_{\epsilon_o | \bx_o} (1 - \alpha / 2) + h_1 - (q_* + b_0) <  q_{\epsilon_o | \bx_o} (1 - \alpha / 2) - q_{\epsilon_o | \bx_o} (\alpha / 2),
\end{equation*}
and by rearranging and solving for $q_{\epsilon_o | \bx_o} (\alpha / 2)$, we have that
\begin{equation*}
    q_{\epsilon_o | \bx_o} (\alpha / 2) < q_* + b_0 < q_*,
\end{equation*}
whence, there exists $h_0 > 0$ such that $q_{\epsilon_o | \bx_o} (\alpha / 2) + h_0 = q_* + b_0$. 
\par
If $q_{\epsilon_o | \bx_o} (1 - \alpha / 2) < \mathrm{mode} (p_{\epsilon | \bx_o} )$. Then, the argument is the same but the upper quantile gets ``pushed" to the right of the mode and the lower one closer to its left. This follows from using integrals and noting that for $q >  \mathrm{mode} (p_{\epsilon | \bx_o} )$, we have $1 - P_{\epsilon | \bx_o} (q) < \alpha / 2$ in this case. Thus,  $h_0, \, h_1 > 0$ and the solution is
\begin{equation*}
    [ f(\boldsymbol{x}_o) + q_{\epsilon_o | \bx_o} ( \alpha / 2) + h_0, \, f(\boldsymbol{x}_o) + q_{\epsilon_o | \bx_o} (1 -  \alpha / 2) + h_1 \big ].
\end{equation*}
{\bf Case (iii):} left-skewed density. That is, $$p_{\epsilon | \bx_o} (q_{\epsilon_o | \bx_o} (1 - \alpha / 2)) < p_{\epsilon | \bx_o} (q_{\epsilon_o | \bx_o} (\alpha / 2)).$$
\par
If $q_{\epsilon_o | \bx_o} (1 - \alpha / 2) > \mathrm{mode} (p_{\epsilon | \bx_o} ) > q_{\epsilon_o | \bx_o} (\alpha / 2)$, then analogously to the argument above we can find $q_*$ such that $p_{\epsilon | \bx_o} (q_{\epsilon_o | \bx_o} (\alpha / 2)) = p_{\epsilon | \bx} (q_*)$ and $\mathrm{mode} (p_{\epsilon | \bx_o} ) < q_* < q_{\epsilon_o | \bx_o} (1 - \alpha / 2)$. Hence, we can ``push" the horizontal line by decreasing $q_{\epsilon_o | \bx_o} (\alpha / 2)$ and increasing $q_*$. That is, in this case $h_0 < 0$ and $b_1 > 0$. This gives the solution
\begin{equation*}
     \int^{q_* + b_1}_{q_{\epsilon_o | \bx_o} (\alpha / 2) + h_0}  p_{\epsilon | \bx_o} (t) \, dt = 1 - \alpha  \, \, \, \, and  \, \, \, \, p_{\epsilon | \bx_o} (q_* + b_1) = p_{\epsilon | \bx_o} (q_{\epsilon_o | \bx_o} (\alpha / 2) + h_0).
\end{equation*}
By~\eqref{eq: critical point cond prop 1}, we arrive at 
\begin{equation*}
    q_* < q_* + b_1 < q_{\epsilon_o | \bx_o} (1 - \alpha / 2),
\end{equation*}
whence, there exists $h_1 < 0$ such that $q_* + b_1 = q_{\epsilon_o | \bx_o} (1 - \alpha / 2) + h_1 < q_{\epsilon_o | \bx_o} (1 - \alpha / 2)$. 
\par
If $q_{\epsilon_o | \bx_o} (\alpha / 2) > \mathrm{mode} (p_{\epsilon | \bx_o} )$, then both quantiles get ``pushed" to the left analogously to the argument above, such that
\begin{equation*}
    q_{\epsilon_o | \bx_o} (\alpha / 2) + h_0 <  \mathrm{mode} (p_{\epsilon | \bx_o} ) < q_{\epsilon_o | \bx_o} (1 - \alpha / 2) + h_1 < q_{\epsilon_o | \bx_o} (1 - \alpha / 2),
\end{equation*}
which also gives $h_0, \, h_1 < 0$. Thus, $h_0, \, h_1 < 0$ and the solution is
\begin{equation*}
    [ f(\boldsymbol{x}_o) + q_{\epsilon_o | \bx_o} ( \alpha / 2) + h_0, \, f(\boldsymbol{x}_o) + q_{\epsilon_o | \bx_o} (1 -  \alpha / 2) + h_1 \big ].
\end{equation*}
Finally, note that $p_{\epsilon |\bx_o} (t) = p_{y | \bx} (f(\bx_o) + t)$, which gives~\eqref{eq:HDR} when substituting the solution into~\eqref{eq: critical point cond prop 1}. This concludes our proof of Proposition~\ref{prop: ideal optimal interval}.
\end{proof}
Given a CDF estimator $\widehat{P}_{\hat{u} | \hat{f}}$ and/or a pdf estimator $\widehat{p}_{\hat{u} | \hat{f}} $, we can quickly find the optimal quantiles using the following procedure.
\subsection*{Estimation of quantile adjustments}
\begin{enumerate}
    \item If $\widehat{p}_{\hat{u} | \hat{f}} (q_{\hat{u} | \hat{f}} (1 - \alpha / 2)) < \widehat{p}_{\hat{u} | \hat{f}} (q_{\hat{u} | \hat{f}} (\alpha / 2))$, then solve the integral equation for $h_0 < 0$ and $b_1 > 0$
    \begin{equation*}
         \int^{q_* + b_1}_{q_{\hat{u} | \hat{f}} (\alpha / 2) + h_0} \widehat{p}_{\hat{u} | \hat{f}} (t) \, dt = 1 - \alpha
    \end{equation*}
    and set $h_1 = q_* + b_1 - q_{\hat{u} | \hat{f}} (1 - \alpha / 2)$.
    \item If $\widehat{p}_{\hat{u} | \hat{f}} (q_{\hat{u} | \hat{f}} (1 - \alpha / 2)) > \widehat{p}_{\hat{u} | \hat{f}} (q_{\hat{u} | \hat{f}} (\alpha / 2))$, then solve the integral equation for $h_1 > 0$ and $b_0 < 0$
    \begin{equation*}
          \int_{q_* + b_0}^{q_{\hat{u} | \hat{f}} (1 - \alpha / 2) + h_1} \widehat{p}_{\hat{u} | \hat{f}} (t) \, dt = 1 - \alpha
    \end{equation*}
    and set $h_0 = q_* + b_0 - q_{\hat{u} | \hat{f}} (\alpha / 2)$.
\end{enumerate}
\subsection{Proof of Lemma~\ref{lem: oracle and ideal coverage}}
\label{app: proof of lemma one}
{\bf Step I:} Show that 
\begin{equation}\label{eq: lem 1 proof dist convergence}
    P_{\hat{\epsilon}_o | \bx_o} (q)  \rightarrow P_{\epsilon | \bx_o} (q),
\end{equation}
for all $q \in \mathbb{R}$ and $\bx_o \in \mathcal{X} \setminus \mathrm{null} (P_{\epsilon | \bx_o})$.
\begin{proof}
    By Markov's inequality for all $\bx_o \in \mathcal{X}$ and $\delta > 0$ we have that
    \begin{equation*}
        \mathbf{Pr} \left ( | \hat{f} (\bx_o) - f(\bx_o) | > \delta \mid \bX_o = \bx_o \right ) \leq \mathbb{E} \, [| \hat{f} (\bx_o) - f(\bx_o) |^2] / \delta^2 \leq \| \hat{f} - f \|_{L_2} / \delta.
    \end{equation*}
    By assumption $\| \hat{f} - f \|_{L_2} / \delta = o_{\mathbf{Pr}} (1) $, hence, 
    \begin{equation*}
        | \hat{f} (\bx_o) - f(\bx_o) | = o_{\mathbf{Pr}} (1).
    \end{equation*}
   The event above has a bounded probability measure for all $\bx_o$. Hence, by Egorov's Theorem~\citep{bartle1982elements} we have that
    \begin{equation} \label{eq: lem 1 proof sup norm convergence}
        \| \hat{f} - f \|_{\infty} = o_{\mathbf{Pr}} (1),
    \end{equation}
    except for points in null sets, i.e., $\bx_o \in \mathrm{null} (P_{\epsilon | \bx_o})$, where $\mathbf{Pr} (\bX_o \in \mathrm{null} (P_{\epsilon | \bx_o})) = 0$. Thus, we have
    \begin{equation*}
        \mathbf{Pr} (| \hat{\epsilon}_o - \epsilon_o | > \delta \mid \bX_o = \bx_o ) =  \mathbf{Pr} \left ( | \hat{f} (\bx_o) - f(\bx_o) | > \delta \mid \bX_o = \bx_o \right ) = o_{\mathbf{Pr}} (1).
    \end{equation*}
    Convergence in probability implies convergence in distribution, so for all $\bx_o \in \mathcal{X} \setminus \mathrm{null} (P_{\epsilon | \bx_o})$, we have that $\hat{\epsilon}_o \overset{\mathbf{d}}{\longrightarrow} \epsilon_o$ given $\bX_o = \bx_o $. This shows~\eqref{eq: lem 1 proof dist convergence}.
\end{proof}
{\bf Step II:} Show that equation~\eqref{eq: oracle and ideal converge} holds. That is, 
\begin{equation*}
    \lambda_{\mathrm{Leb}} \left ( C_{\mathrm{ideal}} \left ( \boldsymbol{x}_o \right ) \Delta \widehat{C}_{\mathrm{oracle}} \left ( \boldsymbol{x}_o \right ) \right ) \overset{\mathbf{Pr}}{\longrightarrow} 0.
\end{equation*}
\begin{proof}
    By~\eqref{eq: lem 1 proof dist convergence} for all $q \in \mathbb{R}$ we have that
    \begin{equation*}
        \alpha_{\hat{\epsilon}_o } (q) =  P_{\hat{\epsilon}_o | \bx_o} (q) = P_{\epsilon | \bx_o} (q) + o(1) = \alpha_{\epsilon_o} (q) + o(1).
    \end{equation*}
    Since $P_{\epsilon | \bx_o}$ is continuous, the quantile function $ P_{\epsilon | \bx_o}^{-1}$ is also continuos. Thus, the sequence $\alpha_{\hat{\epsilon}_o } (q)$ converges for every fixed $q$ when evaluating the quantile function. That is, 
    \begin{equation*}
        P_{\epsilon | \bx_o}^{-1} ( \alpha_{\hat{\epsilon}_o } (q) ) \rightarrow  P_{\epsilon | \bx_o}^{-1} (  P_{\epsilon | \bx_o} (q) ) = q.
    \end{equation*}
    Since $q \in \mathbb{R}$ is arbitrary, for all $\alpha \in (0, \, 1)$ and $\bx_o$ we have that
    \begin{equation}\label{eq: lem 1 proof qunatile convergence}
        q_{\hat{\epsilon}_o | \bx_o } (\alpha) = q_{\epsilon_o | \bx_o} (\alpha) + o(1).
    \end{equation}
    The above also holds for the adjusted ``symmetric" quantiles, so to ease notation we show the result with $h_0 = h_1 =0$. The extension is straightforward by appealing to continuity of the quantile function. Now, note that for all $\bx_o \in \mathcal{X} \setminus \mathrm{null} (P_{\epsilon | \bx_o})$ given $\bX_o = \bx_o $, the symmetric difference (i.e., the length of disjoint subintervals) between the two prediction intervals is
    \begin{align*}
         &\lambda_{\mathrm{Leb}} \left ( C_{\mathrm{ideal}} \left ( \boldsymbol{x}_o \right ) \Delta \widehat{C}_{\mathrm{oracle}} \left ( \boldsymbol{x}_o \right ) \right ) = | \hat{f} (\boldsymbol{x}_o) - f (\boldsymbol{x}_o) + \{q_{\hat{\epsilon}_o | \bx_o } (1 - \alpha/ 2)  - q_{\epsilon_o | \bx_o} (1 - \alpha / 2) \} | + \cdots  \\
         &+ | \hat{f} (\boldsymbol{x}_o) - f (\boldsymbol{x}_o) + \{ q_{\hat{\epsilon}_o | \bx_o } (\alpha/ 2)  - q_{\epsilon_o | \bx_o} (\alpha / 2) \} | \\
         &\overset{(i)}{\leq} 2 \, | \hat{f} (\bx_o) - f(\bx_o) | + | q_{\hat{\epsilon}_o | \bx_o } (\alpha/ 2)  - q_{\epsilon_o | \bx_o} (\alpha / 2) | + | q_{\hat{\epsilon}_o | \bx_o } (1 - \alpha/ 2)  - q_{\epsilon_o | \bx_o} (1 - \alpha / 2) | \\
         &\overset{(ii)}{\leq} 2 \, \| \hat{f} - f \|_{\infty} + | q_{\hat{\epsilon}_o | \bx_o } (\alpha/ 2)  - q_{\epsilon_o | \bx_o} (\alpha / 2) | + | q_{\hat{\epsilon}_o | \bx_o } (1 - \alpha/ 2)  - q_{\epsilon_o | \bx_o} (1 - \alpha / 2) | \\
         &\overset{(iii)}{=} o_{\mathbf{Pr}} (1),
    \end{align*}
    where $(i)$ follows from the triangle inequality, $(ii)$ from the definition of the supremum-norm, and $(iii)$ from~\eqref{eq: lem 1 proof sup norm convergence} and~\eqref{eq: lem 1 proof qunatile convergence}. This concludes our proof of Lemma~\ref{lem: oracle and ideal coverage}.
\end{proof}
\vspace{-10 pt}
\subsection{Proof of Lemma~\ref{lem: stability of the lpm}}
\label{app: proof lemma two}
{\bf Step I:} Show that $\hat{y}_{\psi} (\bx_o)$ is asymptotically normal with mean $y_{\psi} (\bx_o)$, that is,
\begin{equation}\label{eq: lem 2 proof asymp norm}
    \sqrt{a (\ncal)} \{ \yhatpsi (\bx_o) - y_{\psibias} (\bx_o) \} \overset{\boldsymbol{\mathrm{d}}}{\longrightarrow} \mathcal{N} (0, \, \sigma_{\hat{y}}^2 (\bx_o) ).
\end{equation}
\begin{proof}
    By assumption $\hat{y}_{\psi}$ is asymptotically linear. By Definition~\ref{def: asymp linear lpm} we have that
    \begin{equation}\label{eq: lem 2 proof asymp linearity}
       \{ \yhatpsi (\bx_o) - y_{\psibias} (\bx_o) \} =  \frac{ 1}{\ncal } \sum_{\calb{i} = 1}^{\calb{n}} \ell_{\hat{y}} (Y_{\calb{i}}, \, \bx_o, \, \bx_{\calb{i}} ) + o_{\mathbf{Pr}}(1 / \sqrt{a (\ncal)} ).
    \end{equation}
    We define the shorthand 
    $$\sigma_{\ell_i}^2 (\bx_o, \, \bx_{\calb{i}}) := \mathbb{E} \, [ \ell_{\hat{y}}^2 (Y_{\calb{i}}, \, \bx_o, \, \bx_{\calb{i}} ) \mid \hat{f} (\bX_o) = \hat{f} (\bx_o), \, \hat{f} (\bX_{\calb{i}}) = \hat{f} (\bx_{\calb{i}}) ].$$ 
    By assumption, $\sup_{\bx_o, \bx_i} \{ \sigma_{\ell_i}^2 (\bx_o, \, \bx_{\calb{i}}) \} \leq K < \infty$, so by Portmantus Lemma for all $\bx_o, \, \bx_{\calb{i}} \in \mathcal{X}$ and $\delta > 0$ there exists a {\em finite} $M > 0$ such that
    \begin{equation*}
        \mathbf{Pr} \left ( \ell_{\hat{y}}^2 (Y_{\calb{i}}, \, \bx_o, \, \bx_{\calb{i}} ) > M \mid \hat{f} (\bX_o) = \hat{f} (\bx_o), \, \hat{f} (\bX_{\calb{i}}) = \hat{f} (\bx_{\calb{i}}) \right ) \leq \delta \to 0
    \end{equation*}
    for points with non-vanishing measure, i.e., the above holds with probability one as $\calb{n} \to \infty$.
    Thus, for all $\epsilon > 0$ we can derive the following upper bound
    \begin{align*}
        &\int \ell_{\hat{y}}^2 (Y_{\calb{i}}, \, \bx_o, \, \bx_{\calb{i}} ) \, \mathbb{I} (| \ell_{\hat{y}} (Y_{\calb{i}}, \, \bx_o, \, \bx_{\calb{i}} ) | > \epsilon \, \sigma_{\hat{y}} (\bx_o) ) \, d P_{y | \bx_i}\\
        &\leq M \, \int \mathbb{I} (| \ell_{\hat{y}} (Y_{\calb{i}}, \, \bx_o, \, \bx_{\calb{i}} ) | > \epsilon \,\sigma_{\hat{y}} (\bx_o) ) \, d P_{y | \bx_i} \\
        &= M \, \mathbf{Pr} \left ( | \ell_{\hat{y}} (Y_{\calb{i}}, \, \bx_o, \, \bx_{\calb{i}} ) | > \epsilon \,\sigma_{\hat{y}} (\bx_o) \mid \hat{f} (\bX_o) = \hat{f} (\bx_o), \, \hat{f} (\bX_{\calb{i}}) = \hat{f} (\bx_{\calb{i}}) \right )
    \end{align*}
    By Chebyshev's inequality we arrive at the upper bound
    \begin{equation*}
         \int \ell_{\hat{y}}^2 (Y_{\calb{i}}, \, \bx_o, \, \bx_{\calb{i}} ) \, \mathbb{I} (| \ell_{\hat{y}} (Y_{\calb{i}}, \, \bx_o, \, \bx_{\calb{i}} ) | > \epsilon \, \sigma_{\hat{y}} (\bx_o) ) \, d P_{Y | \bx_{i}} \leq \frac{M}{\epsilon^2} \cdot \frac{\sigma_{\ell_i}^2 (\bx_o, \, \bx_{\calb{i}})}{  \sigma_{\hat{y}}^2 (\bx_o) }.
    \end{equation*}
    By assumption, for every fixed $\calb{n}$, each of the individual variances in the sum is bounded. So, as $\calb{n} \to \infty$, we have that
    \begin{equation*}
       \limsup_{\calb{n} \to \infty} \sup_{\bx_i, \bx_o} \frac{ \mathbb{E} \, [\ell_{\psi_{\hat{f}}}^2 ( Y_{\calb{i}}, \, \bx_o, \, \bx_{\calb{i}}) \mid \hat{f} (\bX_o) = \hat{f} (\bx_o), \, \hat{f} (\bX_{\calb{i}}) = \hat{f} (\bx_{\calb{i}}) ]}{ \sum_{\calb{i} = 1}^{ \calb{n} }\mathbb{E} \, [\ell_{\psi_{\hat{f}}}^2 ( Y_{\calb{i}}, \, \bx_o, \, \bx_{\calb{i}}) \mid \hat{f} (\bX_o) = \hat{f} (\bx_o), \, \hat{f} (\bX_{\calb{i}}) = \hat{f} (\bx_{\calb{i}}) ]} \leq \frac{K}{\calb{n} \, K} \to 0.
    \end{equation*}
    Let $ \sigma_{\boldsymbol{\ell}}^2 (\bx_o) := \sum_{i = 1}^{\ncal} \sigma_{\ell_i}^2 (\bx_o, \, \bx_i) $, by the above, for all $\bx_o, \, \bx_{\calb{i}} \in \mathcal{X}$ and $\epsilon > 0$
    \begin{equation}\label{eq: lem 2 proof bounded variance}
       \int \ell_{\hat{y}}^2 (Y_{\calb{i}}, \, \bx_o, \, \bx_{\calb{i}} ) \, \mathbb{I} \left ( | \ell_{\hat{y}} (Y_{\calb{i}}, \, \bx_o, \, \bx_{\calb{i}} ) | > \epsilon \, \sigma_{\boldsymbol{\ell}} (\bx_o) \right ) \, d P_{Y | \bx_i} \leq \frac{M \, \sigma_{\ell_i}^2 (\bx_o, \, \bx_{\calb{i}})}{  \sigma_{\boldsymbol{\ell}}^2 (\bx_o) \, \epsilon^2 } \to 0.
    \end{equation}
    By combining~\eqref{eq: lem 2 proof bounded variance} and the Dominated Convergence Theorem, we have that
    \begin{equation}\label{eq: lem 2 proof Lindberg condition}
        \sum_{\calb{i} = 1}^{\calb{n}} \int \ell_{\hat{y}}^2 (Y_{\calb{i}}, \, \bx_o, \, \bx_{\calb{i}} ) \, \mathbb{I} (| \ell_{\hat{y}} (Y_{\calb{i}}, \, \bx_o, \, \bx_{\calb{i}} ) | > \epsilon \, \sigma_{\boldsymbol{\ell}} (\bx_o) )\, d P_{Y | \bx_i} \to 0,
    \end{equation}
    for all $\bx_o, \, \bx_{\calb{i}} \in \mathcal{X}$ and $\epsilon > 0$. 
    Recall that the $Y_{\calb{i}}$ are conditionally independent, thus, the $ \ell_{\hat{y}} (Y_{\calb{i}}, \, \bx_o, \, \bx_{\calb{i}} )$ are also conditionally independent, and by~\eqref{eq: lem 2 proof Lindberg condition} these satisfy the Lindberg Condition. By the Lindberg-Feller Theorem~\citep[Chapter 2.8]{asym_stats} we see that
     \begin{equation*}
           \frac{ \sqrt{ a (\ncal)} }{\ncal } \sum_{\calb{i} = 1}^{\calb{n}} \ell_{\hat{y}} (Y_{\calb{i}}, \, \bx_o, \, \bx_{\calb{i}} ) \overset{\boldsymbol{\mathrm{d}}}{\longrightarrow} \mathcal{N} (0, \, \sigma_{\hat{y}}^2 (\bx_o) ),
     \end{equation*}
     where $\lim_{\ncal \to \infty} a(\ncal) \, \sigma_{\boldsymbol{\ell}}^2 (\bx_o) / \ncal^2 = \sigma_{\hat{y}}^2 (\bx_o) $. By assumption $\sigma_{\boldsymbol{\ell}}^2 (\bx_o) / \ncal^2 = O (1 / a (\ncal)) \to 0$, thus, by Markov's inequality, we see that
     \begin{equation*}
         \frac{ 1 }{\ncal } \sum_{\calb{i} = 1}^{\calb{n}} \ell_{\hat{y}} (Y_{\calb{i}}, \, \bx_o, \, \bx_{\calb{i}} ) = O_{\mathbf{Pr}} \left( \frac{ \sigma_{\ell}^2 (\bx_o) }{ \ncal^2 } \right ) = O_{\mathbf{Pr}} \left( \frac{ 1 }{ a (\ncal) } \right ) = o_{\mathbf{Pr}}(1 / \sqrt{a (\ncal)} ).
     \end{equation*}
     The remainder in~\eqref{eq: lem 2 proof asymp linearity} is of the same order as above, whence we arrive at~\eqref{eq: lem 2 proof asymp norm}.
\end{proof}
We will use the following preliminary result. Let $T \geq 0$ be a nonnegative random variable with finite mean. Then, for any fixed $M > 0$ we have that
\begin{equation}\label{eq: lem 2 proof pancake theorem}
    \mathbb{E} \, [ T \, \mathbb{I} (T > M) ] = \int_{M}^{\infty} \mathbf{Pr} ( T > t ) \, dt + M \, \mathbf{Pr} ( T > M ).
\end{equation}
{\bf Step II:} Show that the lifted bias vanishes asymptotically, that is,
\begin{equation}\label{eq: lem 2 proof lifted bais cond}
     | \mathbb{E} \, [ \yhatpsi (\bx_o) ] - y_{\psibias} (\bx_o) | = o(1 / \sqrt{a (\ncal)}).
\end{equation}
\begin{proof}
    We denote the sigma-algebra induced by conditioning on the given out-of-sample and calibration set pointwise predictions by the shorthand
    $$\mathcal{F}_{\hat{f}} (\bx_o, \, \calb{\x}) := \sigma (\hat{f} (\bX_o) = \hat{f} (\bx_o), \, \hat{f} (\bX_{\calb{1}}) = \hat{f} (\bx_{\calb{1}}), \dots, \, \hat{f} (\bX_{\calb{n}}) = \hat{f} (\bx_{\calb{n}}) ).$$
    Define $W_{\calb{n}} (\bx_o)$ as the random variable
    \begin{equation}\label{eq: lem 2 proof W defnition}
        W_{\calb{n}} (\bx_o) :=  \frac{ 1 }{ \sigma_{\hat{y}} (\bx_o) } \sum_{\calb{i} = 1}^{\calb{n}} \ell_{\hat{y}} (Y_{\calb{i}}, \, \bx_o, \, \bx_{\calb{i}} ) + R(\bx_o) \overset{\boldsymbol{\mathrm{d}}}{\longrightarrow} \mathcal{N} (0, \, 1).
    \end{equation}
    where $R(\bx_o) = o_{\mathbf{Pr}} (1 / \sqrt{a (\ncal)})$ is the remainder in~\eqref{eq: lem 2 proof asymp linearity}. 
    Let $M > 0$ be fixed. By~\eqref{eq: lem 2 proof pancake theorem},
    \begin{multline}\label{eq: lem 2 proof pancake bound}
        \mathbb{E} \, [ |  W_{\calb{n}} (\bx_o)| \, \mathbb{I} ( | W_{\calb{n}} (\bx_o) | > M ) \mid \mathcal{F}_{\hat{f}} (\bx_o, \, \calb{\x}) ] = \\ \int_{M}^{\infty} \mathbf{Pr} ( | W_{\calb{n}} (\bx_o) | > t \mid \mathcal{F}_{\hat{f}} (\bx_o, \, \calb{\x}) ) \, dt
        + M \, \mathbf{Pr} ( | W_{\calb{n}} (\bx_o) | > M \mid \mathcal{F}_{\hat{f}} (\bx_o, \, \calb{\x}) ).
    \end{multline}
    By Markov's inequality and~\eqref{eq: lem 2 proof pancake theorem}, for all $t > 0$, including $M$, and for all $\calb{n}$ we have that
    \begin{equation}\label{eq: lem 2 proof markov prob bound}
         \mathbf{Pr} ( | W_{\calb{n}} (\bx_o) | > t \mid \mathcal{F}_{\hat{f}} (\bx_o, \, \calb{\x}) ) \leq t^{-2} \, \mathbb{E} \, [  W_{\calb{n}}^2 (\bx_o) \mid \mathcal{F}_{\hat{f}} (\bx_o, \, \calb{\x}) ] \lesssim \frac{1 + o(1)}{t^2}.
    \end{equation}
    Substituting~\eqref{eq: lem 2 proof markov prob bound} into~\eqref{eq: lem 2 proof pancake bound} we see that
    \begin{equation*}
        \limsup_{\calb{n} \to \infty} \mathbb{E} \, [ |  W_{\calb{n}} (\bx_o)| \, \mathbb{I} ( | W_{\calb{n}} (\bx_o) | > M ) \mid \mathcal{F}_{\hat{f}} (\bx_o, \, \calb{\x}) ] \leq \int_{M}^{\infty} t^{-2} \, dt + M / M^2 = 2 / M.
    \end{equation*}
    Thus, $W_{\calb{n}} (\bx_o)$ is asymptotically uniformly integrable, that is, 
    \begin{equation*}
        \lim_{M \to \infty} \limsup_{\calb{n} \to \infty} \mathbb{E} \, [ |  W_{\calb{n}} (\bx_o) | \, \mathbb{I} ( | W_{\calb{n}} (\bx_o) | > M ) \mid \mathcal{F}_{\hat{f}} (\bx_o, \, \calb{\x}) ] = 0.
    \end{equation*}
    By Theorem 2.20 in Chapter 2 of~\cite{asym_stats} and~\eqref{eq: lem 2 proof W defnition} we have that 
    $$\mathbb{E} \, [  W_{\calb{n}} (\bx_o) \mid \mathcal{F}_{\hat{f}} (\bx_o, \, \calb{\x})] \to \mathbb{E} [Z] = 0,$$ 
    where $Z \sim \mathcal{N}(0, \, 1)$. Finally, by~\eqref{eq: lem 2 proof W defnition} and~\eqref{eq: lem 2 proof asymp linearity} we have that
    \begin{equation*}
         \sqrt{a (\ncal)} \left ( \mathbb{E} \, [ \yhatpsi (\bx_o) \mid \mathcal{F}_{\hat{f}} (\bx_o, \, \calb{\x})] - y_{\psibias} (\bx_o) \right ) \to \mathbb{E} \, [  W_{\calb{n}} (\bx_o) \mid \mathcal{F}_{\hat{f}} (\bx_o, \, \calb{\x})] = o(1),
    \end{equation*}
    whence we arrive at~\eqref{eq: lem 2 proof lifted bais cond}. This concludes our proof of Lemma~\ref{lem: stability of the lpm}.
\end{proof}

\vspace{-20 pt}
\subsection{Proof of Theorem~\ref{th: aymp validity of MAPS}}
\label{supp: proof theorem two}
{\bf Step 1:} Show that the quantile function estimator converges uniformly for all $q \in \mathbb{R}$. That is, for all $\bx_o$, given $\hat{f} (\bX_o) = \hat{f} (\bx_o)$, we have
\begin{equation}\label{eq: th 1 proof quantile convergence}
\sup_{\alpha \in (0, \, 1)} \| \widehat{P}_{\hat{u} | \hat{f}}^{-1} - P_{\hat{u} | \hat{f}}^{-1} \|_\infty = o_{\mathbf{Pr}} (1).
\end{equation}
\begin{proof}
    By Assumption~\eqref{eq: consistent dist estimator} and Markov's inequality we have that
    \begin{equation*}
        |  \widehat{P}_{\hat{u} | \hat{f}} (q \mid \hat{f} (\bx_o) ) - P_{\hat{u} | \hat{f}} (q \mid \hat{f} (\bx_o) ) | = o_{\mathbf{Pr}} (1),
    \end{equation*}
    for all $q \in \mathbb{R}$. By Egorov's Theorem and Lemma 2.11 in Chapter 2 of~\cite{asym_stats},
    \begin{equation*}
        \sup_{q \in \mathbb{R}} | \widehat{P}_{\hat{u} | \hat{f}} (q \mid \hat{f} (\bx_o) ) - P_{\hat{u} | \hat{f}} (q \mid \hat{f} (\bx_o) ) | = o_{\mathbf{Pr}} (1). 
    \end{equation*}
    Using the above, by Lemma 1.2.1 of~\cite{politis1999subsampling}, we arrive at~\eqref{eq: th 1 proof quantile convergence}.
\end{proof}
{\bf Step II:} Show that the ``pivots" converge in distribution. That is, for $\widehat{V}_o := \widehat{P}_{\hat{u}_o | \hat{f}} (\hat{u}_o \mid \hat{f} (\bx_o) )$,
\begin{equation}\label{eq: th 1 proof pivot to std unif}
    \mathbf{Pr} ( \widehat{V}_o \leq \alpha ) = \alpha + o_{\mathbf{Pr}} (1),
\end{equation}
where $V_o \sim \mathrm{Unif} (0, \, 1)$ is {\em independent} of  $\hat{f} (\bX_o)$ for all $\bX_{o} \in \mathcal{X}$ and $\mathcal{F}_{\hat{f}} (\bx_o, \, \calb{\x})$.
\begin{proof}
    By~\eqref{eq: consistent dist estimator} we have that 
    \begin{equation*}
        \widehat{P}_{\hat{u}_o | \hat{f}} (\hat{u}_o \mid \hat{f} (\bx_o) ) = P_{\hat{u}_o | \hat{f} } (\hat{u}_o \mid \hat{f} (\bx_o) ) + o_{\mathbf{Pr}} (1),
    \end{equation*}
    where 
    $$P_{\hat{u}_o | \hat{f} } (\hat{u}_o \mid \hat{f} (\bx_o) ) = V_o \sim  \mathrm{Unif} (0, \, 1),$$
    since convergence in probability implies convergence in distribution we arrive at~\eqref{eq: th 1 proof pivot to std unif}.
\end{proof}
{\bf Step III:} Show that MAPS has valid model-agnostic asymptotic conditional coverage.
\begin{equation}\label{eq: th 1 proof maps valid cond coverage}
    \underset{\bx_o}{\sup} \left \{ \mathbf{Pr} \left ( Y_o \notin \widehat{C}_{\mathrm{maps}} (\bx_o ) \, | \, \hat{f} (\bX_o) = \hat{f} (\bx_o) \right ) - \alpha \right \} = o_{\mathbf{Pr}}(1).
\end{equation}
\begin{proof}
    Let $\widehat{q}_{\hat{u} | \hat{f} } (\alpha) := \widehat{P}_{\hat{u}_o | \hat{f}}^{-1} (\alpha \mid \hat{f} (\bx_o) )$, which can be computed from samples generated by $P_{\hat{u}_o | \hat{f} }$. We proceed to compute coverage for $Y_o \mid \hat{f} (\bX_o) = \hat{f} (\bx_o)$,
    \begin{align}
        \mathbf{Pr} ( Y_o - \hat{y}_{\psi} (\bx_o ) &\in [\widehat{q}_{\hat{u} | \hat{f} } (\alpha / 2), \, \widehat{q}_{\hat{u} | \hat{f} } (1 - \alpha / 2) ] \mid \mathcal{F}_{\hat{f}} (\bx_o, \, \calb{\x}) ) \nonumber \\
        &= \mathbf{Pr} ( \hat{u}_o \in [\widehat{q}_{\hat{u} | \hat{f} } (\alpha / 2), \, \widehat{q}_{\hat{u} | \hat{f} } (1 - \alpha / 2) ] \mid \mathcal{F}_{\hat{f}} (\bx_o, \, \calb{\x}) ) \nonumber \\
        &= \mathbf{Pr} (\widehat{q}_{\hat{u} | \hat{f} } (\alpha / 2) \leq \hat{u}_o \leq \widehat{q}_{\hat{u} | \hat{f} } (1 - \alpha / 2)  \mid \mathcal{F}_{\hat{f}} (\bx_o, \, \calb{\x})), \label{eq: th 1 proof kde sampler coverage}
    \end{align}
    recall that the CDF estimator is a monotone function, thus,~\eqref{eq: th 1 proof kde sampler coverage} is equal to
    \begin{equation*}
        \mathbf{Pr} ( \widehat{P}_{\hat{u}_o | \hat{f}} (\widehat{q}_{\hat{u} | \hat{f} } (\alpha / 2)  \mid \hat{f} (\bx_o) ) \leq \widehat{V}_o \leq  \widehat{P}_{\hat{u}_o | \hat{f}} (\widehat{q}_{\hat{u} | \hat{f} } (1 - \alpha / 2)  \mid \hat{f} (\bx_o) ) \mid \mathcal{F}_{\hat{f}} (\bx_o, \, \calb{\x})).
    \end{equation*}
    By~\eqref{eq: th 1 proof pivot to std unif} we see that the pivots are independent of $\mathcal{F}_{\hat{f}} (\bx_o, \, \calb{\x})$, and
    \begin{equation}\label{eq: th 1 proof coverage to std unif}
        \mathrm{\eqref{eq: th 1 proof kde sampler coverage} } = \mathbf{Pr} ( \widehat{P}_{\hat{u}_o | \hat{f}} (\widehat{q}_{\hat{u} | \hat{f} } (\alpha / 2)  \mid \hat{f} (\bx_o) ) \leq \widehat{V}_o \leq \widehat{P}_{\hat{u}_o | \hat{f}} (\widehat{q}_{\hat{u} | \hat{f} } (1 - \alpha / 2)  \mid \hat{f} (\bx_o) )) +  o_{\mathbf{Pr}} (1).
    \end{equation}
    By combining~\eqref{eq: consistent dist estimator},~\eqref{eq: th 1 proof quantile convergence} and~\eqref{eq: th 1 proof pivot to std unif} we have that
    \begin{align}\label{eq: th 1 proof valid asymp coverage kde sampler}
        \mathrm{\eqref{eq: th 1 proof coverage to std unif} } &= \mathbf{Pr} \left ( P_{\hat{u}_o | \hat{f}} (\widehat{q}_{\hat{u} | \hat{f} } (\alpha / 2)  \mid \hat{f} (\bx_o) ) \leq V_o \leq P_{\hat{u}_o | \hat{f}} (\widehat{q}_{\hat{u} | \hat{f} } (1 - \alpha / 2)  \mid \hat{f} (\bx_o) ) \right ) +  o_{\mathbf{Pr}} (1) \nonumber \\
        &\overset{(i)}{=} \mathbf{Pr} ( P_{\hat{u}_o | \hat{f}} (q_{\hat{u} | \hat{f} } (\alpha / 2)  \mid \hat{f} (\bx_o) ) \leq V_o \leq P_{\hat{u}_o | \hat{f}} (q_{\hat{u} | \hat{f} } (1 - \alpha / 2)  \mid \hat{f} (\bx_o) )) + o_{\mathbf{Pr}} (1) \nonumber \\ 
        &\overset{(ii)}{=} \mathbf{Pr} (\alpha / 2 \leq V_o \leq 1 - \alpha / 2) + o_{\mathbf{Pr}} (1) \nonumber \\
        &= 1 - \alpha + o_{\mathbf{Pr}} (1).
    \end{align}
    where $(i)$ follows from~\eqref{eq: th 1 proof quantile convergence} and $(ii)$ from~\eqref{eq: th 1 proof pivot to std unif}. By~\eqref{eq: consistent dist estimator} and~\eqref{eq: th 1 proof quantile convergence} we have that samples generated by $\widehat{P}_{\hat{u}_o | \hat{f}}$, either directly or by inversion, converge to $P_{\hat{u}_o | \hat{f}}$, that is, the iid bootstrap samples satisfy
    \begin{equation}\label{eq: th 1 proof kde sampler equivalence}
        \hat{u}_{i}^* \overset{\boldsymbol{\mathrm{d}}}{\longrightarrow} \hat{u}_{\calb{i}} \, \, \, \, \mathrm{and} \, \, \, \, \, \widehat{U}_{o}^* \overset{\boldsymbol{\mathrm{d}}}{\longrightarrow} \hat{u}_o.
    \end{equation}
    Thus, we have that the bootstrap lifted residual satisfies
    \begin{align}
        \hat{u}^*_o &= Y_o^* - \hat{y}_{\psi}^*(\bx_o) \nonumber \\
        &= \hat{y}_{\psi} (\bx_o) - \hat{y}_{\psi}^*(\bx_o) + \widehat{U}_{o}^* \nonumber \\
        &\overset{(i)}{=} \widehat{U}_{o}^* + o_{\mathbf{Pr}} (1) \nonumber
    \end{align}
    where $(i)$ follows from assumption~\eqref{eq: consistent psi etimator}. Then, by~\eqref{eq: th 1 proof kde sampler equivalence} we arrive at 
    $$\hat{u}^*_o \overset{\boldsymbol{\mathrm{d}}}{\longrightarrow} \hat{u}_o + o_{\mathbf{Pr}} (1).$$ 
    Thus, by~\eqref{eq: th 1 proof quantile convergence} the quantiles of bootstrap samples are consistent, that is, for all $\alpha \in (0, \, 1)$ we have
    \begin{equation*}
        \widehat{q}_{\hat{u}_o^* | \hat{f} } (\alpha) = \widehat{P}_{\hat{u}_o | \hat{f}}^{-1} (\alpha \mid \hat{f} (\bx_o) ) + o_{\mathbf{Pr}} (1).
    \end{equation*}
    By combining the above with~\eqref{eq: th 1 proof coverage to std unif} for all $\bx_o \in \mathcal{X}$ and calibration sets we have that
    \begin{equation*}
         \mathbf{Pr} ( Y_o \in \widehat{C}_{\mathrm{maps}} (\bx_o ) \mid \mathcal{F}_{\hat{f}} (\bx_o, \, \calb{\x}) ) = \mathbf{Pr} \left ( \alpha / 2 \leq V_o \leq 1 - \alpha / 2 \right ) + o_{\mathbf{Pr}} (1).
    \end{equation*}
    Hence, we see that
    \begin{align*}
         \underset{\bx_o}{\sup} \{ \mathbf{Pr} ( Y_o \notin \widehat{C}_{\mathrm{maps}} (\bx_o ) \, &| \, \hat{f} (\bX_o) = \hat{f} (\bx_o) ) - \alpha \} \\
         &= \underset{\bx_o}{\sup} \left \{ 1 - \mathbf{Pr} \left ( \alpha / 2 \leq V_o \leq 1 - \alpha / 2 \right ) - \alpha \right \} + o_{\mathbf{Pr}} (1) \\
         &= (1 - 1) + (\alpha - \alpha) + o_{\mathbf{Pr}} (1) = o_{\mathbf{Pr}} (1),
    \end{align*}
    whence we arrive at~\eqref{eq: th 1 proof maps valid cond coverage}.
\end{proof}
{\bf Step IV:} Show that if~\eqref{eq: asymp: linear lpm} holds, then~\eqref{eq: consistent psi etimator} also holds and the intervals are bootstrap consistent. That is, for all $\bx_o$ and any $\hat{f}$, Algorithm~\ref{alg: MAPS add errors} is {\em Kolmogorov-Smirnov} consistent; see Lemma 23.3 in~\cite{asym_stats}.
\begin{equation}\label{eq: th 1 proof bootstrap residual consistency}
    \sup_{t \in \mathbb{R}} \left \{ \bigg | \mathbf{Pr} \left ( \hat{u}_o^* \leq t \mid \mathcal{F}_{\hat{f}} (\bx_o, \, \mathbf{x}_{\calb{n}} ) \right ) - \mathbf{Pr} \left ( u_o \leq t \mid \hat{f} (\bX_o) = \hat{f} (\bx_o) \right ) \bigg | \right \} =  o_{\mathbf{Pr}} (1).
\end{equation}
\begin{proof}
    By assumption~\eqref{eq: asymp: linear lpm} and Lemma~\ref{lem: stability of the lpm}, and~\eqref{eq: lem 2 proof asymp norm} we have for all $\x_o$ and any $\hat{f}$ that
    \begin{equation}\label{eq: th 1 proof lifted est error to zero}
        \hat{y}_{\psi} (\x_o) = y_{\psi} (x_o) + o_{\mathbf{Pr}} (1 / \sqrt{a_{\calb{n}}}).
    \end{equation}
    Let $\widehat{\Delta}_{\psi} (\bx_o) := \hat{y}_{\psi} (\x_o) - y_{\psi} (x_o)$ denote lifted model estimation error. Decompose the lifted residual into estimation error and irreducible error by writing it as the sum
    \begin{equation}\label{eq: th 1 proof lifted residual decomposition}
        \hat{u}_o = \widehat{\Delta}_{\psi} (\bx_o) + u_o.
    \end{equation}
    We proceed to show that the lifted residuals satisfy the following properties.
    \begin{enumerate}
        \item By~\eqref{eq: th 1 proof lifted est error to zero} and~\eqref{eq: th 1 proof lifted residual decomposition} we have that calibration and out-sample data satisfy
        \begin{equation}\label{eq: th 1 proof pert error to zero lifted}
          \sqrt{a_{\calb{n}}} \, \, \widehat{\Delta}_{\psi} (\bx_{\calb{i}}) = o_{\mathbf{Pr}} (1), \, \, \, \hat{u}_{\calb{i}} \overset{\boldsymbol{\mathrm{d}}}{\longrightarrow} u_{\calb{i}}, \, \, \, \, \,\, \sqrt{a_{\calb{n}}} \, \, \widehat{\Delta}_{\psi} (\bx_o) = o_{\mathbf{Pr}} (1), \, \, \, \hat{u}_o \overset{\boldsymbol{\mathrm{d}}}{\longrightarrow} u_o.
        \end{equation}
        \item By~\eqref{eq: th 1 proof kde sampler equivalence} and~\eqref{eq: th 1 proof pert error to zero lifted} we have that the bootstrap errors satisfy
        \begin{equation}\label{eq: th 1 proof kde sampler pert convergence}
            \hat{u}_{i}^* \overset{\boldsymbol{\mathrm{d}}}{\longrightarrow} u_{\calb{i}} \, \, \, \, \mathrm{and } \, \, \, \, \, \widehat{U}_{o}^* \overset{\boldsymbol{\mathrm{d}}}{\longrightarrow} u_o.
        \end{equation}
        \item By~\eqref{eq: th 1 proof lifted est error to zero} and~\eqref{eq: th 1 proof kde sampler pert convergence}, given $\mathcal{F}_{\hat{f}} (\bx_o, \, \mathbf{x}_{\calb{n}} )$, the bootstrap responses satisfy
        \begin{equation}\label{eq: th 1 proof response sampler pert convergence}
            Y_{i}^* \overset{\boldsymbol{\mathrm{d}}}{\longrightarrow} y_{\psi} (\x_{\calb{i}}) +  u_{\calb{i}} + o_{\mathbf{Pr}} (1 / \sqrt{a_{\calb{n}}}), \, \, \, \, Y_{o}^* \overset{\boldsymbol{\mathrm{d}}}{\longrightarrow} y_{\psi} (\x_o) + u_o +  o_{\mathbf{Pr}} (1 / \sqrt{a_{\calb{n}}}).
        \end{equation}
        \item Let $\widehat{\Delta}_{\psi}^* (\bx_o) := \hat{y}^*_{\psi} (\x_o) - \hat{y}_{\psi} (x_o)$ denote lifted model estimation error in the bootstrap sampling mechanism. By combining~\eqref{eq: asymp: linear lpm}, Lemma~\ref{lem: stability of the lpm} and~\eqref{eq: lem 2 proof asymp norm} we see that 
        \begin{equation}\label{eq: th 1 proof asymp norm pert bootstrap}
            \sqrt{a_{\calb{n}}} \, \, \widehat{\Delta}_{\psi}^* (\bx_o) \overset{\boldsymbol{\mathrm{d}}}{\longrightarrow} \mathcal{N} (0, \, \sigma_{\hat{y}}^2 (\bx_o) ) \, \, \, \, \mathrm{and } \, \, \, \, \,  \sqrt{a_{\calb{n}}} \, \, \widehat{\Delta}_{\psi} (\bx_o) \overset{\boldsymbol{\mathrm{d}}}{\longrightarrow} \mathcal{N} (0, \, \sigma_{\hat{y}}^2 (\bx_o) ).
        \end{equation}
        \item By combining~\eqref{eq: th 1 proof kde sampler pert convergence},~\eqref{eq: th 1 proof asymp norm pert bootstrap} and Lemma 2.11 in~\cite{asym_stats} we have that
        \begin{align}
            &\sup_{t \in \mathbb{R}} \left \{ \left | \mathbf{Pr} \left ( \sqrt{a_{\calb{n}}} \widehat{\Delta}_{\psi}^* (\bx_o) \leq t \mid \mathcal{F}_{\hat{f}} (\bx_o ) \right ) - \mathbf{Pr} \left ( \sqrt{a_{\calb{n}}}  \widehat{\Delta}_{\psi} (\bx_o) \leq t \mid \mathcal{F}_{\hat{f}} (\bx_o ) \right )  \right | \right \} =  o_{\mathbf{Pr}} (1),  \label{eq: th 1 proof est error boot consistnent} \\
             &\sup_{t \in \mathbb{R}} \left \{ \left | \mathbf{Pr} \left ( \widehat{U}^*_o \leq t \mid \mathcal{F}_{\hat{f}} (\bx_o, \, \mathbf{x}_{\calb{n}} ) \right ) - \mathbf{Pr} \left ( u_o \leq t \mid \hat{f} (\X_o ) = \hat{f} (\x_o) \right )  \right | \right \} =  o_{\mathbf{Pr}} (1),  \label{eq: th 1 proof pred error boot consistnent}
        \end{align}
        where $\mathcal{F}_{\hat{f}} (\bx_o) \equiv \mathcal{F}_{\hat{f}} (\bx_o, \, \mathbf{x}_{\calb{n}} )$.
    \end{enumerate}
    \par
    Recall that $\hat{u}^*_o = Y^*_o - \hat{y}_{\psi}^* (\bx_o)$, which we can write as $\hat{u}^*_o = - \widehat{\Delta}_{\psi}^* (\bx_o) + \widehat{U}^*_o$. By Slutsky's Lemma (Lemma 2.8 in~\citealp{asym_stats}) and~\eqref{eq: th 1 proof est error boot consistnent} we have that
    \begin{equation}\label{eq: th 1 proof boot root to lifted error}
        \hat{u}^*_o = \widehat{U}^*_o + o_{\mathbf{Pr}} (1 / \sqrt{a_{\calb{n}}}).
    \end{equation}
    By combining~\eqref{eq: th 1 proof kde sampler pert convergence} and~\eqref{eq: th 1 proof boot root to lifted error}, we see that $\hat{u}^*_o  \overset{\boldsymbol{\mathrm{d}}}{\longrightarrow} u_o$. By noting that $u_o$ is independent of the calibration set and applying Lemma 2.11 in~\cite{asym_stats}, we arrive at~\eqref{eq: th 1 proof bootstrap residual consistency}, which finishes our proof of Theorem~\ref{th: aymp validity of MAPS}.
\end{proof}
{\bf Pertinent prediction intervals.} Conditions~\eqref{eq: th 1 proof pert error to zero lifted}--\eqref{eq: th 1 proof pred error boot consistnent} are model-agnostic pertinent, which generalise the ones proposed by~\cite{politis2015model, wang2021modelfreebootstrapconformalprediction}. By construction $u_o$ is independent of data in the calibration set and randomness in LPM estimation error solely comes from the calibration set. 
Hence, the distribution of $\hat{u}_o$ is the convolution of the distribution of $u_o$ and that of $\widehat{\Delta}_{\psi} (\bx_o)$, and the one of $\hat{u}^*_o$ is the convolution of the distribution of $\widehat{U}^*_o$ and that of $-\widehat{\Delta}_{\psi}^* (\bx_o)$.
\begin{assumptions}\label{assum: A3}
        There exist a $\tau, \, \kappa > 0$ and $n_0$ such that if $\calb{n} \geq n_0$, then for all $ t > \kappa$ and $\sigma \in (0, \tau)$, we have that $\widehat{P}_{\hat{u} | \hat{f}}, \, P_{\hat{u} | \hat{f} } \in \mathcal{P}_{\tau, \kappa}$, where $ \mathcal{P}_{\tau, \kappa}$ is the following family of distributions
            \begin{equation*}
                 \mathcal{P}_{\tau, \kappa} := \left \{ U \sim P  : \, \mathbf{Pr} \left ( | U | \leq  t \right ) > \mathbf{Pr} \left ( | U + T_{\sigma} | \leq t \right ), \, \, \mathrm{where} \, \, T_{\sigma} \sim \mathcal{N} (0, \sigma^2), \, \, t \in \mathbb{R} \right \}.
            \end{equation*}
\end{assumptions}
\par
\cite{wang2021modelfreebootstrapconformalprediction} introduced Assumption~\ref{assum: A3}, which says that a {\em smoothed} bootstrap algorithm (e.g., Algorithm~\ref{alg: MAPS add errors}) would be an appropriate approach, since it will have a faster concentration rate; see~\cite{bootstrap_pred_intervals}. This condition is satisfied by a large family of distributions. Intuitively, it says that the density functions have positive mass across all quantiles and decay in a smooth enough manner, i.e., the tails of the distribution do not decrease too roughly, and can be upper-bounded by a smooth tail, by adding a Gaussian error. By~\eqref{eq: th 1 proof asymp norm pert bootstrap}, we see that if Assumption~\ref{assum: A3} holds, then bootstrap prediction intervals are longer for $\calb{n} \geq n_0$ (the asymptotic distribution of $\hat{u}^*_o$ is a convolution with heavier tails). Hence, MAPS prediction intervals should provide better finite-sample conditional coverage approximations than DCP~\citep{dcp} or KDE sampling based ones for this family of distributions~\citep[Theorem 5.4 in][]{wang2021modelfreebootstrapconformalprediction}.
\subsection{Proof of Corollary~\ref{cor: binary class interval}}
Our argument can be thought of as a {\em delta method}~\citep[Chapter 3 in][]{asym_stats} tailored for pointwise ``prediction" rather than estimation.
\begin{proof}
    Recall that $\varphi (t) = (1 + 1 / e^t)^{-1}$ is the \texttt{sigmoid} function. Thus, we have that
    \begin{align}
        &\varphi' (t) = (1 - \varphi (t) ) \, \varphi (t), \label{eq: cor 1 proof sigmoid 1-diff} \\
        &\varphi'' (t) = 2 \, (1 / 2 - \varphi (t) ) \, (1 - \varphi (t) ) \, \varphi (t) \label{eq: cor 1 proof sig 2-diff}.
    \end{align}
    By~\eqref{eq: cor 1 proof sigmoid 1-diff} and~\eqref{eq: cor 1 proof sig 2-diff}, we see that $| \varphi' (t) | \leq 1$ and $ | \varphi'' (t) | \leq 1$. By Taylor's Theorem, if $h \to 0$, then
    \begin{equation}\label{eq: cor 1 proof taylor error}
        \mathrm{e} (h) := \varphi (\hat{y}_{\psi} (\x_o )+ h) - \varphi (\hat{y}_{\psi} (\x_o)) - h \, \varphi' (\hat{y}_{\psi} (\x_o)) = \varphi'' (\hat{y}_{\psi} (\x_o)) h^2 / 2 + o (|h|^3),
    \end{equation}
    since $| \varphi'' (\hat{y}_{\psi} (\x_o)) | \leq 1$ with probability one, we have that $| \mathrm{e} (h) | = o(h^2) \to 0 $. 
    \par
    Recall that $R_o^* = \hat{y}_{\psi}^* (\x_o) - \hat{y}_{\psi} (\x_o)$, and that by~\eqref{eq: consistent psi etimator} we have that $R_o^* = o_{\mathbf{Pr}} (1)$. By Lemma 2.12 in~\cite{asym_stats}, we can swap $h$ for $R_o^*$ in~\eqref{eq: cor 1 proof taylor error} to get $| \mathrm{e} (R_o^*) | = o (| R_o^*|^2)$, and by the continuous mapping theorem, $ \mathrm{e} (R_o^*) = o_{\mathbf{Pr}} (1)$. Since $| \varphi' (\hat{y}_{\psi} (\x_o)) \, R_o^* | \leq | R_o^*| $ with probability one, we have by Slutsky's Lemma that $\varphi' (\hat{y}_{\psi} (\x_o)) \, R_o^* = o_{\mathbf{Pr}} (1)$ and arrive at
    \begin{equation}\label{eq: cor 1 proof taylor dist coonvergence}
        \varphi (\hat{y}_{\psi} (\x_o ) + R_o^*) \overset{\boldsymbol{\mathrm{d}}}{\longrightarrow} \varphi (\hat{y}_{\psi} (\x_o)) + o_{\mathbf{Pr}} (1).
    \end{equation}
    \par
    By applying Lemma 2.11 in~\cite{asym_stats} to~\eqref{eq: cor 1 proof taylor dist coonvergence}, where $\mathcal{F}_{\hat{f}} (\bx_o) \equiv \mathcal{F}_{\hat{f}} (\bx_o, \, \mathbf{x}_{\calb{n}} )$,
    \begin{equation}\label{eq: cor 1 proof kol-smir consistent}
        \sup_{t \in \mathbb{R}} \left \{ \bigg | \mathbf{Pr} \left ( \varphi (\hat{y}_{\psi} (\x_o ) + R_o^*) \leq t \mid \mathcal{F}_{\hat{f}} (\bx_o ) \right ) - \mathbf{Pr} \left ( \varphi (\hat{y}_{\psi} (\x_o))  \leq t \mid \mathcal{F}_{\hat{f}} (\bx_o ) \right ) \bigg | \right \} = o_{\mathbf{Pr}} (1).
    \end{equation}
    \par
    Now, recall that Algorithm~\ref{alg: maps for binary class} produces a bootstrap sample of $\hat{y}_{\psi} (\x_o ) + R_o^*$, and computes the intervals in~\eqref{eq: MAPS binary class}--\eqref{eq: MAPS binary class opt} using the transform $\varphi (\hat{y}_{\psi} (\x_o ) + R_o^*)$. By~\eqref{eq: cor 1 proof kol-smir consistent} and Lemma 23.3 in~\cite{asym_stats} these are {\em Kolmogorov-Smirnov} consistent. We then have one of the following cases. 
    \begin{enumerate}
        \item {\bf Ambiguous-in-distribution.} The random variable $\hat{y}_{\psi} (\x_o )$ has no concentration point (i.e., it does not converge to a fixed point, in probability). That is, the prediction intervals do not collapse to a fixed calibration probability.
        \item {\bf Biased lifted logit.} The random variable $\hat{y}_{\psi} (\x_o )$ is a consistent estimator of a fixed point $\overline{y}_{\psi} (\x_o)$. That is, $\hat{y}_{\psi} (\x_o) = \overline{y}_{\psi} (\x_o) + o_{\mathbf{Pr}} (1)$. Then, by the continuous mapping theorem, we have that $\varphi (\hat{y}_{\psi} (\x_o) ) = \varphi ( \overline{y}_{\psi} (\x_o)) + o_{\mathbf{Pr}} (1)$, which combined with~\eqref{eq: cor 1 proof taylor dist coonvergence} and~\eqref{eq: cor 1 proof kol-smir consistent} gives
        \begin{equation}\label{eq: cor 1 proof concentration point}
            \varphi (\hat{y}_{\psi} (\x_o ) + R_o^*) = \varphi ( \overline{y}_{\psi} (\x_o)) + o_{\mathbf{Pr}} (1).
        \end{equation}
        \item {\bf Asymp. linear lifted logit.} The random variable $\hat{y}_{\psi} (\x_o )$ satisfies~\eqref{eq: asymp: linear lpm}. Then, by combining Lemma~\ref{lem: stability of the lpm} and the delta method~\citep[Chapter 3 in][]{asym_stats}, we arrive at 
        \begin{equation*}
            \sqrt{a_{\calb{n}}} \{ \varphi (\hat{y}_{\psi} (\x_o )) - \varphi (y_{\psi} (\x_o)) \} \overset{\boldsymbol{\mathrm{d}}}{\longrightarrow} \mathcal{N} (0, \, \varphi' (y_{\psi} (\x_o)) \, \sigma_{\hat{y}}^2 (\bx_o) \, \varphi' (y_{\psi} (\x_o))),
        \end{equation*}
        which upon rearranging, recalling that $\sigma_{\hat{y}}^2 (\bx_o) / a_{\calb{n}} \to 0$ and $|\varphi' (y_{\psi} (\x_o))| \leq 1$, and combining with~\eqref{eq: cor 1 proof taylor dist coonvergence}--\eqref{eq: cor 1 proof kol-smir consistent} gives
        \begin{equation}\label{eq: cor 1 proof collapse at logit for asymp linear LPM}
            \varphi (\hat{y}_{\psi} (\x_o ) + R_o^*) = \varphi ( y_{\psi} (\x_o)) + o_{\mathbf{Pr}} (1 /  \sqrt{a_{\calb{n}}}),
        \end{equation}
        and applying Egorov's Theorem~\citep{bartle1982elements} to~\eqref{eq: cor 1 proof collapse at logit for asymp linear LPM} gives~\eqref{eq: binary class maps asymp validity}.
    \end{enumerate}
    Combining the three cases above finishes our proof of Corollary~\ref{cor: binary class interval}.
\end{proof}

\vspace{-10 pt}
\subsection{Proof of Lemma~\ref{lem: coverage comparison and optimal f-homos}}
\label{app: proof of lemma 3}
{\bf Step I:} Show that the given $f$, the probability is weighted along the contour, that is, for all $y \in \mathbb{R}$ such that $f^{-1} (y) \neq \varnothing$, given $f(\X) = f(\x) = y$, we have
\begin{equation}\label{eq: lem 3 proof contour cond prob}
    \mathbf{Pr} \big ( \epsilon \leq t \mid f(\X) = f(\x) \big ) = \frac{ 1 }{ \int_{f^{-1} (y)} d P_{\X} } \int_{f^{-1} (y)} P_{\epsilon | \x} (t) \, d P_{\X}.
\end{equation}
\begin{proof}
    Let $P_{\epsilon | f} (t)$ denote the conditional distribution of $\epsilon \mid f(\X) = f(\x)$, where $f(\x) = y$, $f^{-1} (y) \neq \varnothing$ and $f^{-1} (y) \subseteq \mathcal{X}$. Then, we have that
    \begin{align*}
        P_{\epsilon | f} (t) &=  \mathbf{Pr} \left ( \epsilon \leq t \mid f(\X) = f(\x) \right ) \\
        &=  \mathbf{Pr} \left ( \epsilon \leq t \mid \X \in f^{-1} (y), \, f(\x) = y \right ),
    \end{align*}
    by the law of total probability, we then have
    \begin{align*}
        P_{\epsilon | f} (t) &= \int_{ f^{-1} (y)} \mathbf{Pr} \left ( \epsilon \leq t \mid \X = \x \right ) \, p_{\X} (\x \mid \X \in f^{-1} (y) ) \, d \boldsymbol{\lambda}\\
        &= \left ( \int_{ f^{-1} (y)} p_{\X} (\x ) \, d \boldsymbol{\lambda} \right )^{-1} \, \int_{ f^{-1} (y)} \mathbf{Pr} \left ( \epsilon \leq t \mid \X = \x \right ) \, p_{\X} (\x ) \, d \boldsymbol{\lambda},
    \end{align*}
    where $\boldsymbol{\lambda}$ denotes the $d$-dimensional Lebesgue measure. By rewriting the above as Riemann-Stieltjes integrals, we arrive at~\eqref{eq: lem 3 proof contour cond prob}.
\end{proof}
{\bf Step II:} Show that the conditional probability bounds in~\eqref{eq: coverage comparison global} hold.
\begin{proof}
    Since $P_{\epsilon | \x}$ is continuous for every $\x \in \mathcal{X}$, the quantiles are well-defined and {\em finite} for each $\alpha \in (0, \, 1)$. Hence, for each $y \in \mathbb{R}$ such that $f^{-1} (y) \neq \varnothing$, given $f(\X) = y$, we can define
    \begin{align}
        q_{\alpha}^{-} (y) &:= \inf \{ q_{\epsilon | \x} (\alpha) : \, \x \in f^{-1} (y) \} > -\infty, \label{eq: lem 3 proof inf quantile def} \\
        q_{\alpha}^{+} (y) &:= \sup \{ q_{\epsilon | \x} (\alpha) : \, \x \in f^{-1} (y) \} < +\infty. \label{eq: lem 3 proof sup quantile def}
    \end{align}
    \par
    Then, since $P_{\epsilon | \x}$ is monotone increasing for each $\x$, and~\eqref{eq: lem 3 proof inf quantile def}--\eqref{eq: lem 3 proof sup quantile def} are finite, there exist
    \begin{align}
        \alpha^{-} (y) &:= \min \{ P_{\epsilon | \x} (t) : \, t \in [q_{\alpha}^{-} (y), \, q_{\alpha}^{+} (y)] \} > 0, \label{eq: lem 3 proof min alpha} \\
        \alpha^{+} (y) &:= \max \{ P_{\epsilon | \x} (t) : \, t \in [q_{\alpha}^{-} (y), \, q_{\alpha}^{+} (y)] \} < 1. \label{eq: lem 3 proof max alpha}
    \end{align}
    \par
    Thus, for all $\alpha \in (0, \, 1)$ and $\x \in f^{-1} (y)$ we have that $ q_{\alpha}^{-} (y) \leq q_{\epsilon | \x} (\alpha) \leq q_{\alpha}^{+} (y)$, and since $P_{\epsilon | \x}$ is monotone non-decreasing, we see that for all $\x \in f^{-1} (y)$
    \begin{equation}\label{eq: lem 3 proof alpha bound for in contour}
        \alpha^{-} (y) \leq P_{\epsilon | \x } ( q_{\alpha}^{-} (y)) \leq P_{\epsilon | \x } ( q_{\epsilon | \x} (\alpha) ) = \alpha \leq P_{\epsilon | \x } ( q_{\alpha}^{+} (y)) \leq \alpha^{+} (y),
    \end{equation}
    hence, for all $t \in [q_{\alpha}^{-} (y), \, q_{\alpha}^{+} (y)]$ and $\x \in f^{-1} (y)$ we have that
    \begin{equation}\label{eq: lem 3 proof alphas bound for t}
         \alpha^{-} (y) \leq P_{\epsilon | \x } ( t) \leq  \alpha^{+} (y).
    \end{equation}
    By~\eqref{eq: lem 3 proof alphas bound for t}, we arrive at 
    \begin{equation}
         \frac{ \alpha^{-} (y) \, \int_{f^{-1} (y)} d P_{\X} }{ \int_{f^{-1} (y)} d P_{\X} } \leq  \frac{ \int_{f^{-1} (y)} P_{\epsilon | \x} (t) \, d P_{\X} }{ \int_{f^{-1} (y)} d P_{\X} } \leq \frac{ \alpha^{+} (y) \, \int_{f^{-1} (y)} d P_{\X} }{ \int_{f^{-1} (y)} d P_{\X} },
    \end{equation}
    which by~\eqref{eq: lem 3 proof contour cond prob} gives
    \begin{equation}\label{eq: lem 3 proof contour prob bound}
         \alpha^{-} (y) \leq \mathbf{Pr} \big ( \epsilon \leq t \mid f(\X) = f(\x) \big ) \leq  \alpha^{+} (y).
    \end{equation}
    So, by the intermediate value theorem and~\eqref{eq: lem 3 proof contour cond prob}--\eqref{eq: lem 3 proof contour prob bound}, $q_{\epsilon | f} (\alpha) \in [q_{\alpha}^{-} (y), \, q_{\alpha}^{+} (y)]$ for all $\alpha \in (0, \, 1)$. Hence, for all $\x \in f^{-1} (y)$ and $\alpha \in (0, \, 1)$, we arrive at
    \begin{equation}\label{eq: lem 3 proof bound for f-contour dist}
        \alpha^{-} (y) \leq P_{\epsilon | \x } ( q_{\alpha}^{-} (y)) \leq P_{\epsilon | f } ( q_{\epsilon | f} (\alpha) ) \leq P_{\epsilon | \x } ( q_{\alpha}^{+} (y)) \leq \alpha^{+} (y).
    \end{equation}
    Recall that for any $\alpha \in (0, \, 1)$ there exists an $h \in [-\alpha / 2, \, \alpha / 2]$ such that
    \begin{equation}\label{eq: lem 3 proof f-cont ideal interval}
        C_{f\text{-}\mathrm{ideal}} \left ( \boldsymbol{x}_o \right ) = \big [ f(\boldsymbol{x}_o) + q_{\epsilon_o | f} ( \alpha / 2 + h), \, f(\boldsymbol{x}_o) + q_{\epsilon_o | f} (1 -  \alpha / 2 + h) \big ].
    \end{equation}
    By~\eqref{eq: lem 3 proof bound for f-contour dist}, for the shape-adjusted levels, we have
    \begin{align*}
         \alpha^{-}_h (y) / 2 + h \leq P_{\epsilon_o | \x_o} \big  ( q_{\epsilon_o | f} ( \alpha / 2 + h) \big ) \leq \alpha^{+}_h (y) / 2 + h, \\
         \alpha^{-}_h (y) / 2 - h \leq 1 - P_{\epsilon_o | \x_o} \big  ( q_{\epsilon_o | f} (1 -  \alpha / 2 + h) \big ) \leq \alpha^{+}_h (y) / 2 - h,
    \end{align*}
    which combined with~\eqref{eq: lem 3 proof f-cont ideal interval} gives~\eqref{eq: coverage comparison global}, that is,
    \begin{equation*}
         \alpha^{-} (y) \leq \mathbf{Pr} \left ( Y_o \notin  C_{f\text{-}\mathrm{ideal}} \left ( \boldsymbol{x}_o \right ) \mid \X_o = \x_o \right ) \leq  \alpha^{+} (y).
    \end{equation*}
    Finally, if~\eqref{eq: f-homos assump} holds, then for all $\x_o \in f^{-1} (y)$ and $\alpha \in (0, \, 1)$, we have that $P_{\epsilon_o | f} \equiv P_{\epsilon_o | \x_o}$, which by~\eqref{eq: lem 3 proof contour cond prob} and the above gives $q_{\epsilon_o | f} ( \alpha) = q_{\epsilon_o | \x_o} ( \alpha)$, $\alpha^{-}_h (y) = \alpha_h = \alpha^{+}_h (y)$ and~\eqref{eq: optimal interval for f-homos}.
\end{proof}
\subsection{Proof of Theorem~\ref{th: dist-free consistent model properties}}
\label{supp: proof theorem three}
{\bf Step I:} Show the results in~\eqref{eq: maps consistency} hold. The LPM either inherits consistency from $\hat{f}$ or kills it.
\begin{proof}
    By Egorov's Theorem and assumption, we have that $\|\hat{f} - f \|_{\infty} = o_{\mathbf{Pr}} (1)$. By construction, we have that $\mathbb{E} \, [ u_o \mid \hat{f} (\X_o) = \hat{f} (\x_o) ] = 0$ for all $\x_o \in \mathcal{X}$. Hence, we must have that
    \begin{equation*}
        f (\x_o) = \mathbb{E} \, [ \psi (\hat{f} (\x_o)) \mid \hat{f} (\X_o) = \hat{f} (\x_o)],
    \end{equation*}
    and since $\psi \circ \hat{f}$ minimises MSPE given $\hat{f} (\X_o) = \hat{f} (\x_o)$, by the tower property and monotonicity of expectation, where the expectation is taken using $P_{\X}$, we also have that
    \begin{equation*}
        \|\psi \circ \hat{f} - f \|_{L_2}^2 = \mathbb{E} \, [ \big ( f(\X_o) - \psi( \hat{f} (\X_o)) \big )^2] \leq \mathbb{E} \, [ \big ( f(\X_o) - \hat{f} (\X_o) \big )^2] = \|\hat{f} - f \|_{L_2}^2,
    \end{equation*}
     and since $\|\hat{f} - f \|_{L_2} = o_{\mathbf{Pr}} (1)$, we arrive at $\|\psi \circ \hat{f} - f \|_{L_2} = o_{\mathbf{Pr}} (1)$, which, by Egorov's theorem, gives $\|\psi \circ \hat{f} - f \|_{\infty} = o_{\mathbf{Pr}} (1)$. By~\eqref{eq: asymp: linear lpm} and Lemma~\ref{lem: stability of the lpm}, we have that $|\hat{y}_{\psi} (\x_o) - y_{\psi} (\x_o)| = o_{\mathbf{Pr}} (1)$ for all $\x_o \in \mathcal{X}$, which, by Egorov's theorem, gives $\|\hat{y}_{\psi} - y_{\psi} \|_{\infty} = o_{\mathbf{Pr}} (1)$, and
     \begin{align}
         \| \hat{y}_{\psi} - f \|_{\infty} &=  \| \hat{y}_{\psi} - y_{\psi} + y_{\psi} - f \|_{\infty} \nonumber \\
         &\overset{(i)}{\leq} \| \hat{y}_{\psi} - y_{\psi} \|_{\infty} + \| y_{\psi} - f \|_{\infty} \nonumber \\
         &=  \| \hat{y}_{\psi} - y_{\psi} \|_{\infty} + \|\psi \circ \hat{f} - f \|_{\infty} \nonumber \\
         &\overset{(ii)}{=} o_{\mathbf{Pr}} (1). \label{eq: th 2 proof consistent for f LPM}
     \end{align}
     where $(i)$ follows from the triangle inequality and $(ii)$ from the two results above.
\end{proof}
{\bf Step II:} Show that MAPS produces consistent $f$-contour homoscedastic residuals, show~\eqref{eq: maps f-homos optimal}. 
\begin{proof}
    By Theorem~\ref{th: aymp validity of MAPS} and~\eqref{eq: th 1 proof bootstrap residual consistency}, we have that $\| P^*_{\hat{u} | \hat{f} } - P_{u_o | \hat{f}} \|_{\infty} = o_{\mathbf{Pr}} (1)$ for all $\x_o \in \mathcal{X}$. By the continuous mapping theorem, we have
    \begin{equation}\label{eq: th 2 proof dist convergence}
        \mathbf{Pr} \big ( \epsilon_o \leq t \mid \hat{f} (\X_o) = \hat{f} (\x_o) \big ) = \mathbf{Pr} \big ( \epsilon_o \leq t \mid f (\X_o) = f (\x_o) \big ) + o_{\prob} (1),
    \end{equation}
    and since $\|\psi \circ \hat{f} - f \|_{\infty} = o_{\mathbf{Pr}} (1)$, the lifted error is $u_o = \epsilon_o + o_{\mathbf{Pr}} (1)$, which combined with~\eqref{eq: th 2 proof dist convergence} and Lemma 2.11 in~\cite{asym_stats}, gives
    \begin{equation}\label{eq: th 2 proof u dist to epsilon dist}
        \| P_{u_o | \hat{f} } - P_{\epsilon_o | f} \|_{\infty} = o_{\prob} (1).
    \end{equation}
    We decompose the target into two parts as follows
    \begin{align}
         \| P^*_{\hat{u}_o | \hat{f} } - P_{\epsilon_o | f} \|_{\infty} &=  \| P^*_{\hat{u}_o | \hat{f} } - P_{u_o | \hat{f} } + P_{u_o | \hat{f} } - P_{\epsilon_o | f} \|_{\infty} \nonumber \\
         &\overset{(i)}{\leq} \| P^*_{\hat{u}_o | \hat{f} } - P_{u_o | \hat{f} } \|_{\infty} + \| P_{u_o | \hat{f} } - P_{\epsilon_o | f} \|_{\infty} \nonumber \\
         &\overset{(ii)}{=} o_{\mathbf{Pr}} (1), \label{eq: th 2 proof uhat to epsilon in dist}
    \end{align}
    where $(i)$ follows from the triangle inequality, and $(ii)$ from combining~\eqref{eq: th 2 proof u dist to epsilon dist} and~\eqref{eq: th 1 proof bootstrap residual consistency}. Hence, we arrive at~\eqref{eq: maps f-homos optimal}.
\end{proof}
{\bf Step III:} Show that~\eqref{eq: maps optimality} holds, i.e., MAPS is consistent for $f$-contour homnoscedastic distributions.
\begin{proof}
    Recall that the endpoints of~\eqref{eq: maps optimal interval dist-free} are
\begin{align*}
    \hat{c}_1 (\bx_o) &= \widehat{\psi} (\hat{f} (\bx_o)) + \widehat{q}_{\hat{u}^*} (\alpha / 2 + \hat{h}^* ), \\
     \hat{c}_2 (\bx_o) &= \widehat{\psi} (\hat{f} (\bx_o)) + \widehat{q}_{\hat{u}^*} (1 - \alpha / 2 + \hat{h}^*).
\end{align*}
So, by combining~\eqref{eq: th 2 proof consistent for f LPM} and~\eqref{eq: th 2 proof uhat to epsilon in dist}, we have that
\begin{align*}
     \hat{c}_1 (\bx_o) &= f(\x_o) + q_{\epsilon_o | f} (\alpha /2 + h) + o_{\mathbf{Pr}} (1), \\
     \hat{c}_2 (\bx_o) &= f(\x_o) + q_{\epsilon_o | f} (1 - \alpha /2 + h) + o_{\mathbf{Pr}} (1).
\end{align*}
By applying Lemma~\ref{lem: coverage comparison and optimal f-homos} and~\eqref{eq: optimal interval for f-homos}, we get that
\begin{align*}
     \hat{c}_1 (\bx_o) &= f(\x_o) + q_{\epsilon_o | \x_o} (\alpha /2 + h) + o_{\mathbf{Pr}} (1), \\
     \hat{c}_2 (\bx_o) &= f(\x_o) + q_{\epsilon_o | \x_o} (1 - \alpha /2 + h) + o_{\mathbf{Pr}} (1),
\end{align*}
for all $\x_o \in \mathcal{X}$, which by Proposition~\ref{prop: ideal optimal interval} finishes our proof.
\end{proof}
\section{Proofs for the spline estimator}
\label{app: spline proofs}
{\bf Notation.} Set $\lambda \in (\calb{n}^{-4 / 5}, \, \calb{n}^{-3/2})$ and $\calb{N} \in [\calb{n}^{1/5}, \, \calb{n}^{4/5} ]$, we write $a_m \lesssim b_m$ if there exists a constant $K > 0$ such that $a_m \leq K \, b_m$ for all sufficiently large $m \in \mathbb{N}$. Throughout this Appendix the variables $s, \, t \in  [y_{\min}, \, y_{\max}]$, uppercase letters, say $\X$, denote random variables and lowercase ones, say $\x$, realizations (as in the rest of the paper). The fitted spline is $\psibiashat$, $\psibar$ minimises the theoretical lifted loss, and $\psitrue$ is the unknown ``true" lifted regression function, $\kernel{t}{s}$ is the resulting Mercer kernel~\eqref{eq: spline proofs spline kernel}. 
\newline
We will show that if~\eqref{eq: spline family assumption} and~\eqref{eq: spline cond proper lifted support} are valid, then the spline estimator satisfies 
\begin{align}
    &\psibiashat (t) - \psibar (t) = \frac{1}{\calb{n}} \sum_{i = 1}^{\calb{n}} u_i \, \kernel{t}{\hat{y}_{i}} + o_{\prob} (1), \label{eq: spline proofs kernel expansion}\\
    &\sup_t | \psibar (t) - \psitrue (t) |^2 \lesssim \lambda + o_{\prob} (1), \label{eq: spline proofs bias term} \\
    &\sup_t \mathbb{E} \, [ | \psibiashat (t) - \psibar (t)|^2 ] = M_u \, O_{\prob}  \left ( \frac{1}{ \ncal \, \lambda^{1/4}} \right ), \label{eq: spline proofs variance term}
\end{align}
where $0 < \sigma_u^2 (\x) \leq M_u < \infty$ for all $\x \in \hat{f}^{-1} ( [y_{\min}, \, y_{\max}])$. %
\subsection{Preliminary Results}
Let $\mathcal{G}$ be the function family in~\eqref{eq: spline family assumption}, $\mathcal{W}^2$ the Sobolev space of order-two defined in $[y_{\min}, \, y_{\max}]$, and $\mathcal{S}_{N} := \{ g \in \mathcal{W}^2 : \, g (t) = \beta_0 + \beta_1 t + \sum_{j}^{\calb{N}} \eta_j \, (t - y_k)^3_+ \}$ be the natural cubic spline finite-dimensional function space~\citep[Chapter 5][]{statLearn} with $\calb{N}$ interior knots ${y_k}$. Define the penalty functional $J_2 (g, \, h) := \int_{y_{\min}}^{y_{\max}} g'' (t) \, h''(t) \, d t$ and the dot-product $\dotprod{g}{h}{2} := \int_{y_{\min}}^{y_{\max}} g(t) \, h(t) \, d t$ for $g, \, h \in \mathcal{W}^2$. We will use the following result~(Propositions 2.4--2.5 in~\citealp{penalised_splines_consistency}).
\begin{proposition}[\citealp{penalised_splines_consistency}]\label{prop: spline proof orthonormal expansion}
    The dot-product functional $\dotprod{g}{g}{2} = \| g \|_2^2$ is completely continuous with respect to $J_2$. Thus, there exists an orthonormal basis $\{ \phi_{\nu}: \, \nu = 0, \, 1, \dots, \}$ of $\mathcal{W}^2$, and of $\mathcal{G}$, such that for any $g, \, h \in \mathcal{G}$, we have
    \begin{align}
        &\dotprod{g}{h}{2} = \sum_{\nu} g_{\nu} h_{\nu}, \, \, \| g \|_2^2 = \sum_{\nu} g_{\nu}^2, \, \, J_2 (g, \, h) = \sum_{\nu} \rho_{\nu} g_{\nu} h_{\nu}; \label{eq: spline proof orth expansion and norm}\\
        &\| g \|_2^2 + \lambda \, J_2 (g) = (\dotprod{\cdot}{\cdot}{2} + \lambda J_2) (g) = \sum_{\nu} (1 + \rho_{\nu} ) g_{\nu}^2 < \infty, \label{eq: spline proof penalty diagonalised} \\
        &\sum_{\nu} (1 + \lambda \, \rho_{\nu})^{-1} = O (\lambda^{-1/4}), \label{eq: spline proofs bound for eigenvalue sum}
    \end{align}
    where $\rho_{\nu} \lesssim \nu^4 < \infty$ for all sufficiently large $\nu$ and $g_{\nu} = \dotprod{g}{\phi_{\nu}}{2}$
\end{proposition}
\par
By construction, we see that for all $g \in \mathcal{G}$ and $t \in [y_{\min}, \, y_{\max} ]$, we have $|g(t)| \leq K_{\mathcal{G}} + |\beta_0 + \beta_1 t |$, thus, there exist two constants such that
$$\sup_{g \in \mathcal{G} \setminus \{ 0 (\cdot) \} } \| g \|_{\infty} \leq A_1 < \infty, \, \, \, \, \mathrm{and} \, \,\, \, \inf_{g \in \mathcal{G} \setminus \{ 0 (\cdot) \} } \| g \|_{2} \geq A_2 > 0,$$
hence, motivated by~\cite{polynomial_spline_asymp}, we can define a measure of the complexity of $\mathcal{G}$ as 
\begin{equation}\label{eq: spline proofs sup-norm linf stability}
    \sup_{g \in \mathcal{G} \setminus \{ 0 (\cdot) \} } \left \{ \frac{\| g \|_{\infty}}{ \| g \|_2} \right \} = A_{\mathcal{G}} \leq A_1 / A_2 < \infty.
\end{equation}
Our proof makes heavy use of this $L_{\infty}$- and $L_2$-norm stability, which is shared by spline spaces, $\sup_{g \in \mathcal{S}_N \setminus \{ 0 (\cdot) \} } \left \{ \| g \|_{\infty} / \| g \|_2 \right \} \lesssim \calb{N^{1/2}}$~\citep[Proposition 2.2][]{penalised_splines_consistency}. To exploit the norm-continuity of these spaces, we will use Theorem 14.1 in~\cite{wainwright_2019} and Proposition 2.3 in~\cite{penalised_splines_consistency}. We state these results below.
\begin{theorem}
    Let $\dotprod{h}{g}{\calb{n}} := \sum_{i = 1}^{\calb{n}} h ( \hat{f} (\X_i) ) g ( \hat{f} (\X_i) ) / \calb{n}$ be the empirical dot-product and $\| \cdot \|_{\calb{n}}$ the associated norm. Since $\calb{N} \log (\calb{n}) / \calb{n} \to 0$, we have
    \begin{equation}\label{eq: spline proofs norm-continuity}
        \sup_{g \in \mathcal{G}} \big | \| g \|_{\calb{n}} - \| g \|_2 \big | = o_{\prob} (1), \, \, \, \, \mathrm{and} \, \,\, \,  \sup_{g \in \mathcal{S}_N} \big | \| g \|_{\calb{n}} - \| g \|_2 \big | = o_{\prob} (1).
    \end{equation}
\end{theorem}
We will also use global convergence results. These are Theorems 3.1--3.2 and Corollary 3.3 of~\cite{penalised_splines_consistency}. Let $\mathcal{L}_{\ell pm} (\calb{\boldsymbol{Y}}; \, g)$ be the $L_2$-norm penalised spline loss in~\eqref{eq: smooth spline criterion}. 
\begin{theorem}[\citealp{penalised_splines_consistency}]
    Let $\psibar : = \mathrm{argmin}_{g \in \mathcal{S}_N } \{\mathbb{E} \, [\mathcal{L}_{\ell pm} (\calb{\boldsymbol{Y}}; \, g) ] \} $ be the theoretical loss-minimiser, $\psibiashat = \mathrm{argmin}_{g \in \mathcal{S}_N } \{\mathcal{L}_{\ell pm} (\calb{\boldsymbol{Y}}; \, g) \}$ be the empirical one, and $\psitrue \in \mathcal{G}$ be the {\em``true"} unknown lifted regression function. Then, we have that
    \begin{equation}\label{eq: spline proofs global convergence}
        \| \psibiashat - \psitrue \|_{\infty} = o_{\prob} (1), \, \, \,  \| \psitrue - \psibar \|_{\infty} = o(1), \,\, J_2 (\psibiashat) = O_{\prob} (1), \, \,  J_2 (\psibar) = O (1).
    \end{equation}
\end{theorem}
\par
We further exploit reproducing kernel Hilbert spaces (RKHS). See Chapter 12 in~\cite{wainwright_2019} for an overview of RKHS. Define the following kernel using Proposition~\ref{prop: spline proof orthonormal expansion}, 
\begin{equation}\label{eq: spline proofs spline kernel}
    \kernel{s}{t} := \sum_{\nu} \frac{\phi_{\nu} (s) \, \phi_{\nu} (t)}{1 + \lambda \rho_{\nu}}, \, \,  (s, \, t) \in [y_{\min}, \, y_{\max}] \times [y_{\min}, \, y_{\max}].
\end{equation}
By Mercer's Theorem~\citep[Theorem 12.20][]{wainwright_2019} and Corollary 12.26 in~\cite{wainwright_2019}, $\kernel{\cdot}{\cdot}$ induces the RKHS, say $\mathcal{H}_{\lambda} \subsetneq \mathcal{W}^2$, with inner product $\sum_{\nu} (1 + \lambda \rho_{\nu} ) \langle g, \, \phi_{\nu} \rangle \langle h, \, \phi_{\nu} \rangle$.
\begin{proposition}\label{prop: spline proofs L2 implies sup convergence}
    Assume that the sequence of functions $\{ g_{m} \} \subset \mathcal{G}$ converges in $L_2$-norm to $g \in \mathcal{G}$. Then, it also converges uniformly (pointwise for all $t \in  [y_{\min}, \, y_{\max}]$ at the same rate). 
\end{proposition}
\begin{proof}
    By assumption, we have $\| g_{m} - g \|_2 \to 0^+$, so by~\eqref{eq: spline proofs sup-norm linf stability}, we see that $\| g_{m} - g \|_{\infty} \leq A_{\mathcal{G}} \| g_{m} - g \|_2$ for every $m \in \mathbb{N}$. Thus, $\| g_{m} - g \|_{\infty} \to 0^+$.
\end{proof}
By assumption~\eqref{eq: spline cond proper lifted support}, we see that the Jacobian of the function $\X_o \mapsto (\hat{f} (\X_o), \, X_2, \dots, \, X_d) \in \mathbb{R}^d$ is well-defined for all $\X_o \in \hat{f}^{-1} ( [y_{\min}, \, y_{\max}])$. Thus, the density function of $\hat{f} (\X_o)$ is continuous and bounded given that $p_{\X}$ is bounded. Hence, there exist two constants $0 < a_1 \leq a_2 < \infty$ such that
\begin{equation}\label{eq: spline proofs proper support}
    a_1 \leq p_{\hat{f}} (\hat{f} (\x_o) ) \leq a_2, \, \, \, \text{for all} \, \, \, \x_o \in \hat{f}^{-1} ( [y_{\min}, \, y_{\max}]).
\end{equation}
In view of~\eqref{eq: spline proofs proper support} we assume the bounded mesh ratio condition~\citep{penalised_splines_consistency}, that is, there exist two constants $a_1, a_2$ that do not depend on $\ncal$ such that
\begin{equation*}
    a_1 \leq \frac{\max (y_{k + 1} - y_{k})}{\min (y_{k + 1} - y_{k})} \leq a_2.
\end{equation*}
\subsection{Proof of Lemma~\ref{lem: L2 spline lpm influence}}
{\bf Step I:} Show that the solutions are in $\mathcal{G}$ and admit the kernel expansion in~\eqref{eq: spline proofs kernel expansion}.
\begin{proof}
Let $\beta (t) = \beta_0 + \beta_1 t$ denote the line around which $\psitrue$ concentrates, $\| \psitrue - \beta (\cdot) \|_{\infty} \leq K_{\mathcal{G}}$. By~\eqref{eq: spline proofs global convergence} and the triangle inequality, we see that
\begin{equation*}
    \| \psibar - \beta (\cdot) \|_{\infty} = \| \psibar - \psitrue + \psitrue - \beta (\cdot) \|_{\infty} \leq \| \psibar - \psitrue \|_{\infty} + \| \psitrue - \beta (\cdot) \|_{\infty} \leq K_{\mathcal{G}} + o(1),
\end{equation*}
where $\psibar \in \mathcal{S}_N$, similarly for $\psibiashat \in \mathcal{S}_N$, we have that
\begin{equation*}
    \| \psibiashat - \beta (\cdot) \|_{\infty} = \| \psibiashat - \psitrue + \psitrue - \beta (\cdot) \|_{\infty} \leq \| \psibiashat - \psitrue \|_{\infty} + \| \psitrue - \beta (\cdot) \|_{\infty} \leq K_{\mathcal{G}} + o_{\prob} (1).
\end{equation*}
Since $J_2 (\psibiashat) = O_{\prob} (1)$ and $J_2 (\psibar) = O (1)$, we conclude by the above that $\psibar, \, \psibiashat \in \mathcal{G}$ with probability one as $\calb{n} \to \infty$. Thus, by Propositions~\ref{prop: spline proof orthonormal expansion}--\ref{prop: spline proofs L2 implies sup convergence}, we see that both admit Fourier expansions that also converge pointwise for all $t \in [y_{\min}, \, y_{\max}]$.
\par
Following~\cite{penalised_splines_consistency}, we see that the theoretical minimiser $\psibar$, and the empirical one $\psibiashat$ satisfy the following stationary equations for any $g \in \mathcal{S}_N$. 
\begin{align}
    &-\frac{1}{\calb{n}} \sum_{i = 1}^{\ncal} (Y_i - \psibiashat ( \hat{f} (\X_i)) ) \, g(\hat{f} (\X_i)) + \lambda J_2 (\psibiashat, \, g) = 0, \label{eq: spline proofs emprical stationary equation} \\
    &-\mathbb{E} \ [ \psitrue (\hat{f} (\X_o) )] + \mathbb{E} \, [ g (\hat{f} (\X_o)) \psibar (\hat{f} (\X_o))] + \lambda J_2 (\psibar, \, g) = 0. \label{eq: spline proofs theoretical minimiser}
\end{align}
By~\eqref{eq: spline proofs proper support}, we see that the expectations in~\eqref{eq: spline proofs theoretical minimiser} exist, and that these (because of the bounds) are equivalent to the usual $L_2$-norm $\| \cdot \|_2$. Substituting the Fourier expansions: $\psibar = \sum_{\nu} \psibar_{\nu} \phi_{\nu}$, $\psibiashat = \sum_{\nu} \psibiashat_{\nu} \phi_{\nu}$, $\psitrue = \sum_{\nu} \psi_{0, \nu} \phi_{\nu}$ and $g = \sum_{\nu} g_{\nu} \phi_{\nu}$ into~\eqref{eq: spline proofs emprical stationary equation}--\eqref{eq: spline proofs theoretical minimiser}, and using Proposition~\ref{prop: spline proof orthonormal expansion},
\begin{align*}
    \sum_{\nu} \psi_{0, \nu} g_{\nu} &= \sum_{\nu} (1 + \lambda \rho_{\nu}) \psibar_{\nu} g_{\nu}, \\
    \sum_{\nu} g_{\nu} \dotprod{\calb{\boldsymbol{Y}}}{\phi_{\nu}}{\ncal} &\overset{(i)}{=} \sum_{\nu} g_{\nu} \dotprod{\psibiashat}{\phi_{\nu}}{2} + \lambda \rho_{\nu} \psibiashat_{\nu} g_{\nu} + o_{\prob} (1),
\end{align*}
where $(i)$ follows from using~\eqref{eq: spline proofs norm-continuity} for $ \dotprod{\psibiashat}{\phi_{\nu}}{\ncal} = \dotprod{\psibiashat}{\phi_{\nu}}{2} + o_{\prob} (1)$. Since these two equations hold for all $g_{\nu} \in \mathbb{R}$, we have that the solution is 
\begin{equation}\label{eq: spline proofs coeff solution}
    \psibar_{\nu} = \frac{\psi_{0, \nu}}{ (1 + \lambda \rho_{\nu})}, \, \, \, \mathrm{and} \, \, \, \psibiashat_{\nu} = \frac{\dotprod{\calb{\boldsymbol{Y}}}{\phi_{\nu}}{\ncal}}{(1 + \lambda \rho_{\nu})} + o_{\prob} (1). 
\end{equation}
Recall that $\psibar, \, \psibiashat \in \mathcal{G}$, hence, $\psibiashat - \psibar \in \mathcal{G}$, so we can compute $\psibiashat (t) - \psibar (t)$ for all $t \in [y_{\min}, \, y_{\max}]$ using Fourier expansions, which gives
\begin{align*}
    \psibiashat (t) - \psibar (t) &= \sum_{\nu} (\psibiashat_{\nu} - \psibar_{\nu} ) \phi_{\nu} (t) \\
    &= \frac{1}{\ncal} \sum_{i = 1}^{\ncal} \sum_{\nu} \frac{Y_i \phi_{\nu} (\hat{f} (\X_i)) }{(1 + \lambda \rho_{\nu})} \phi_{\nu} (t) - \sum_{\nu} \frac{\dotprod{\psitrue}{\phi_{\nu}}{2}}{(1 + \lambda \rho_{\nu})} \phi_{\nu} (t) \\
    &\overset{(i)}{=} \frac{1}{\ncal} \sum_{i = 1}^{\ncal} \sum_{\nu} \frac{Y_i \phi_{\nu} (\hat{f} (\X_i)) - \psitrue (\hat{f} (\X_i)) \phi_{\nu} (\hat{f} (\X_i))}{(1 + \lambda \rho_{\nu})} \phi_{\nu} (t) + o_{\prob} (1) \\
    &= \frac{1}{\ncal} \sum_{i = 1}^{\ncal} \sum_{\nu} u_i \frac{\phi_{\nu} (\hat{f} (\X_i))  \phi_{\nu} (t)}{(1 + \lambda \rho_{\nu})} + o_{\prob} (1) \\
    &= \frac{1}{\ncal} \sum_{i = 1}^{\ncal} u_i \kernel{t}{\hat{f} (\X_i)} + o_{\prob} (1),
\end{align*}
where $(i)$ follows from the definition of $\dotprod{\cdot}{\cdot}{\ncal}$ and~\eqref{eq: spline proofs norm-continuity}. Given $\hat{f} (\X_i) = \hat{y}_i$, we arrive at~\eqref{eq: spline proofs kernel expansion}.
\end{proof}
{\bf Step II:} Show that the variance is of the order in~\eqref{eq: spline proofs variance term}.
\begin{proof}
    We begin by bounding $\dotprod{\kernel{t}{\cdot}}{\kernel{t}{\cdot}}{2}$ by $\kernel{t}{t}$. By direct substitution and recalling that $\dotprod{\phi_{\nu}}{\phi_{\mu}}{2} = \mathbb{I} (\nu = \mu)$ and $(1 + \lambda \rho_{\nu}) \geq 1$ for all $\nu \in \mathbb{N}$, we see that
    \begin{equation}\label{eq: spline proofs dot prod bound by kernel}
        \dotprod{\kernel{t}{\cdot}}{\kernel{t}{\cdot}}{2} = \sum_{\nu} \frac{\phi_{\nu}^2 (t) }{(1 + \lambda \rho_{\nu} )^2} \leq \sum_{\nu} \frac{\phi_{\nu}^2 (t) }{(1 + \lambda \rho_{\nu} )} = \kernel{t}{t}.
    \end{equation}
    We proceed to bound $\kernel{t}{t}$ for all $t \in [y_{\min}, \, y_{\max}]$. Following~\cite{local_polynomial_spline}, we notice that for all $g \in \mathcal{G}$, by the Cauchy-Schwarz inequality, we have
    \begin{equation*}
        \sup_{(g_{\nu})} \frac{|\sum_{\nu} g_{\nu} \phi_{\nu} (t)|}{\sqrt{\sum_{\nu} g_{\nu}^2}} = \sqrt{\sum_{\nu} \phi_{\nu}^2 (t) }, 
    \end{equation*}
    hence, we see by direct comparison that
    \begin{align*}
        \sup_t \sqrt{\sum_{\nu} \phi_{\nu}^2 (t) } &=  \sup_t \sup_{(g_{\nu})} \frac{|\sum_{\nu} g_{\nu} \phi_{\nu} (t)|}{\sqrt{\sum_{\nu} g_{\nu}^2}} \\
        &\leq \sup_{(g_{\nu})} \frac{\sup_t |\sum_{\nu} g_{\nu} \phi_{\nu} (t)|}{\sqrt{\sum_{\nu} g_{\nu}^2}} \\
        &= \sup_{g \in \mathcal{G}} \left \{ \frac{\| g \|_{\infty}}{ \| g \|_2} \right \} = A_{\mathcal{G}},
    \end{align*}
    which, combined with $\sup_t \phi_{\nu}^2 (t) \leq \sup_t \sum_{\nu} \phi_{\nu}^2 (t)$, gives for all $g \in \mathcal{G}$
    \begin{equation}\label{eq: spline proofs bound for eigen sum}
        \sup_t \phi_{\nu}^2 (t) \leq A^2_{\mathcal{G}}.
    \end{equation}
    We can now bound $\kernel{t}{t}$ as follows
    \begin{align}
        \sup_t \, |\kernel{t}{t}| &= \sup_t \sum_{\nu} \frac{\phi_{\nu}^2 (t) }{(1 + \lambda \rho_{\nu} )} \nonumber \\
        &\overset{(i)}{\leq} A^2_{\mathcal{G}} \sum_{\nu} (1 + \lambda \, \rho_{\nu})^{-1} \nonumber \\
        &\overset{(ii)}{\lesssim} A^2_{\mathcal{G}} \, \lambda^{-1/4}, \label{eq: spline proofs kernel bound}
    \end{align}
    where $(i)$ follows from~\eqref{eq: spline proofs bound for eigen sum}, and $(ii)$ from~\eqref{eq: spline proofs bound for eigenvalue sum}. Interestingly, this bound recovers the one we get when plugging in the approximation suggested by~\cite{silverman_kernel}.
    \par
    Recall that $u_o$ is orthogonal to any measurable (e.g., continuous) function of $\hat{f} (\X_o)$, that is, $\mathbb{E} \, [u_o \, h (\hat{f} (\x_o)) \mid \hat{f} (\X_o) = \hat{f} (\x_o))] = 0$ for all continuous $h$, and that the $u_i$ are independent. Thus, by the tower property of expectation, we see that $\mathbb{E} \, [u_i \mid \hat{f} (\X_i) ] \, \kernel{t}{\hat{f} (\X_i)} ] = 0$. We are ready to bound the variance term. For all $t \in [y_{\min}, \, y_{\max}]$, we have that
    \begin{align*}
        \mathbb{E} \, [| \psibiashat (t) - \psibar (t) |^2] &= \frac{1}{\ncal^2} \sum_{i = 1}^{\ncal} \mathbb{E} \, [u_i^2 \mid \hat{f} (\X_i) ] \, \kernel{t}{\hat{f} (\X_i)}^2 \\
        &\overset{(i)}{\leq} \frac{M_u}{\ncal} \left ( \frac{1}{\ncal} \sum_{i = 1}^{\ncal} \kernel{t}{\hat{f} (\X_i)} \, \kernel{t}{\hat{f} (\X_i)} \right ) \\
        &= \frac{M_u}{\ncal} \dotprod{\kernel{t}{\cdot}}{\kernel{t}{\cdot}}{\ncal} \\
        &\overset{(ii)}{=} \frac{M_u}{\ncal} \dotprod{\kernel{t}{\cdot}}{\kernel{t}{\cdot}}{2} + o_{\prob} (1) \\
        &\overset{(iii)}{\leq} \frac{M_u}{\ncal} \kernel{t}{t} + o_{\prob} (1),
    \end{align*}
    where $(i)$ follows from the assumption $0 < \sup_{\x} \sigma_{u}^2 (\x) \leq M_u$, $(ii)$ from~\eqref{eq: spline proofs norm-continuity} and $(iii)$ from~\eqref{eq: spline proofs dot prod bound by kernel}. Thus, by~\eqref{eq: spline proofs kernel bound} we arrive at
    \begin{equation}
        \sup_t  \mathbb{E} \, [| \psibiashat (t) - \psibar (t) |^2] \lesssim \frac{M_u \, A^2_{\mathcal{G}}}{\ncal \, \lambda^{1/4}},
    \end{equation}
    which results in~\eqref{eq: spline proofs variance term}.
\end{proof}
{\bf Step III:} Show that the bias is of the order in~\eqref{eq: spline proofs bias term}.
\begin{proof}
    For any $g \in \mathcal{G}$, we have that $g_{\nu} \phi_{\nu} \in \mathcal{G}$, and by~\eqref{eq: spline proofs sup-norm linf stability} we see that
    \begin{equation}\label{eq: spline proofs bounding eigen function in bias term}
        \| g_{\nu} \phi_{\nu} \|_{\infty} \leq A_{\mathcal{G}} \, \| g_{\nu} \phi_{\nu} \|_2 = A_{\mathcal{G}} \, | g_{\nu} | < \infty.
    \end{equation}
    Further, since $J_2 (g ) < \infty$ for all $g \in \mathcal{G}$, we must have that $|g_{\nu}|^2$ goes to zero faster than $\rho_{\nu}$ grows, that is, $g_{\nu}^2 \rho_{\nu} = o(1)$ for all $g \in \mathcal{G}$. The closer $g$ is to a line, the faster the Fourier coefficients $g_{\nu}$ decay to zero, by construction, all $g \in \mathcal{G}$ concentrate around a line, thus, by assumption~\eqref{eq: spline family assumption}, there exists a constant $B_{\mathcal{G}} > 0$ such that 
    \begin{equation}\label{eq: spline proofs bias bound using eigenvalue decay}
        B_{\mathcal{G}} \nu \geq | g_{\nu} | \rho_{\nu}, \, \, \, \, \text{for all} \, \, \, g \in \mathcal{G}.
    \end{equation}
    Recall that $0 < 1 - \lambda^{1/4} < 1$, hence, by~\eqref{eq: spline proofs bound for eigenvalue sum}, we have that
    \begin{equation*}
        \sum_{\nu} (1 + \lambda \rho_{\nu} )^{-1} \lesssim \sum_{\nu} (1 - \lambda^{1/4})^{\nu} = \lambda^{-1/4},
    \end{equation*}
    which combined with~\eqref{eq: spline proofs bias bound using eigenvalue decay} and the dominated convergence theorem~\cite{bartle1982elements} gives
    \begin{equation}\label{eq: spline proofs bias sum bound}
         \sum_{\nu} |g_{\nu} | \rho_{\nu} \, (1 + \lambda \rho_{\nu} )^{-1} \lesssim B_{\mathcal{G}} \sum_{\nu} \nu (1 - \lambda^{1/4})^{\nu} \lesssim \lambda^{-1/2} \, \, \, \text{for all} \, \, \, g \in \mathcal{G}.
    \end{equation}
    Now, we examine the penalty evaluated at $g, \, \kernel{\cdot}{t}$, 
    \begin{align*}
        | J_2 (g, \, \kernel{\cdot}{t}) | &= \left | \sum_{\nu} \frac{g_{\nu} \rho_{\nu}}{(1 + \lambda \rho_{\nu}) } \phi_{\nu} (t) \right | \\
        &\overset{(i)}{\leq} \sum_{\nu} \frac{ A_{\mathcal{G}} \, |g_{\nu}| \rho_{\nu}}{(1 + \lambda \rho_{\nu}) } \\
        &\overset{(ii)}{\lesssim} A_{\mathcal{G}} \, B_{\mathcal{G}} \, \lambda^{-1/2},
    \end{align*}
    for all $g \in \mathcal{G}$ and $t \in [y_{\min}, \, y_{\max}]$, where $(i)$ follows from~\eqref{eq: spline proofs bias bound using eigenvalue decay} and $(ii)$ from~\eqref{eq: spline proofs bias sum bound}. Hence,  
    \begin{equation}\label{eq: spline proofs penalty bound for kernel}
        \sup_t \, \sup_{g \in \mathcal{G}} | J_2 (g, \, \kernel{\cdot}{t}) | = O (\lambda^{-1/2}).
    \end{equation}
    We are ready to bound the bias term, by direct computation we see that
    \begin{align}
        \sup_t |\psibar (t) - \psitrue (t)|  &= \sup_t \left | -\lambda \sum_{\nu} \frac{\rho_{\nu} \psi_{0, \nu}}{1 + \lambda \rho_{\nu}} \phi_{\nu} (t) \right | \nonumber \\
        &= \sup_t \left | - \lambda \, J_2 (\psitrue, \kernel{\cdot}{t}) \right | \nonumber \\
        &\overset{(i)}{\leq} \lambda \, O ( \lambda^{-1/2}) \nonumber \\
        &= O (\lambda^{1/2}), \label{eq: spline proofs bias bound}
    \end{align}
    where $(i)$ follows from~\eqref{eq: spline proofs penalty bound for kernel}. By squaring both sides of~\eqref{eq: spline proofs bias bound}, we arrive at~\eqref{eq: spline proofs bias term}.
\end{proof}
{\bf Step IV:} Show that the lifted prediction influence function in Lemma~\ref{lem: L2 spline lpm influence} satisfies~\eqref{eq: asymp: linear lpm}.
\begin{proof}
    Let $h_{\lambda} := \lambda^{1/4}$ and recall that $\ncal h_{\lambda} \to \infty$. By~\eqref{eq: spline proofs variance term}, and Markov's inequality we see that
    \begin{equation}\label{eq: spline proofs spline variance to zero}
         \mathbb{E} \, [ (\psibiashat ( t ) - \psibar( t) )^2 ] \lesssim 1 / (\ncal h_{\lambda}) + o_{\prob} (1), \, \, \, \, \prob ( |  \psibiashat (t) - \psibar (t) | > \delta ) \lesssim 1  / (\ncal h_{\lambda} \delta^2) \to 0^+,
    \end{equation}
    thus, we get that $ \sqrt{\ncal h_{\lambda}} \{  \psibiashat (t) - \psibar (t) \} = o_{\prob} (1)$. By construction $\lambda \ncal h_{\lambda} \to 0$, so by~\eqref{eq: spline proofs bias term}, we get
    \begin{equation}\label{eq: spline proofs bias to zero}
        \ncal h_{\lambda} \, | \psibar (t) - \psitrue (t) |^2 = O_{\prob} (\lambda \ncal h_{\lambda}) = o_{\prob} (1).
    \end{equation}
    By combining\eqref{eq: spline proofs spline variance to zero}--\eqref{eq: spline proofs bias to zero}, we arrive at
    \begin{equation}
        \sqrt{ \ncal h_{\lambda}} \{ \psibiashat (t) - \psitrue (t) \} = \sqrt{ \ncal h_{\lambda}} \{ \psibiashat (t) - \psibar (t) \} + \sqrt{\ncal h_{\lambda}} \{ \psibar (t) - \psitrue (t) \} = o_{\prob} (1).
    \end{equation}
    Thus, by substituting~\eqref{eq: spline proofs kernel expansion} and noting that $0 < \sigma_u^2 (\x) \leq M_u < \infty$,  we have that
    \begin{equation*}
         \sqrt{ \ncal h_{\lambda}} \{ \psibiashat (t) -  \psitrue (t) \} = \frac{ \sqrt{ \ncal h_{\lambda}} }{\calb{n}} \sum_{i = 1}^{\calb{n}} \frac{u_i \,  \kernel{t}{\hat{y}_{i}}}{\sigma_{u} (\x_i) } + o_{\prob} (1),
    \end{equation*}
    which shows that the function $\ell_{\psi_{\hat{f}}}$ defined in Lemma~\ref{lem: L2 spline lpm influence} satisfies condition~\eqref{eq: asymp: linear lpm}.
\end{proof}
\subsection{Proof of Theorem~\ref{th: cond coverage spline}}
We combine the results in Theorems~\ref{th: aymp validity of MAPS}--\ref{th: dist-free consistent model properties} with Lemmas~\ref{lem: stability of the lpm}--\ref{lem: L2 spline lpm influence} to show Theorem~\ref{th: cond coverage spline}.
\begin{proof}
    By Lemma~\ref{lem: L2 spline lpm influence} we see that $\psibiashat$ satisfies condition~\eqref{eq: asymp: linear lpm} for all $\x_o \in \hat{f}^{-1} ( [y_{\min}, \, y_{\max}])$. Thus, by Lemma~\ref{lem: stability of the lpm} and Theorem~\ref{th: aymp validity of MAPS}, we see that the lifted bootstrap residuals are Kolmogorov-Smirnov consistent for the lifted error, $$ \| P_{\hat{u} | \hat{f} }^* -  P_{u | \hat{f} } \|_{\infty} = o_{\prob} (1).$$ 
    \par
    By assumption, the errors $u_o$ are $\hat{f}$-contour homoscedastic, so by Lemma~\ref{lem: coverage comparison and optimal f-homos}, we have that coverage for $Y_o \mid \hat{f} (\X_o) = \hat{f} (\x_o)$ is equal to one for $Y_o \mid \X_o = \x_o$ for all $\x_o \in \hat{f}^{-1} ( [y_{\min}, \, y_{\max}])$, and we arrive at~\eqref{eq: maps asymp validity at xo}. Further, if $ \| \hat{f} - f \|_{\infty} = o_{\prob} (1)$ for all $\x_o \in \hat{f}^{-1} ( [y_{\min}, \, y_{\max}])$, then since~\eqref{eq: asymp: linear lpm} holds by Lemma~\ref{lem: L2 spline lpm influence} and Theorem~\ref{th: dist-free consistent model properties}, we arrive at~\eqref{eq: maps spline optimal consistent}.
\end{proof}
\setlength{\bibsep}{0pt plus 0.3ex}
\bibliography{jmlr-style-file-master/main_manuscript/references}
}

\end{document}